\documentclass{article}

\PassOptionsToPackage{numbers}{natbib}
\PassOptionsToPackage{hyphens}{url}
\usepackage[preprint]{neurips_2026}

\usepackage{amsmath,amssymb,amsthm,mathtools}
\usepackage{amsfonts}       
\usepackage[utf8]{inputenc} 
\usepackage[T1]{fontenc}    

\usepackage{url}            
\usepackage{booktabs}       
\usepackage{caption}
\usepackage{nicefrac}       
\usepackage{microtype}      
\usepackage{xcolor}         

\usepackage{dashbox}
\usepackage{enumerate}
\usepackage{enumitem}
\usepackage{float}

\usepackage{graphicx}

\usepackage{longtable}
\usepackage{tabularx}
\usepackage{mathrsfs}
\usepackage{mathtools}
\usepackage{mdframed}
\usepackage{microtype}
\usepackage{multirow}
\usepackage{pgfplots}
\pgfplotsset{compat=newest}
\usepackage{pifont}
\newcommand{\cmark}{\text{\ding{51}}}
\newcommand{\xmark}{\text{\ding{55}}}%

\usepackage{relsize}
\usepackage{stackengine}
\usepackage{subcaption}
\usepackage{tcolorbox}
\usepackage{thmtools,thm-restate}
\usepackage{tikz}
\usepackage{tikz-cd}
\usetikzlibrary{calc,shapes.geometric,arrows,fit,matrix,positioning} 
\tikzset
{
    treenode/.style = {circle, draw=black, align=center, minimum size=0.5cm},
    subtree/.style  = {isosceles triangle, draw=black, align=center, minimum height=0.5cm,
    minimum width=1.2cm, shape border rotate=90, anchor=north}
}
\usepackage{todonotes}
\usepackage{xcolor}
\usepackage{url}
\usepackage{xspace}

\usepackage{hyperref}
\hypersetup{
  colorlinks=true,
  citecolor=blue,
  urlcolor=black
}
\usepackage[capitalise,noabbrev]{cleveref}

\newcommand{\gls}[2]{\hyperlink{gls:#1}{#2\textsuperscript{\tiny$\diamond$}}}
\newcommand{\glstarget}[2]{\hypertarget{gls:#1}{}\textbf{#2}}

\theoremstyle{plain}
\newtheorem{definition}{Definition}

\newtheorem{myremark}[definition]{\textbf{Remark}}
\let\oldmyremark\myremark
\renewcommand\myremark{\oldmyremark{\!\!\!\bf}\normalfont}

\title{Proposals and challenges for zero-knowledge verification of frontier AI training}

\author{%
  Pierre Peigné - Lefebvre \\
  General-Purpose AI Policy Lab
  Paris \\
  \texttt{ppeigne@gpaipolicylab.org} \\
\And
  Ky Nguyen \\
  Sorbonne Université, CNRS, LIP6 \\
  F-75005 Paris, France\\
  \texttt{ky.nguyen@lip6.fr} \\
  \AND
    Paul Wang \\
   Sorbonne Université, CNRS, LIP6 \\
  F-75005 Paris, France\\
  \texttt{pwang@phare.normalesup.org} \\
}

\newcommand{\N}{\mathbb{N}}

\newcommand{\zo}{\{0,1\}}

\newcommand{\defeq}{\stackrel{ {\scriptscriptstyle def}}{=}}

\newcommand{\concat}{\mathbin{\|}}

\DeclarePairedDelimiterX{\innerp}[2]{\langle}{\rangle}{#1,#2} 

\NewDocumentCommand\range{om}{\IfNoValueTF{#1}{[{#2}]}{\ifnum1=0#1\relax[{#2}]\else[#1;{#2}]\fi}}

\newcommand\sdash{\!\!\!\:\:\text-}

\newcommand{\negl}{\operatorname{negl}}

\makeatletter
\newcommand{\boldparagraph}{%
  \@startsection{paragraph}{4}%
  {\z@}{3ex \@plus 1ex \@minus .2ex}{-0.4em}%
  {\normalfont\normalsize\textbf}%
}

\makeatother

\newcommand{\cX}{\mathcal{X}}

\newcommand{\cW}{\mathcal{W}}
\newcommand{\cA}{\mathcal{A}}
\newcommand{\cB}{\mathcal{B}}
\newcommand{\cC}{\mathcal{C}}

\newcommand{\cE}{\mathcal{E}}
\newcommand{\cF}{\mathcal{F}}

\newcommand{\cM}{\mathcal{M}}
\newcommand{\ctildeM}{\widetilde{\mathcal{M}}}

\newcommand{\cO}{\mathcal{O}}

\newcommand{\cR}{\mathcal{R}}

\newcommand{\cT}{\mathcal{T}}

\newcommand{\AND}{\wedge}

\newcommand{\sep}{\lambda} 

\newcommand{\crs}{\ensuremath{{\mathsf{crs}}}\xspace}

\newcommand{\EECom}{\ensuremath{\mathsf{EECom}}\xspace} 

\NewDocumentCommand\AdvSBindExt{O{\EECom}O{\cA}}{\Adv^{\mathsf{s\text{-}bind\text{-}ext}}_{#1,#2}(\sep)}
\NewDocumentCommand\AdvSSimInd{O{\EECom}O{\cA}}{\Adv^{\mathsf{s\text{-}sim\text{-}ind}}_{#1,#2}(\sep)}

\newcommand{\Adv}{\mathsf{Adv}}

\newcommand\Mes\cM 

\newcommand{\LSSS}{\mathsf{LSSS}}

\newcommand{\notionfont}[1]{\texorpdfstring{\ensuremath{\textsf{#1}}}{#1}}

\newcommand{\aone}{\notionfont{aone}\xspace} 

\newcommand\LObf{\mathsf{LObf}} 

\NewDocumentCommand\AdvLO{O{\LObf}O{\cA}}{\Adv^{\mathsf{lock}}_{#1,#2}(\sep)}

\newcommand{\LF}{\mathsf{LF}}

\NewDocumentCommand\AdvLF{O{\LF}O{\cA}}{\Adv^{\mathsf{keydist}}_{#1,#2}(\sep)}

\newcommand\Eaone{\cE^\aone}
\newcommand\Emcfepos{\cE^{\mathsf{pos}}}
\newcommand\Emcfe{\cE}

\newcommand\dcent{} 
\newcommand\cpa{\mathsf{cpa}}

\newcommand\ochal{{\mathsf{1chal}\sdash}}
\newcommand\pos{{\mathsf{pos}\sdash}}

\newcommand\xxx{{\mathsf{xxx}\sdash}}

\NewDocumentCommand\advd{ooom}{\Adv^{\dcent\mathsf{mc\text{-}w\!\!\!\:\:\text{-}rep}\text{-}#4\cpa}_{#1,#2,#3}(1^\sep)}

\NewDocumentCommand\AdvDOchalXxx{O{\Emcfe}O{\cF^{\,\mathsf{IP,poly}}_{(N_i)^n_{i=1},q,\LSSS}}O{\cA}}{\advd[#1][#2][#3]{\xxx\ochal}}
\NewDocumentCommand\AdvAoNEOchalXxx{O{\Eaone}O{\cF^{\,\mathsf{IP,poly}}_{(N_i)^n_{i=1},q,\LSSS}}O{\cB_2}}{\advd[#1][#2][#3]{\xxx\ochal}}

\NewDocumentCommand\AdvDOchalPosXxx{O{\Emcfepos}O{\cF^{\,\mathsf{IP,poly}}_{(N_i)^n_{i=1},q,\LSSS}}O{\cB_1}}{\advd[#1][#2][#3]{\pos\xxx\ochal}}

\newcounter{openproblem}
\newcommand{\opitem}[2]{%
  \refstepcounter{openproblem}%
  \paragraph{OP-\theopenproblem. #2}%
  \label{op:#1}%
}
\newcommand{\opref}[1]{OP-\ref{op:#1}}

\begin{document}

\maketitle

\begin{abstract}
Proposals for Frontier AI rules often use cumulative training compute as the main criterion
for identifying high-impact models. Recent governance analyses call zero-knowledge proofs promising but currently impractical at
frontier scale~\citep{shavit2023chinchilla,baker2025verifying}. We argue this barrier
reflects the current paradigm, not a fundamental limit. Our frontier dense-pre-training
design combines a pre-committed specification, inter-node network observations, and live Merkle
commitments to intermediate computation. Our reference zero-knowledge virtual machine
(zkVM) provides a succinct proof layer with native BF16/FP32 precompiles. We qualitatively compare
zkVMs, custom circuits, and direct interactive audits by transferability, soundness, and
cost. Public randomness beacons are a separate audit-selection choice.

At challenged positions, proofs check the committed computation against declared
BF16/FP32 semantics bit-for-bit. A private training
specification targets model-architecture confidentiality. Genesis proofs cover
initialisation; then, in-training step proofs verify selected steps across the run.
Ex-ante attestations enforce policy-relevant claims as running
invariants on every proved step. Together they create a governance-auditable training
record. Detection-probability formulas quantify sampling guarantees from the sampling
budget and the share of computation altered. We give numerical examples and explain
what sampling does not cover.

We note, however, that weight smuggling could reuse or merge weights trained elsewhere without charging
that work to the current run. We then argue that regimes with global training budgets
are more robust than regimes with per-run FLOP caps. This advantage depends on
counting all contributing training. We catalogue remaining research and engineering problems for external contributors.

\end{abstract}

\section{Introduction}
Training runs for frontier AI systems are increasingly at the centre of emerging
governance frameworks that rely on claims the trainer makes about its run:
what data was used, what procedure was followed, what compute was consumed.
No technical verification primitive exists to check such claims at frontier scale today,\footnote{The
main non-cryptographic proposal, proof-of-learning~\citep{jia2021pol}, attests a run by
logging intermediate checkpoints together with the data points that produced them; its
security rested on the empirical claim that spoofing a proof costs as much as honest
training --- falsified by Zhang et al.~\citep{zhang2022polattack} and then more broadly
by the scheme's own authors, who conclude that a provably robust mechanism awaits
``further understanding of optimization in deep learning''~\citep{fang2023polbroken}.}
and enforcement rests on trainer self-reporting.
Any future international agreement that would govern the development of frontier AI faces
the same problem at higher stakes~\citep{baker2025verifying,scher2025international}.
Past international agreements on technologies with significant externalities show that the
credibility of substantive commitments usually rests on the compliance architecture
negotiated with them
\footnote{Examples include IAEA safeguards under the NPT (material accountancy, on-site
inspections, and environmental sampling since the 1997 Additional Protocol) and the OPCW
regime under the CWC (industry declarations, routine inspections of declared facilities;
the Article~IX challenge mechanism has never been invoked). The Montreal Protocol is a
partial case: state-level reporting under Article~7 from 1987, with the Non-Compliance
Procedure added in 1992 and atmospheric science as a parallel independent input. The
Biological Weapons Convention is the counterexample: in force since 1975 with no
verification regime.}.
For the option of such an agreement to remain available to states on frontier AI, the
verification primitive must exist before negotiations begin, not after. This paper
proposes such a primitive.

A \emph{verification primitive} for training must produce proof of faithful execution of the
committed procedure, verifiable claims about the run (compute consumed, training regime,
data-content filters), and privacy for the trainer, within a training-overhead budget low
enough to be economically rational at frontier cost. Single-digit-percent overhead, not
factor-of-two. Prior zero-knowledge (ZK) machine-learning systems
(\Cref{sec:current-zkverif-limits}) are \emph{monolithic}\footnote{That is, these prior ZK systems can be performed only \emph{per sample} or \emph{per step}, and not efficiently for a full run.},
and are confined to inference, to LoRA fine-tuning, or to full training of models orders
of magnitude below frontier scale, operate
over finite-field proxies rather than native BF16/FP32 GPU execution, and plateau at
${\sim}13$B parameters.
Alternative verification approaches based on dedicated hardware,
such as \gls{tee}{TEE}-attested accelerators\footnote{Throughout the paper, a
superscript diamond ($\diamond$) marks a term's first use and links it to the glossary
(\Cref{app:glossary}).}, \gls{hbm}{HBM}-monitoring coprocessors~\citep{petrie2025guaranteeable},
and NICs retrofitted with verification chiplets ~\citep{petrie2025flexheg}, face two structural obstacles independent
of cryptographic feasibility. Their silicon is proprietary to a handful of vendors, a
liability for international verification where trust cannot rest on any single firm's
supply chain. The research-and-development cycle for verification-grade custom silicon
(specification, tape-out, formal RTL verification, manufacturing, certification, and
cluster-wide deployment) is realistically six to ten years before the primitive becomes
available to regulators\footnote{Open-hardware alternatives such as flexible hardware-enabled
guarantees~\citep{petrie2025flexheg} address the vendor-concentration problem but push the
deployment timeline further by requiring international coordination on the hardware design
itself.}.

Our introductory discussion thus begs for answering the following research question:
{\it Is verification for frontier AI training, in a zero-knowledge and efficient non-monolithic manner, possible?}

\paragraph{Our contribution.}
This paper answers the above question positively.
We argue that a verification architecture is feasible at frontier scale, deployable within
approximately 36 months:
\textbf{\it (i)} A blueprint for zero-knowledge verification of frontier dense pre-training
at single-digit-percent training-side overhead under sampled coverage (the
${\sim}5{\times}10^{5}$ figure of~\citep{baker2025verifying} is an inference-derived,
exhaustive-coverage estimate, so the two are not directly comparable), verifying bit-exact
conformance to the declared BF16/FP32 execution semantics and targeting
model-architecture confidentiality via a private \texttt{arch\_spec};
\textbf{\it (ii)} A characterisation of the ex-ante attestation as
a detection-grade governance-enforcement primitive, with compute-threshold enforcement as the immediate
use case\footnote{The EU AI Act~\citep{euaiact_art51} sets a rebuttable presumption of
systemic risk for general-purpose AI models at cumulative training compute greater than
$10^{25}$ FLOPs (Art.~51(2), amendable by delegated act under Art.~51(3)); this remains
in force. The United States executive order 14110~\citep{us_eo14110_2023} used $10^{26}$ FLOPs on similar lines
before its revocation in 2025. Both regimes rely on trainer self-reporting.};
\textbf{\it (iii)} An explicit quantification of the probabilistic guarantees of the
protocol's sampling-based checks---detection probability as a function of sampling
budgets and cheating fractions---together with an honest account of their limits
(\Cref{sec:sampling-guarantees});
\textbf{\it (iv)} A
structured roadmap of independently-attackable open problems delineating the
remaining work to reach a fielded verification primitive.

\paragraph{Our main perspectives and suggested open problems.}
From an \emph{architectural} point of view, we circumvent the fixed-point conversion
required by prior zero-knowledge ML systems by anchoring the proof in three
complementary objects: a \emph{pre-committed training specification}, \emph{inter-node network}
observations by an auditor-aligned anchor (physical \gls{tap}{TAP} or attested \gls{smartnic}{SmartNIC}), and
\emph{on-the-fly Merkle commitments} of intermediate computation. The anchors are verified
through a succinct proof layer equipped with native BF16/FP32 precompiles---a generic
zkVM in our reference design, though \Cref{sec:proof-options} compares alternative
instantiations (bespoke SNARK circuits, and interactive audits whose challenges come
from a public randomness beacon)---so the proof checks the declared BF16/FP32
floating-point semantics of the computation rather than a fixed-point
approximation of it.
Our AI verification protocol employs a \emph{genesis proof} at initialisation,
\emph{in-training step proofs} across the run, and \emph{ex-ante attestations} enforcing
policy-relevant claims as running invariants during training.
We highlight that the question of when random local verifiability of ``snapshots''
can yield global verifiability for NP relations (a faithful execution in our case)
is a long-standing subject in theoretical computer science, reflected via the
celebrated \emph{PCP theorem}
(from the seminal~\citep{ALMSS98:PCP} to recent results~\citep{ABSSW26:PCP});
our raw-trace protocol obtains deviation-fraction-dependent detection
(\Cref{sec:sampling-guarantees}) rather than PCP-style global soundness, and
closing that gap is part of the open problems of \Cref{app:open-problems}.
We extend this framework to AI training verification by structuring the training
specifications, network observations, and random-sampling verification
checkpoints/indices during model training (\Cref{architecture,sec:verifying}). A zkVM
with precompiles proves these objects in a form compatible with step composition.
The architecture is
developed for frontier \emph{dense pre-training} as a first target, because it is
architecturally the simplest setting, accounts for the largest single block of frontier
training compute, and is the root of trust for every post-trained or fine-tuned
derivative.
The architecture does not yet cover sparse mixture-of-experts, reinforcement-learning
post-training, multi-datacentre training, or intra-node \gls{nvlink}{NVLink} traffic. We outline
candidate extensions for each; whether they preserve the base protocol and its trust
assumptions without redesign is itself open. The relevant open research and
engineering problems are catalogued in \Cref{app:open-problems}.

\paragraph{Organisation of the appendices.} \Cref{app:open-problems} catalogues the
open problems with success criteria; \Cref{app:zkvm-details} details the zkVM
instantiation and its precompile catalogue; \Cref{appendix:proof-cost} derives proof
cost and trainer-side overhead; \Cref{app:determinism-details} reports determinism
benchmarks; \Cref{app:network-anchor} develops network anchoring;
\Cref{appendix:toy-example} walks a complete two-node example end to end;
\Cref{appendix:arch-spec} specifies the \texttt{arch\_spec} schema;
\Cref{appendix:formalization} formalises the protocol and its security statements
(with a PAC-style comparison); and \Cref{app:glossary} is a glossary of technical
terms. \Cref{app:zkvm-details,appendix:proof-cost} are the prerequisites for the
cost claims; \Cref{appendix:toy-example} is the gentlest entry point.

\section{The prior ZK-ML paradigm and its limits}\label{sec:current-zkverif-limits}

\paragraph{ZK for inference.}
Recent work has applied zero-knowledge proofs to verify ML model inference.
ZKML~\citep{chen2024zkml} builds an optimizing compiler from TensorFlow to halo2 ZK-SNARK
circuits, proving inference for models up to GPT-2 (81M parameters) in approximately one
hour.
zkLLM~\citep{sun2024zkllm} introduces specialized protocols for LLM inference
(\texttt{tlookup} for non-arithmetic operations, \texttt{zkAttn} for the attention
mechanism) with GPU-accelerated sumcheck proving, scaling to 13B parameters in under 15
minutes.
South et al.~\citep{south2024ezkl} propose a general-purpose framework that converts any
ONNX model into verifiable evaluation attestations via ezkl, demonstrating proofs across
CNNs, LSTMs, and small transformers.
These systems enable a model provider to prove that a committed set of weights produces a
claimed output on a given input, without revealing the weights.

\paragraph{ZK for training and fine-tuning.}
Extending ZK beyond inference to the training process is more challenging, as it requires
proving backward passes and parameter updates.
VeriLoRA~\citep{liao2026verilora} is the first framework to achieve this for LoRA
fine-tuning, proving forward propagation, backward propagation, and parameter updates for
a single mini-batch on models up to 13B parameters (OPT-13B, $\sim$250\,s proving time per
step on A100).
Earlier work on full training includes Kaizen~\citep{abbaszadeh2024kaizen}, which
recursively composes per-iteration proofs of gradient descent so that proof size and
verifier time are independent of the number of iterations (${\sim}15$ minutes of proving
per mini-batch iteration for VGG-11 at $10$M parameters, over a quantised fixed-point
encoding), and zkDL~\citep{sun2024zkdl}, which flattens CNN training into a single
circuit architecture (FAC4DNN) for an 8-layer network with 10M parameters.

\paragraph{Structural limitations.}
Recent governance analyses describe zero-knowledge proofs as a promising candidate for
training verification in principle, but currently impractical at frontier
scale~\citep{shavit2023chinchilla,baker2025verifying}.
Shavit~\citep{shavit2023chinchilla} notes that such techniques ``cannot efficiently execute
computationally intensive programs, like long sequences of gradient updates on
billion-parameter models''.
Baker et al.~\citep{baker2025verifying} give a back-of-the-envelope ${\sim}5{\times}10^{5}$
compute-overhead estimate --- derived from zkLLM's \emph{inference}-verification throughput
against plain inference on the same GPU --- and note the loss of model-architecture
confidentiality.
These assessments accurately characterise the existing ZK-ML paradigm.
All systems in this paradigm share two structural constraints.
First, they operate over finite fields, where arithmetic is \emph{exact} and \emph{associative}.
This is a \emph{structural requirement of SNARKs and sumcheck protocols}, not a design choice
(our proposal also proves over a finite field; what it removes is the fixed-point \emph{model}
arithmetic, by arithmetising the bit-level floating-point relation instead).
In this paradigm, the systems reviewed above therefore convert the model to fixed-point
arithmetic before proving.
The proof verifies a \emph{different} computation than what GPUs execute (BF16/FP32 with
\emph{non-associative floating-point} arithmetic and hardware-specific rounding).
For inference this gap is acceptable; for training verification it is not, because the
goal is to prove a specific sequence of GPU operations was performed as claimed.
Second, proving cost makes monolithic proofs of training prohibitively inefficient: the
fastest existing system (zkLLM) takes 15 minutes per forward pass at 13B, far from a
frontier run involving millions of forward-backward steps at hundreds of billions of
parameters across \gls{fsdp}{FSDP}/\gls{tensor-parallelism}{TP}/\gls{pipeline-parallelism}{PP}-distributed clusters that no existing ZK-ML system models.
Both constraints are properties of this paradigm, not of zero-knowledge verification per se.
The architecture proposed below is designed to lift them.
Outside the ZK paradigm entirely, verifiable training over native floating point does
exist: Verde~\citep{arun2025verde} verifies FP32 training steps (forward, backward, and
optimizer-state updates) bitwise-reproducibly via reproducible operators and refereed
delegation among $n$ compute providers of whom one is assumed honest, and optimistic
verifiable training~\citep{srivastava2024optimistic} replicates FP32 training across
heterogeneous GPUs by controlling hardware nondeterminism. Both rely on a 1-of-$n$
honesty assumption and offer neither zero knowledge nor a succinct transferable
certificate --- but they show that the native-FP target is not intrinsically tied to
fixed-point encodings.

\begin{table}[h!]
  \centering
  \small
  \setlength{\tabcolsep}{5pt}%
  \begin{tabular}{lccccc}
    \toprule
    & ZKML~\citep{chen2024zkml} & zkLLM~\citep{sun2024zkllm} & ezkl~\citep{south2024ezkl} & VeriLoRA~\citep{liao2026verilora} & \textbf{Target} \\
    \midrule
    Scope & Inference & Inference & Inference & Fine-tuning & \textbf{Pre-training} \\
    Arithmetic & Finite field & Finite field & Finite field & Finite field & \textbf{Native BF16} \\
    Max scale & 81M & 13B & 250K & 13B & \textbf{100-1000B (est.)} \\
    Multi-step & Per-sample & Single pass & Per-sample & 1 LoRA step & \textbf{Full run (sampled)} \\
    Distributed & \xmark & \xmark & \xmark & \xmark & \textbf{\cmark} \\
    Native-FP semantics & \xmark & \xmark & \xmark & \xmark & \textbf{\cmark} \\
    \bottomrule
  \end{tabular}
  \caption{Comparison of ZK-ML verification systems.
  Existing approaches prove computations over finite-field proxies of the model. In the
  Target column, ``Native BF16'' means checking the declared BF16/FP32 execution
  semantics; the proof constraints themselves still operate over a finite field.
  The Target column states the architectural objectives of this proposal, justified in
  \Cref{architecture} and \Cref{appendix:proof-cost}; values are not yet measured, and
  run-level coverage is sampled, detection-grade (\Cref{sec:sampling-guarantees}).}
  \label{tab:related-work}
\end{table}

\section{Proposed solution}\label{sec:proposed-solution}
\paragraph{Scope.}
The architecture developed in this paper targets \emph{dense pre-training}. Today's
frontier spans dense models at several hundred billion parameters (e.g.\ Llama~3.1 at
405B) and sparse mixture-of-experts architectures approaching one trillion total
parameters with a small fraction active per token (e.g.\ DeepSeek-V3 at 671B, Kimi~K2 at
roughly 1T with ${\sim}32$B active per token).
We begin with dense pre-training because it is architecturally the simplest setting,
because every post-trained model is rooted in a pre-training run (if pre-training is
not verifiable, nothing downstream is), and because pre-training remains the largest
single block of training compute\footnote{Two further regimes, (\Cref{app:open-problems}, \opref{moe} for MoE and
\opref{rl} for RL), are increasingly load-bearing for frontier capability: \emph{sparse
mixture-of-experts (MoE)}, and \emph{reinforcement-learning (RL) post-training} (RLHF, RLAIF, rule-based RL
for reasoning).}.

\begin{figure}[h!]
  \centering
  \includegraphics[width=12.5cm]{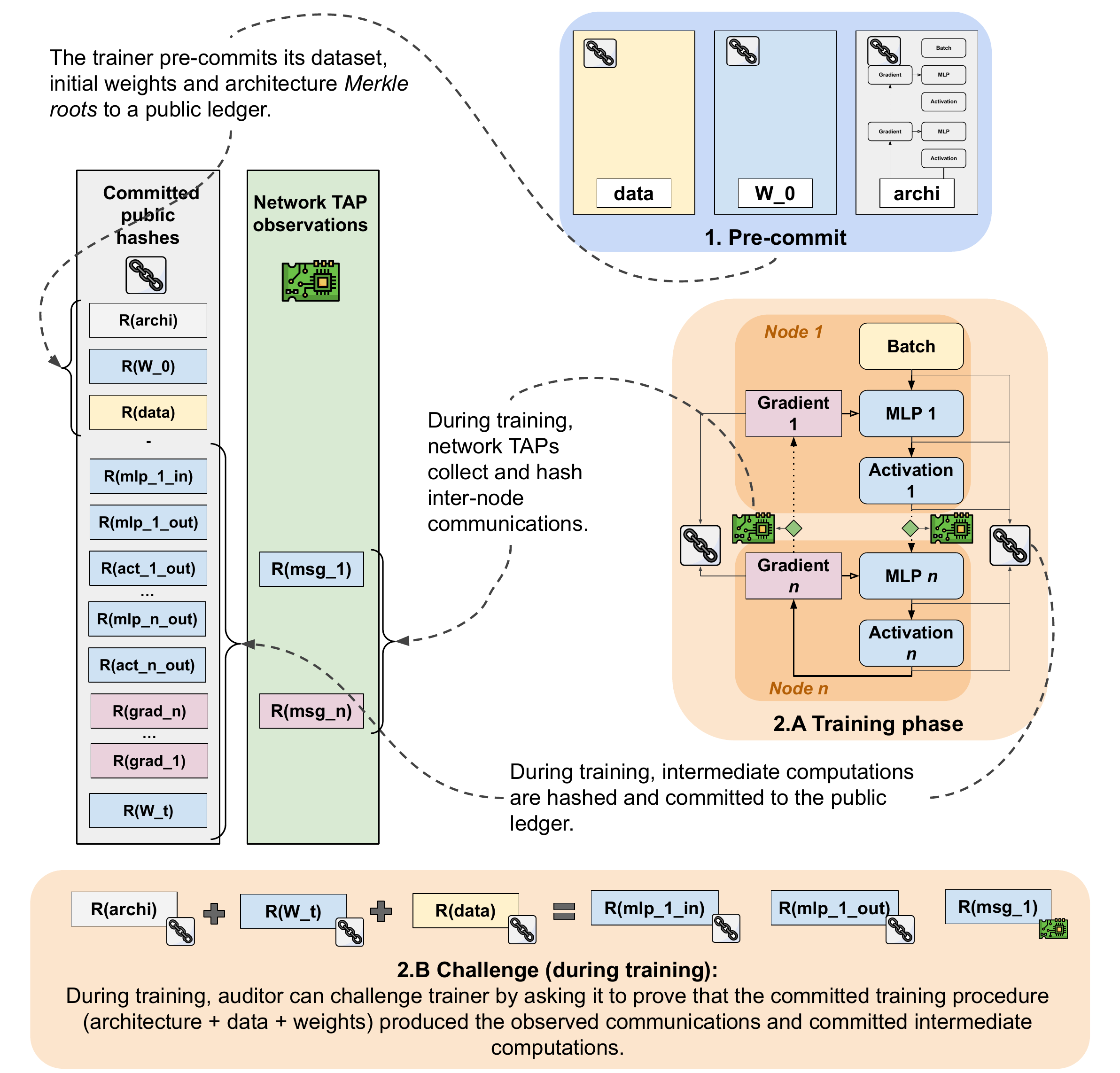}
  \caption{Overview of the verification protocol.
  \emph{(Top)} {\bf 1.} The trainer publishes Merkle roots of the training specification, initial weights, and dataset to a public ledger. \emph{(Middle)} {\bf 2.A} During training, the trainer commits Merkle roots of intermediate tensors per node to a per-step hash chain, while auditor-aligned network TAPs independently hash inter-node communications. \emph{(Bottom)} {\bf 2.B} The auditor asks for proofs that the committed procedure (architecture, weights, data) produced the committed intermediate computations and the observed network messages.
  }
  \label{fig:overview-verif-protocol}
\end{figure}       
\paragraph{Overview.} 
As \Cref{fig:overview-verif-protocol} shows, our verification architecture combines three
complementary trust anchors: {\it (I)} a pre-committed training procedure, model weights,
and dataset (\Cref{pre-commit}),
  bound to the physical cluster via a challenge-response genesis proof (Protocol~\ref{offline});
  {\it (II)} external network TAPs that provide public hashes of ongoing inter-node
  communications and anchor the training to independently observable data
  (\Cref{network-tap});    
  and {\it (III)} Merkle roots of selected intermediate computations, committed on-the-fly
  (\Cref{constant-hashing}) and available for retrospective sampling challenges
  (Protocol~\ref{online}).

\subsection{Modification of current AI training procedures}\label{sec:training-modifications}
We start by detailing modifications that we believe necessary for integration of our AI verification architecture.
\begin{enumerate}[label=\textbf{MOD.~\arabic*},leftmargin=*]
  \item \label{enforce-determinism} {\bf (Enforcing determinism)}
Verification requires bit-exact determinism so the proof checker can reproduce committed
values; this requirement is instantiation-independent---every proof-layer option of
\Cref{sec:proof-options} that binds values bit-exactly relies on it. Standard GPU
training is non-deterministic because floating-point reductions in
\gls{cublas}{cuBLAS}, \gls{nccl}{NCCL} collectives, and attention kernels depend on thread
scheduling.
For \textbf{hardware-level determinism},
per-operator kernel selection (FlashAttention-2 deterministic mode, pinned cuBLAS
algorithms, NCCL \gls{nvls}{NVLS} reductions) achieves bit-exact determinism with
$1.6$--$8.2\%$ end-to-end overhead on Llama~7B FSDP ($8{\times}$H100), dominated by the
attention backward pass. Recent deterministic FlashAttention-3 scheduling~\citep{qiang2026dash}
reduces this further; details in \Cref{app:determinism-details}.
For \textbf{network-level determinism beyond FSDP},
FSDP's collectives are deterministic under NVLS. For PP, the requirement reduces to per-layer
guarantees; for \gls{expert-parallelism}{EP}, it reduces to NVLS-deterministic all-to-all
guarantees. TP is the hard case: its intra-node allreduce over \gls{nvswitch}{NVSwitch} is
\emph{not} deterministic under NVLS above 128\,MB, and deterministic NCCL configurations
incur 64--89\% bandwidth loss. The proposed path for TP is a custom kernel that exploits its
fixed topology; \Cref{app:determinism-details} gives the design.

\item \label{pre-commit} {\bf (Pre-committing public Merkle roots)}
Before training, the trainer publishes:
\(
  h_{\text{commit}} = H\!\big(H(\texttt{arch\_spec}) \;\|\;
  \text{MerkleRoot}(\text{dataset}) \;\|\; \text{MerkleRoot}(w_0)\big)
\).
Merkle paths enable membership proofs for individual batches or layer weights at
logarithmic cost without revealing the full dataset or model.
The genesis proof checks all three components; an in-training step proof
(Protocol~\ref{online}) needs only paths for the relevant batch and layer weights.

\item \label{constant-hashing} {\bf (On-the-fly Merkle root computation)}
A GPU concurrent stream computes Merkle roots of six tensors per layer at every step
(layer input/output, Q, K, V, attention output) using the zkVM-native hash (Poseidon;
precompile~3 of the catalogue in \Cref{tab:precompiles}), published in a \emph{per-step
hash chain} for retrospective verification.
When challenged, the trainer must produce values consistent with the published roots, which
become immutable anchors once the chain is frozen by an auditor-signed terminal anchor
before the audit seed is opened.
%
Pure-HBM-read overhead is $0.2\%$; a 2-pass reduction kernel adds $1.6$--$3.3\%$ (SM
contention). Estimated production overhead with an optimised kernel: $0.5$--$1.5\%$.
The stream can alternatively be offloaded to a SmartNIC/\gls{dpu}{DPU} or CPU
(\Cref{appendix:proof-cost}).

\item \label{network-tap} {\bf (Network anchoring)}
Without an independent witness, all commitments are trainer-produced, leaving open
fabrication of internally consistent commitments for phantom computations. Both tiers
commit a Merkle tree over inter-node collective payloads via SHA-256 (precompile~8,
\Cref{tab:precompiles}), enabling $\log n$-depth path verification at challenge time.
At \textbf{Tier~1 (auditor-owned physical TAP)},
a passive auditor-installed device captures frames at wire speed, canonicalises payloads,
and hashes them into the Merkle tree. Under pinned NCCL the wire sequence is a
deterministic function of step inputs, so the trainer reconstructs the tree from a replay
(\Cref{enforce-determinism}); trust root is physical.
At \textbf{Tier~2 (attested SmartNIC)}, signed DPU firmware exposes its root over an attested channel rooted in secure boot and
device identity; trust root is cryptographic. Weaker than Tier~1 against firmware or
supply-chain adversaries, but deployable on existing cluster infrastructure.
%
{\bf Tier~2 }serves early adopters while {\bf Tier~1} infrastructure is standardised; the long-term
goal is uniform Tier~1, required by any regime holding against well-resourced adversaries.
Both tiers observe inter-node traffic only; intra-node NVLink is invisible (hybrid
FSDP+TP covered at the FSDP layer only). Intra-node coverage via GPU-resident trusted
execution is discussed in \Cref{app:open-problems}. The canonical tensor-to-wire mapping
is the central open engineering problem; see \Cref{sec:future-work} and
\Cref{app:network-anchor}.

\item \label{weight-storage} {\bf (Weight storage rolling window)}
The trainer stores a rolling window of weights at every step on NVMe/Lustre/GPFS arrays.
At 405B parameters in BF16, each snapshot is $\approx$810\,GB; 100 steps require 81\,TB.
When challenged on step~$N$, the trainer loads weights directly and runs a single forward
pass. Whole-run sampling requires either full retention or deterministic replay from
durable checkpoints with committed replay metadata (optimizer, data-loader, and
stochastic-component state; kernel/reduction-tree and precision metadata; checkpoint
interval commitments; committed seed schedule). At frontier scale, 81\,TB is modest
relative to existing cluster storage.

\end{enumerate}
Verification costs (genesis proof, sampling challenges) are borne by the proving system
and detailed in \Cref{sec:verifying}.
Total trainer-side overhead at Llama~3.1 405B scale on a \$100M budget\footnote{The
\$100M figure is a round working assumption, not a published estimate for this specific
run. The Llama~3 report~\citep{dubey2024llama3} gives ${\sim}3.8\times10^{25}$ FLOPs on
$16{,}384$ H100s; frontier-run cost accounting varies with the convention used (Epoch
AI~\citep{cottier2024training} reports, e.g., ${\sim}\$40$M amortised hardware and energy
for GPT-4, with staff and R\&D multiplying totals severalfold). The dollar costs in
this paper scale with the assumed budget.} is in~\Cref{tab:training-overhead}.

\begin{table}[h!]
  \centering
  \small
  \begin{tabular}{lll}
    \toprule
    Component & Cost & \% of training \\
    \midrule
    Determinism (\ref{enforce-determinism}, 1.6--8.2\%)          & \$1.6--8.2M & 1.6--8.2\% \\
    Merkle computation (\ref{constant-hashing}, $\sim$0.5--1.5\%) & \$0.5--1.5M & 0.5--1.5\% \\
    Weight storage (\ref{weight-storage}, 81\,TB rolling)   & negligible  & $<$0.1\%   \\
    \midrule
    \textbf{Total (FSDP)} & \textbf{\$2--10M} & \textbf{$\sim$2--10\%} \\
    \bottomrule
  \end{tabular}
  \caption{Estimated training-side overhead at Llama~3.1 405B scale. The dominant cost is
  determinism (attention backward pass). FSDP+TP setups incur additional overhead until a
  custom TP allreduce kernel is deployed. Dollar values use the \$$100$M round working
  assumption of the text and scale with that budget; the determinism range is measured on
  the cited 7B FSDP benchmark, while the Merkle range is estimated.}
  \label{tab:training-overhead}
\end{table}

\subsection{Architecture of the proving system: necessary components}\label{architecture}
\begin{figure}[!tbp]
  \centering
  \includegraphics[width=12.5cm]{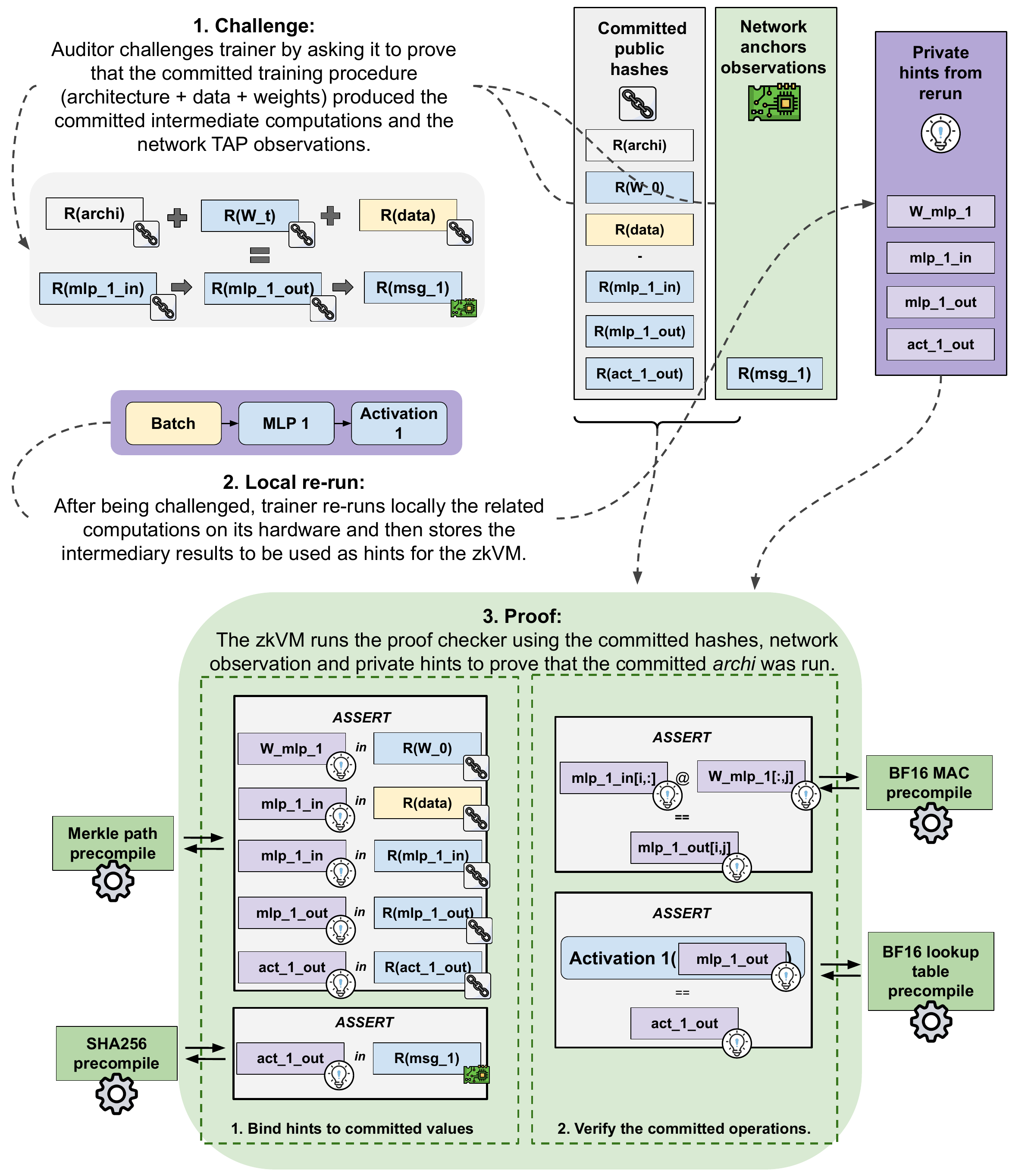}
  \caption{Proof generation from an in-training challenge.
  \emph{(Top)} The auditor challenges the trainer to prove the committed procedure
  (architecture, weights, data) produced the committed intermediates and observed network
  messages.
  \emph{(Middle)} The trainer re-runs the challenged operations locally and supplies the
  intermediate tensors as private hints to the zkVM.
  \emph{(Bottom)} The zkVM verifies in two phases: {\bf 1.} Merkle path verification for
  each private hint, then {\bf 2.} random sampling to verify the bound hints.
  In the diagram, $R(x)$ is the committed Merkle root for $x$, ``in'' is a
  Merkle-membership check, \texttt{@} is matrix multiplication, and each
  \texttt{ASSERT} is a condition enforced by the proof checker.
  }
  \label{fig:proof-flow}
\end{figure}

The proving system has four components: \emph{a training specification} (private, known only to the trainer),
\emph{a proof layer} (run by the trainer; instantiated in our reference design by a zkVM, one option among several --- see \Cref{sec:proof-options}),
\emph{a proof checker} (compiled and given to the proof layer), and
a set of \emph{precompiles} (native circuits the proof layer exposes).
Its central principle is that the proof layer never
re-executes the training computation. When challenged on a step, the trainer loads
stored weights, re-executes forward and backward on GPU, and provides intermediate
tensors as hints; the proof layer only verifies hints are consistent with the Merkle roots
committed during training.
Two data categories enter the proof. \textbf{Public inputs} (Merkle roots) are
committed during training on a concurrent GPU stream and linked into a per-step hash
chain. They become immutable trust anchors when an auditor-signed terminal anchor
freezes the chain before the audit seed is opened. \textbf{Private hints} (full tensor
values) are produced by the host at challenge time by re-executing from stored weights;
they must be verified against the public roots. Verification then proceeds in two
layers. First, each hinted value is bound to its committed root via a \emph{path check},
preventing fabricated hints. Second, the computation on the bound values is verified
using the appropriate \emph{precompile}, preventing incorrect commitments.

The {\bf training specification} \texttt{arch\_spec} is a structured, machine-readable description of one complete
training step. It declares: model architecture as typed operations per layer (\gls{gemm}{GEMM}s
with dimensions, activations, normalisations); numerical precision (BF16 parameters,
FP32 accumulators, IEEE~754 \gls{rne}{RNE}); the distributed communication pattern (allgather,
reduce-scatter); data loading (batch size, shuffling seed, master RNG seed\footnote{Training RNG material is distinct from the auditor's audit seed: the
former determines stochastic execution; the latter selects committed positions to
open after the stream segment is frozen.},
registered stochastic components); and tensor boundaries at which Merkle roots are
committed. The \texttt{arch\_spec} fully describes the
procedure---no separate code commitment is needed; the genesis proof (\Cref{offline})
validates that GPU execution is consistent with it. The concrete structure of \texttt{arch\_spec} is given in \Cref{appendix:arch-spec}.
Commitment structure is in
\Cref{pre-commit}.

The {\bf proof layer} turns the committed material into assurance for the auditor: it
establishes that the committed intermediate tensors are consistent with the operations
declared in \texttt{arch\_spec}. Our reference design instantiates it with a
\emph{zero-knowledge virtual machine} (zkVM) producing succinct, non-interactive proofs of correct
execution of the proof checker, on a \emph{hint-and-verify} model: the host (trainer's
GPU) runs the full computation outside the proof and provides intermediate values as
hints; the guest (proof checker inside the zkVM) does not re-execute, but verifies
hints by calling precompiles against Merkle-committed values. Host computation
contributes zero constraints; only the guest's verification logic does, which reduces
cost by roughly five orders of magnitude versus full re-execution (${\sim}10\times$ from
native precompiles versus software emulation, the rest from sampled rather than
exhaustive coverage; see the footnote in \Cref{app:zkvm-details}). Hint-based witness
computation is standard in ZK-ML inference systems~\citep{kang2022scaling, chen2024zkml,
sun2024zkllm}, and proof-of-training constructions exist for smaller model
classes~\citep{garg2023zktraining,abbaszadeh2024kaizen}; reusing the same native-FP primitives for the
backward pass --- what makes verifying \emph{training}, not just inference, possible ---
is a design this paper specifies but does not yet demonstrate (\opref{zk-backprop}). The zkVM is, however, one instantiation among several: the
proof layer can equally be realised by bespoke SNARK circuits or by an interactive
audit protocol whose challenges come from a live auditor or a public randomness
beacon. \Cref{sec:proof-options} compares these options; zkVM requirements and
mechanics are detailed in \Cref{app:zkvm-details}, and a complete walkthrough is
provided in \Cref{appendix:toy-example}.
%
The {\bf proof checker} is a public, auditable program compiled to the zkVM's instruction
set. It reads \texttt{arch\_spec}, iterates over declared operations, and dispatches
each to the appropriate precompile. The same binary handles models using the
supported operation set. Only \texttt{arch\_spec} differs between a 10M-parameter prototype and a frontier
trillion-parameter model. The \texttt{arch\_spec} is a \emph{private} input.
The auditor sees only $H(\texttt{arch\_spec})$ as part of $h_{\text{commit}}$,
under the zero-knowledge assumptions in \Cref{appendix:formalization}.
From \emph{public metadata} only the number of layers (chain length) and
approximate scale (storage footprint or regulatory filings) can be inferred---both
already disclosed under the regulatory regimes this work serves. Within the proof
layer, the auditor's model-dependent code-review surface reduces to a single public
binary, the precompile specifications, and the zkVM implemented proof system's
cryptographic soundness; the end-to-end trust base additionally includes the anchor
placement, key-custody, determinism, and availability assumptions stated in
\Cref{appendix:formalization}.

The workhorses of the reference design are the \textbf{precompiles}: native
constraint circuits that encode IEEE~754 BF16/FP32 semantics directly over the proof
system's field, so that the proof verifies bit-exact conformance to the declared
BF16/FP32 floating-point semantics rather than a fixed-point approximation of the
model --- the inversion of prior ZK-ML (cf.~\Cref{tab:related-work}). Eight precompiles cover a Transformer
training step (BF16/FP32 MAC chain, BF16 activation lookup, Merkle path, FP32
exp/sqrt/div/log nonlinearities, and SHA-256 for network-anchor reconciliation); the
MAC chain dominates with ${\approx}98\%$ of constraints at 405B scale. The full catalogue
with constraint counts and per-precompile detail is given in
\Cref{app:zkvm-details} (\Cref{tab:precompiles}). GEMMs --- the dominant cost ---
are verified by \emph{interactive per-entry sampling} rather than
exhaustively\footnote{Exact verification for GEMM is costly; we make use of
randomness and interaction to achieve efficiency, at the cost of non-perfect
soundness, see~\Cref{appx:proof-cost:gemm-baseline} for further discussion.}: the
host hints the full result, the auditor selects $k$ random output entries, and the
proof checker recomputes each via the MAC precompile, binding inputs and output to
their committed roots. The resulting detection probabilities, as functions of the
sampling budgets and of the deviation fraction, are the subject of
\Cref{sec:sampling-guarantees}. Approximate
sumcheck~\citep{bitan2025approxsumcheck} suggests a path to recovering Freivalds-like
efficiency for FP GEMMs (\Cref{sec:future-work}).
%
%
Each layer is \emph{decomposed and proved independently}. Cross-layer soundness is enforced
by root chaining: each layer's proof verifies
$R(\text{layer}_\ell\text{\_out})=R(\text{layer}_{\ell+1}\text{\_in})$.
Per-layer sub-proofs run in parallel and are folded into a
${\sim}200$\,KB aggregate via \emph{recursive composition} (details and cost in \Cref{appx:proof-cost:composition}). 
\emph{Fused GPU kernels} (Flash Attention, fused activations) do not materialise intermediate tensors in HBM, so
those tensors cannot be Merkle-committed individually; the proof treats each fused
block as a unit verified end-to-end at its input/output commitments (see details of costs for these fused blocks in~\Cref{appx:proof-cost:commitments}). 

\subsection{Ways to implement the proof layer}\label{sec:proof-options}

The commitments in \Cref{pre-commit,constant-hashing,network-tap} do not determine
how the proof layer is built. Any route must link the committed tensors to the
declared operations. All routes retain the determinism and network-anchor
assumptions.

Three choices are separate:
\begin{itemize}
  \item \textbf{Implementation.} Three overlapping routes are useful: zkVM
  execution, custom-circuit proofs, and direct interactive audits.
  \Cref{tab:proof-layer-options} compares them.
  \item \textbf{Coverage.} A zkVM or custom circuit can prove sampled checks or
  every operation within selected steps. Exhaustive coverage removes sampling
  gaps only within the proved steps. Proof-system assumptions remain.
  \item \textbf{Audit selection.} An auditor, a transcript hash, or a
  \gls{public-beacon}{public randomness beacon} can provide the randomness that
  selects audited steps and entries.
\end{itemize}

\begin{table}[h!]
  \centering
  \footnotesize
  \begin{tabularx}{\textwidth}{>{\raggedright\arraybackslash}p{2.5cm}>{\raggedright\arraybackslash}X>{\raggedright\arraybackslash}X}
    \toprule
    Route & How it checks & Main engineering implications \\
    \midrule
    zkVM proof \emph{(reference)} &
    A public program runs sampled checks through specialised floating-point
    precompiles, which are custom circuits. &
    One checker covers models within its supported operation set and specification.
    Under the stated proof-system assumptions, zero knowledge keeps
    \texttt{arch\_spec} private. A sampled matrix product needs GPU re-execution,
    openings, and $O(kd)$ numerical constraints.
    The current estimate for the recursively combined proof is
    \(\sim 200\,\mathrm{KB}\). \\
    \addlinespace
    Custom-circuit proof &
    A circuit encodes the checker more directly, for example through
    GKR/sumcheck-style checks over exact integer encodings. &
    It may be specialised or reusable. A bit-exact claim requires the same
    sequence of GPU rounding operations (\Cref{sec:current-zkverif-limits}).
    Certificate size depends on the proof system. We do not establish a cost
    ranking between the proof implementations. \\
    \addlinespace
    Direct interactive audit &
    The checker receives values and Merkle paths linking them to their commitments.
    It tests those values against the disclosed operation specification. &
    This avoids cryptographic proving only when those details are disclosed. Either
    proof implementation can instead provide a zero-knowledge proof. Its cost must
    then be counted. The artefact is an authenticated transcript, plus any proofs. \\
    \bottomrule
  \end{tabularx}
  \caption{Engineering differences for a fixed set of checks. Coverage, privacy,
  randomness for audit selection, and storage schedules are shared choices.
  Here \(k\) is the number of sampled entries and \(d\) is the inner dimension.
  The zkVM proof size is a reference-design estimate, not a general property.}
  \label{tab:proof-layer-options}
\end{table}

\paragraph{How much to check and what to store.}
For sampled matrix products, work grows with the number of samples. Under the same
uniform, unpredictable sampler, the local guarantees and floors in
\Cref{sec:sampling-guarantees} apply. Exhaustive coverage instead makes work grow
with the full output size. At frontier dimensions, that means substantially more
proving work per step.

Either proof implementation can combine successive proofs into one compact proof
(recursive accumulation). Incrementally verifiable computation (IVC) maintains
that proof as certified steps are appended during training. Either can also
finish the required proofs before deleting the supporting data (a
proof-before-eviction schedule). How long data must be kept depends on the challenge
and proving schedule, not the implementation. Pending checks need stored opening
material or deterministic replay fixed by prior commitments. An accumulated proof
cannot answer later challenges outside the statement it certifies.

\paragraph{Audit selection and third-party verification.}
The proof system's internal challenges differ from the audit's selection of steps
and entries. \gls{fiat-shamir}{Fiat--Shamir} can make the proof object
non-interactive and publicly verifiable. Its soundness analysis must account
separately for any system-dependent hash-query loss. The audit still needs its own
rule for selecting steps and entries.

The reference audit uses an auditor's commit/reveal seed
(\Cref{sec:protocol-sec-stmts}). Releasing that seed is still an auditor action.
Once opened, the seed lets others reproduce the selected positions. It does not
establish unbiased choice, secrecy before the freeze, or non-collusion. Public
proof verification and confidence in the selection are therefore distinct. The
same selection assumptions apply to direct and proof-wrapped audits.

A \gls{transferability}{transferable} transcript convinces third parties, not only
the initiating auditor. This requires an authenticated public transcript and a
public pass/fail rule that anyone can apply. Interaction alone does not prevent
transferability.

A public transcript hash can remove the separate seed-opening action. However, the
trainer can test candidate commitments until the resulting positions miss a
deviation. This is \gls{grinding}{grinding}. With \(Q\) tested candidates, the miss
probability can rise by a factor of up to \(Q\). Compensation needs a
logarithmic-in-\(Q\) increase in samples, as described in
\Cref{sec:sampling-guarantees}. Exhaustively proved steps have no audit samples to
grind.

A public beacon is unpredictable before release and publicly verifiable
afterwards. Candidate designs include threshold beacons such as drand~\citep{syta2017scalable},
\gls{vdf}{verifiable delay functions}~\citep{boneh2018vdf}, and TEE-attested
generators. Before the relevant beacon output, the trainer must bind one admissible
commitment per audit round. The protocol also needs an authenticated public
transcript and a public pass/fail rule that anyone can apply. Abort or non-response
must count as failure.

These rules make retries later and publicly countable. If the system admits \(Q\)
candidates, the miss probability can still rise by a factor of up to \(Q\)
(\opref{beacons}). Trust in the beacon, its availability on schedule, and
registration infrastructure remain required. A beacon changes assumptions about
randomness and event ordering. It changes neither coverage nor the cost of checking
a fixed challenge.

The reference design uses a zkVM so that supported models can share one checker.
The authors are also studying beacon-timed and hybrid schedules. One combines
non-interactive spot-check proofs with additional bit-exact checks scheduled by a
beacon (\Cref{sec:hybrid}, \opref{beacons}--\opref{hybrid}). The windowed audit in
\Cref{appendix:formalization} sketches one schedule. It fixes a pre-registered
public-beacon rule before each storage window. The same uniform, unpredictable
sampler preserves the local sampling formulas. However, full protocol soundness
needs separate accounting for each window and a model of timing and the public
bulletin board. The soundness bound for the single terminal-anchor protocol does
not carry over.

\subsection{Protocol for AI training verification}\label{sec:verifying}

The protocol produces three proof types, described below as phases \ref{offline}--\ref{ex-ante-attestation}.
\Cref{fig:overview-verif-protocol} overviews the commitment, training, and challenge
flow shared by \ref{offline} and \ref{online}; \ref{ex-ante-attestation} adds
pre-committed policy claims and running counters to proved steps.
The \emph{genesis proof} runs once at
initialisation and binds the committed procedure to physical execution.
The
\emph{in-training step proof} fires many times across a run and verifies sampled
steps against the published hash chain and the network anchor. The \emph{ex-ante
attestation} binds policy-relevant claims (compute, regime, data-content) at
initialisation and enforces them as running invariants.
All three share the
commitments of \Cref{pre-commit,constant-hashing,network-tap} and the precompile
catalogue (\Cref{precompiles}).
We define and analyse our protocol in Appendix~\ref{appendix:formalization}:
an overview of the computational complexity-theoretic framework is given in~\Cref{sec:formal-background};
the target relation is defined in~\Cref{sec:defs-proto-formal};
the desired properties and their security analysis are given in~\Cref{sec:protocol-sec-stmts}.

\begin{enumerate}[label=\textbf{P.~\arabic*},leftmargin=*]
  \item\label{offline} {\bf (Genesis proof)} The genesis proof binds the committed training procedure to actual GPU execution,
  run once before the hash chain begins. Without it, $h_{\text{commit}}$ is a signed
  declaration of intent with no tie to physical work.
  The \textbf{genesis protocol} proceeds as follows.
  The auditor supplies \emph{random sample indices} $\{i_1,\ldots,i_b\}$ into the committed
  dataset. The trainer (i) constructs $B=(\text{data}[i_1],\ldots,\text{data}[i_b])$
  in the committed canonical order; (ii) executes one full training step from $W_0$ on
  $B$ (forward, backward, optimiser update); (iii) produces a zkVM proof that each
  sample is a committed element of $R(\text{dataset})$ at its claimed index (one
  Merkle path per sample via precompile~3), the step executes faithfully per
  \texttt{arch\_spec}, and the resulting $W_1$ is the first link of the public hash
  chain. \emph{For soundness,} the indices are chosen randomly \emph{after}
  $h_{\text{commit}}$ is published. Precomputing the challenged step would therefore
  require anticipating the sampled batch among combinatorially many candidates, which is
  infeasible given the challenge entropy. Merkle-path checks prevent batch substitution.
  Finally, full-step execution exercises the entire training pipeline, so the proof fails
  if a divergence between the claimed procedure and the executed code affects the
  challenged step's outputs. The genesis traffic tag
  $h_{\mathrm{gen}}=F_K(\mathrm{traffic}_{\mathrm{gen}})$, where $F_K$ is a keyed
  hash under the anchor's run secret $K$
  (\Cref{appendix:formalization}), additionally binds the genesis step to
  physically observed inter-node traffic, though no component attests the
  executing device itself. The \emph{cost} is about that of proving one 2{,}048-token microbatch, when the genesis batch is no larger (\Cref{appx:proof-cost:in-training}).
  %
  Genesis establishes faithful execution at step~0; it says nothing about
  learning-theoretic properties or aggregate compute---those are the subject of the
  ex-ante attestation (\Cref{ex-ante-attestation}).
  
  \item\label{online}  {\bf (In-training step proof)} Fired many times across a run, live or post-hoc against archived snapshots, this
phase gives the auditor assurance that any sampled step is consistent with the hash
chain and network anchor.
%
  The \textbf{step protocol} proceeds as follows.
The auditor selects a step $t\in\{1,\ldots,T\}$\footnote{Adaptive, root-dependent
forensic challenges are operationally permitted, but the security-budget
challenges analysed in \Cref{sec:sampling-guarantees} are those drawn uniformly
from the frozen commitment stream via the commit-then-reveal audit seed; adaptive
challenges are supplementary and claim no additional soundness.}, a tensor identifier (layer $\ell$
output, gradient at node $i$, an all-reduce payload), and $k$ random entry indices.
For tensors committed only at challenge time (the lazy FFN intermediates of
\Cref{appx:proof-cost:commitments}) the challenge is \emph{nested}: the step and
tensor identifiers are revealed first, the trainer publishes the lazy roots, and
only then are the $k$ entry indices drawn from fresh, domain-separated randomness
(\Cref{sec:sampling-guarantees}).
The trainer supplies the following hints: the requested entries and their Merkle paths
against the committed $R(\cdot)$ for step~$t$. For wire-bound tensors, the trainer also
supplies the SHA-256 Merkle path against $R(\mathtt{msg}_*)$. The proof checker dispatches
Merkle-path verification (precompile~3), operation verification (precompiles 1, 2, and 4--7), and,
where relevant, anchor-consistency verification (precompile~8). A deviation affecting a
fraction $f$ of a challenged tensor's entries is detected with probability at least
$1-(1-f)^{k}$ (\Cref{sec:sampling-guarantees}). At the GEMM default $k=4{,}605$, any
deviation with $f \geq 1\%$ escapes with probability at most $10^{-20}$.
Fused-attention outputs use the reduced budget $k_{\mathrm{attn}}=500$ and therefore
have correspondingly weaker per-object detection.
%
\emph{Soundness} across a run composes via two chaining relations, ensuring {\bf cross-step binding}.
Within a step, layer boundaries link by
$R(\text{layer}_\ell\text{\_out})=R(\text{layer}_{\ell+1}\text{\_in})$. Across steps,
the step proof verifies, from openings bound to the committed roots, that the
challenged components of $(W_{t+1},\text{optimiser state}_{t+1})$ are the declared
optimiser's update of $(W_t,\text{optimiser state}_t)$ given $\mathtt{grad}_t$,
binding $R(W_{t+1})$ to $R(W_t)$ and $R(\mathtt{grad}_t)$.
Inductively, this forms a directed chain from $R(W_0)$ to $R(W_T)$, verified at
the sampled steps.
%
Sampled-step and per-layer proofs aggregate via \emph{recursive STARK composition} into a
single ${\sim}200$\,KB artefact. 

  \item\label{ex-ante-attestation} {\bf (Ex-ante attestation)} The ex-ante attestation binds policy-relevant claims at initialisation and enforces
them as running invariants throughout the online phase. Categories include: \emph{compute}
(total FLOPs, tokens, GPU-hours), \emph{procedure} (regime, absence of a particular
phase such as RL), and \emph{data-content} (fraction flagged by a public classifier
below a threshold). 
Under an ex-ante attestation, a trainer cannot pass a \emph{challenged} step whose
running counter violates a committed bound without breaking the \emph{soundness}
property of the proof system. Run-wide assurance additionally inherits the
step-sampling guarantees of \Cref{sec:sampling-guarantees} and the structural
detections of \Cref{appx:proof-cost:defense-depth}. Universal per-step coverage is
instead the V5-IVC upgrade path discussed in \Cref{appendix:formalization}. The
proof-system, sampling, and structural mechanisms together turn the training record into
a governance-auditable artefact rather than a mere audit trail.
This phase involves a \textbf{commitment extension.}
That is, the trainer publishes a structured \texttt{ex\_ante\_claims} containing committed
bounds and references (e.g.\ \texttt{max\_total\_flops}, \texttt{max\_steps},
\texttt{training\_regime}, \texttt{data\_filter\_hash}). The genesis commitment
extends to
$h_{\text{commit}}=H(H(\texttt{arch\_spec}) \,\|\, R(\text{dataset}) \,\|\, R(W_0)
\,\|\, H(\mathtt{ex\_ante\_claims}))$.
Bound values are public; only \texttt{arch\_spec} stays private. Opening can be
upfront or on-demand via selective disclosure against $H(\mathtt{ex\_ante\_claims})$.
%
Each step proof from~\ref{online} performs \textbf{running enforcement},
by carrying an additional public running counter, e.g.\
$F_{\mathrm{cum}}(t)$ for cumulative FLOPs, computed in-circuit as
$F_{\mathrm{cum}}(t)=F_{\mathrm{cum}}(t-1)+F_{\text{step}}(\texttt{arch\_spec},t)$ with
the constraint $F_{\mathrm{cum}}(t)\le\mathtt{max\_total\_flops}$\footnote{The same pattern applies to step counts, per-regime
tallies, and any monotone declared quantity.}. 
More generally, the attestations involve a public \emph{recursive} functional $g$ over committed state
and prove
$g(\cdot)=y$ or $g(\cdot)\le y$.
The committed state varies with the
sub-type: $\texttt{arch\_spec}$ and chain length for compute claims; $\texttt{arch\_spec}$
structure for procedure claims; sampled dataset entries plus a public classifier for
data-content claims\footnote{Data-content attestations combine Merkle-path openings into
$R(\text{dataset})$ with in-circuit evaluation of the classifier and a concentration bound
(Hoeffding) over the sampled fraction.}.
The recursive structure, together with the soundness of the recursive STARK
composition, ensures that an aggregated proof is accepted only if every
\emph{proved} step satisfies the enforcement; steps that are never challenged are
covered by the sampling and structural arguments above, not by proof soundness.
The per-step overhead for this enforcement is one addition and one
comparison---negligible vs.\ the MAC budget.
We discuss \emph{computation under-declaration} and \emph{ex-post variants} in~\Cref{sec:protocol-sec-stmts}.

\end{enumerate}

\section{Probabilistic guarantees of sampling-based verification}\label{sec:sampling-guarantees}

The protocol of \Cref{sec:verifying} keeps verification affordable through
\gls{spot-checking}{spot-checking}. It samples the computation instead of checking
all of it. Sampling occurs at two levels.

\emph{Macroscopically}, the auditor samples the $T$ training steps. Within each
sampled step, it challenges one or more operations. \emph{Microscopically}, the
proof recomputes only $k$ sampled entries of a committed result. For example, it
may sample $k$ entries of a GEMM output matrix. Both sample sizes are tunable.

Sampling makes detection uncertain. Its probability depends on the sample sizes
and the amount of tampering. This section answers three practical questions:
\begin{itemize}
  \item How many checks buy how much confidence (\Cref{tab:sample-count})?
  \item How does the spread of cheating affect its detection cost
  (\Cref{tab:step-sampling})?
  \item What can no affordable sampling budget catch?
\end{itemize}
Throughout, the trainer is an adversary. It may use any strategy that tampers with
at least the stated share of computation. Each bound covers every such strategy,
not only accidental errors. Every sample is drawn at random only \emph{after} the
trainer's records are frozen. The formal statement is the soundness bound of
\Cref{sec:protocol-sec-stmts}.

\paragraph{The basic rule.}
Suppose that $M \geq fN$ of $N$ items have been tampered with. The auditor checks
$k$ distinct items drawn uniformly at random. The audit misses the tampering only
if every checked item is clean. This happens with probability
\begin{equation}\label{eq:entry-miss}
  P_{\mathrm{miss}} \;=\; \prod_{i=0}^{k-1}\frac{N-M-i}{N-i} \;\le\; (1-f)^{k},
\end{equation}
because each factor is at most $1-f$. To reduce the miss probability below a
target $\tau$, it is therefore enough to check
\begin{equation}\label{eq:entry-budget}
  k \;\geq\; \frac{\ln(1/\tau)}{-\ln(1-f)}
\end{equation}
items. This bound holds for any total $N$.

Two consequences drive the results below. \emph{Confidence is cheap.} Each
tenfold reduction in miss probability costs a fixed number of extra checks:
$\ln(10)/(-\ln(1-f))$. At $f=1\%$, this is about $229$ checks.
\emph{Rare tampering is expensive.} The number of checks grows roughly as $1/f$.
Thus, tampering that is ten times rarer costs ten times more to catch.

Our tables use the right-hand side of \Cref{eq:entry-miss}. This bound does not
depend on $N$. The exact value is smaller. The difference becomes noticeable only
when the sample covers a sizeable share of all items. Checking \emph{every} item
is a different regime. It makes detection certain, at full cost.

\paragraph{Microscopic-level guarantee: entry sampling within a challenged operation.}
Here, the items are the entries of one committed output. A GEMM output, for
example, has $m \cdot n$ entries. The fraction $f$ is the share that differ from
what the declared operation produces on the committed inputs.

The proof recomputes each sampled entry and checks it against the committed Merkle
roots. For a GEMM output, each sample costs one dot product.
\Cref{tab:sample-count} gives the required sample counts. The GEMM default is
$k=4{,}605$. For a $1\%$ alteration, its miss probability is below $10^{-20}$.

Fused-attention outputs cost more to sample. They use
$k_{\mathrm{attn}}=500$
(\Cref{appx:proof-cost:commitments,appx:proof-cost:per-layer-budget}). For the same
alteration, their miss probability is ${\approx}7\times10^{-3}$. Each output kind
must therefore be read with its own budget.

\begin{table}[h!]
  \centering
  \small
  \begin{tabular}{lccc}
    \toprule
    Target miss probability $\tau$ & $f = 10\%$ & $f = 1\%$ & $f = 0.1\%$ \\
    \midrule
    $10^{-3}$ & $k = 66$ & $k = 688$ & $k = 6{,}905$ \\
    $10^{-6}$ & $k = 132$ & $k = 1{,}375$ & $k = 13{,}809$ \\
    $10^{-10}$ & $k = 219$ & $k = 2{,}292$ & $k = 23{,}015$ \\
    $10^{-20}$ & $k = 438$ & $k = 4{,}583$ & $k = 46{,}029$ \\
    $2^{-128}$ & $k = 843$ & $k = 8{,}828$ & $k = 88{,}679$ \\
    \bottomrule
  \end{tabular}
  \caption{Samples needed within one checked output. Each entry is the smallest
  $k$ satisfying \Cref{eq:entry-budget}. Thus, an alteration of at least a
  fraction $f$ escapes with probability at most $\tau$. The last row matches the
  nominal security level of proof systems. Down each column, another tenfold gain
  in confidence adds a fixed number of samples. The increases are about $22$,
  $229$ and $2{,}301$, respectively. Across a row, an alteration ten times rarer
  costs ten times more to catch. The GEMM default $k=4{,}605$ is just above the
  $4{,}583$ needed for $(\tau,f)=(10^{-20},1\%)$. Even the largest entry is a small
  sample. The $88{,}679$ entries are ${\approx}0.26\%$ of a
  $2{,}048\times16{,}384$ GEMM output. This small share makes the check cheap. It
  also lets a much smaller alteration slip through; see ``What sampling cannot
  catch'' below. Each $\tau$ applies to one checked output. Guarantees over many
  outputs require a union bound.}
  \label{tab:sample-count}
\end{table}

\paragraph{Macroscopic-level guarantee: step sampling across the run.}
At 405B scale, one microbatch contains about $10^{3}$ operations that a challenge
can check. This estimate uses 126 layers times roughly eight operations each
(\Cref{appx:proof-cost:in-training}). A training step processes many microbatches.
Thus, its number $N_{\mathrm{ops}}$ of checkable operations is at least $10^{3}$.

A challenge catches the cheating only if it checks a tampered operation. Suppose
that operation has at least $1\%$ of its entries altered. Checking it then exposes
the tampering almost surely. At the default $k$, a GEMM output has miss probability
below $10^{-20}$.

Under this condition, the key quantity is the share of all checkable operations
that were tampered with. This share depends on how the trainer spreads the
cheating. Two patterns bracket the possibilities. Let
$\alpha_{\mathrm{step}}$ be the fraction of dishonest steps.
\begin{itemize}
  \item \textbf{Systematic cheating.} A dishonest step was not executed as
  declared. The trainer may have skipped computation or substituted weights or
  data. Essentially all its operations are wrong. Each has at least $1\%$ of its
  entries altered. The tampered share is $\alpha_{\mathrm{step}}$.
  \item \textbf{Concentrated cheating.} A dishonest step is honest except for one
  operation. That operation has at least $1\%$ of its entries altered. The
  tampered share falls to $\alpha_{\mathrm{step}}/N_{\mathrm{ops}}$. It is at
  least a thousand times smaller.
\end{itemize}
The number of checks grows inversely with the tampered share
(\Cref{eq:entry-budget}). Concentrated cheating is therefore at least
$N_{\mathrm{ops}}$ times more expensive to catch (\Cref{tab:step-sampling}).

Checking larger units does not avoid this cost. A challenge that checks more
operations requires proportionally more proving work. Thus, every challenge size
buys roughly the same detection per unit of work.

\begin{table}[h!]
  \centering
  \small
  \begin{tabular}{lrr}
    \toprule
    & \multicolumn{2}{c}{Challenges needed to catch it with probability} \\
    \cmidrule(lr){2-3}
    How the trainer cheats & $99\%$ & $99.99\%$ \\
    \midrule
    Systematic, on $1\%$ of steps & $459$ & $917$ \\
    Systematic, on $0.1\%$ of steps & $4{,}603$ & $9{,}206$ \\
    Concentrated, on $1\%$ of steps & $\geq 4.6\times10^{5}$ & $\geq 9.2\times10^{5}$ \\
    Concentrated, on $0.1\%$ of steps & $\geq 4.6\times10^{6}$ & $\geq 9.2\times10^{6}$ \\
    \bottomrule
  \end{tabular}
  \caption{Single-operation challenges needed to check at least one tampered
  operation. Each GEMM challenge recomputes $k=4{,}605$ sampled output entries
  inside the proof. Each entry is a dot product whose length is the GEMM's inner
  dimension. See \Cref{appx:proof-cost:in-training} for proving costs. The challenge count
  $k_{\mathrm{ch}}$ is the smallest integer satisfying
  \Cref{eq:entry-budget}. Here, the tampered operation share replaces $f$. This
  share is $\alpha_{\mathrm{step}}$ for systematic cheating. It is
  $\alpha_{\mathrm{step}}/N_{\mathrm{ops}}$ for concentrated cheating. The
  concentrated rows use $N_{\mathrm{ops}}=10^{3}$, or one microbatch per step.
  They are therefore lower bounds. Every row assumes that checking a tampered GEMM
  output fires: $f \geq 1\%$ with the default $k$. The counts do not depend on run
  length while they remain far below the total number of checkable operations.}
  \label{tab:step-sampling}
\end{table}

\paragraph{The general bound.}
Let $s$ lower-bound the probability that one challenge lands on a tampered
operation. Let $p$ lower-bound the probability that checking that operation
exposes it. If at least fraction $f$ of its entries are altered, then
$p \geq 1-(1-f)^{k}$ for entry budget $k$.

The audit makes $k_{\mathrm{ch}}$ challenges without replacement.
Single-operation challenges draw from the run's checkable operations. Full-step
challenges draw from its steps. Each challenge uses fresh, independent
entry-sampling randomness. The audit then misses the cheating with probability
\begin{equation}\label{eq:endtoend}
  P_{\mathrm{miss}} \;\le\; \bigl(1-s\,p\bigr)^{k_{\mathrm{ch}}}.
\end{equation}
Because sampling is without replacement, the challenged tampered-operation count
is hypergeometric. The map
$x \mapsto (1-p)^{x}$ is convex. Therefore, the expected miss probability is at
most its value under sampling with replacement~\citep[Thm.~4]{hoeffding1963}.
That value is the right-hand side above.

For single-operation challenges, $s \approx \alpha_{\mathrm{step}}$ against
systematic cheating. Against concentrated cheating,
$s \approx \alpha_{\mathrm{step}}/N_{\mathrm{ops}}$. A full-step challenge checks
every operation in a step. It has $s = \alpha_{\mathrm{step}}$ in both cases.
However, it needs about
$N_{\mathrm{ops}}$ times more proving work.

\Cref{eq:endtoend} is a simplified, idealised counterpart of the formal bound in
\Cref{sec:protocol-sec-stmts}. There, branch M3 is the step-sampling term. Its
$k_{\mathrm{step}} \le T$ counts sampled steps. It corresponds to the
$\alpha_{\mathrm{step}}$ factor of $s$. The $1/N_{\mathrm{ops}}$ factor of $s$
and the factor $p$ correspond to branch M2. The formal sampler checks layers and
entries rather than single operations. Aligning the statements exactly remains
part of the formalisation roadmap.

The remaining branches cover commitment binding, proof-system soundness and
commit-then-reveal. Auxiliary terms cover pseudorandom functions (PRFs) and
signatures. These cryptographic terms are designed to sit below the sampling
terms. Their concrete values are tracked in \Cref{appendix:formalization}.

\emph{Example.} Consider a run with $T=10^{5}$ steps. The trainer cheats
systematically on $1\%$ of them. Each operation has at least $1\%$ of its entries
altered. Then $10^{3}$ single-operation challenges into GEMM outputs catch the
cheating with probability above $99.99\%$. The miss probability is at most
$0.99^{1000} \approx 4.3\times10^{-5}$. Suppose the same trainer instead alters
one operation in each dishonest step. The same $10^{3}$ challenges then catch the
cheating with probability at most about $1\%$.

\paragraph{What sampling cannot catch.}
The basic rule has a flip side. The smallest tampering is the hardest to find. At
the extreme, the audit does not find it at all.
\begin{itemize}
  \item \emph{A single altered entry} in one operation is caught with probability
  $k/(mn)$ when that operation is checked. At the default $k$, this is about
  $1.4\times10^{-4}$ for a $2{,}048\times16{,}384$ GEMM output.
  \item \emph{A single dishonest step} is caught with probability about
  $k_{\mathrm{step}}/T$ if it is systematic. Here, $k_{\mathrm{step}}$ is the
  number of sampled steps. If the cheating is concentrated, the probability is
  about $k_{\mathrm{step}}/(T N_{\mathrm{ops}})$. This expression assumes one
  single-operation challenge per sampled step. Neither case gives cryptographic
  certainty (\Cref{sec:protocol-sec-stmts}). Reliable detection would require
  checking essentially every step.
\end{itemize}
Sampling therefore protects against \emph{widespread} cheating across many steps
and many operations within them. It does not protect against an adversary served
by a handful of altered values. Targeted model manipulation is one example.

The frauds relevant to compute-threshold attestation plausibly need widespread
footprints to pay off. Examples include under-declared compute, substituted models
or datasets, and skipped work. Whether a given fraud \emph{must} be widespread is
open (\opref{algebraic-reduction}). It may instead have a concentrated variant.
The bounds above do not settle this question. We consider this limit as important
to state as the headline numbers.

Two mechanisms narrow the gap without closing it. First, consider a checked entry
that is wire-bound. A mismatch between the trainer's commitment and the network
anchor's record is caught deterministically. This requires the wire-to-tensor
reconciliation of \opref{wire-to-tensor}. Second, the committed hash chain prevents
after-the-fact changes to the record. Both mechanisms act only on checked objects
(\Cref{appx:proof-cost:defense-depth}). A principled solution would check every
value, but less precisely. This motivates the hybrid protocols of
\Cref{sec:hybrid} and open problems \opref{algebraic-reduction} and \opref{hybrid}.

\paragraph{A limit that better verification does not remove.}
Larger sampling budgets shrink the gaps above. Full coverage would remove them.
However, this architecture aims to certify that a run followed a declared recipe
at a declared cost. It therefore relies on \emph{training compute as a proxy for
capability}. At least three cases escape this proxy.
\begin{itemize}
  \item \emph{Undeclared preparation compute.} A trainer can execute the declared
  run faithfully on inputs prepared with undeclared compute. The cheapest such
  input is a dataset. For example, the trainer may distil it from an undeclared
  model. The verified run can thereby reach a capability that the declared budget
  would not buy. No check in this paper fires because the executed computation contains
  no falsehood.

  Whether this counts as under-declaration depends on the accounting boundary.
  Article~51(2) states that the threshold covers \emph{cumulative} training
  compute~\citep{euaiact_art51}. Recital~111 of the same Regulation states that
  this cumulative amount includes activities intended to enhance capabilities
  before deployment. It names ``such as pre-training, synthetic data generation
  and fine-tuning''. What remains
  unaddressed is the \emph{attribution} of third-party compute to the model under
  assessment. This includes compute spent by an upstream synthetic-data generator
  or an undeclared teacher. 
  \item \emph{Compute declared by someone else.} A trainer distilling from a
  publicly released model under-declares nothing. Yet its run buys more than its
  own FLOPs would.
  \item \emph{A moving compute--capability map.} Even with nothing undeclared, a
  recipe may genuinely cost what it declares. It may still produce a more capable
  model than the threshold's authors assumed.
\end{itemize}
The last two cases limit compute as a governance instrument, not this protocol.
This architecture binds a training \emph{run}, not the capability of the resulting
model.

\paragraph{Preconditions: fresh, unpredictable challenges.}
Every number above assumes two conditions:
\begin{itemize}
  \item[(i)] Random choices are made \emph{after} the trainer's commitments are
  frozen.
  \item[(ii)] The trainer cannot predict or influence those choices.
\end{itemize}
The protocol enforces (i) through
\gls{commit-then-reveal}{commit-then-reveal}. It opens the audit seed, or issues
the challenge, only after freezing the commitment stream. An auditor-signed
terminal anchor performs this freeze (\Cref{sec:protocol-sec-stmts}).

Some tensor roots are committed only at challenge time. These include the lazy
FFN intermediates in \Cref{appx:proof-cost:commitments}. They need an additional,
nested order. Their roots must be published before drawing the entries that check
them.

Requirement (ii) depends on the source of audit randomness (\Cref{sec:proof-options}). Suppose
Fiat--Shamir derives challenges from the public transcript instead of the hidden
commit-then-reveal seed. The trainer can then recompute the challenges. It may try
up to $Q$ candidate transcripts and keep one whose challenges miss the tampering.
Every miss probability above is then multiplied by up to $Q$.

At the entry level, compensation costs
$\ln(Q)/(-\ln(1-f))$ extra samples. This is about $2{,}760$ samples for $Q=2^{40}$
at $f=1\%$. Step-level budgets grow in the same way. Removing this exposure
requires randomness that the trainer cannot query offline. It can come from a live
auditor or a public randomness beacon.

Finally, sampled positions need domain-separated samplers. Guarantees over many
checked objects require a union bound across all objects. The per-object bound
alone is insufficient. See the operational caveats in
\Cref{appendix:formalization}.

\section{Towards hybrid verification protocols}\label{sec:hybrid}

The results from~\Cref{sec:sampling-guarantees} above bracket a design gap. At one end, sampling-based verification is
affordable at frontier scale. However, it provides detection-grade assurance with
$k/(mn)$-type floors. A coefficient that no check reads can absorb an arbitrary
deviation. At the other end, a cryptographic proof could bind every coefficient of
every step bit-exactly. This would remove the floors. Yet proving every step is
currently too expensive (\Cref{appx:proof-cost:flows}).

\gls{hybrid-protocol}{\emph{Hybrid verification protocols}} aim between these
endpoints. The authors are investigating them in companion work. These protocols
change \emph{what is bound per coefficient}, not \emph{which coefficients are
bound}. The construction has two tiers:
\begin{itemize}
  \item A \textbf{strong sampling-based component} uses sparse, unpredictable,
  bit-exact spot-checks, and/or exact anchors. An example anchor is a checkpoint
  bound to the network anchor. The anchors occur at randomly drawn, unpredictable
  steps, as above. For a class defined by a stated share of tampering, the
  spot-check budget depends on target confidence but not run length.
  \item A \textbf{weak but complete per-step component} verifies every step. It
  uses a declared tolerance rather than bit-exact checks. This gives full coverage
  at relaxed precision. It is expected to cost substantially less to prove than
  bit-exact binding.
\end{itemize}
The hypothesis is that \emph{relaxing precision degrades gracefully, while
relaxing coverage can fail abruptly}. An unread coefficient permits an arbitrary
deviation. By contrast, a coefficient bound at tolerance $\eta$ can deviate by at
most $\eta$ per step. Even adversarially correlated within-tolerance deviations
therefore create a drift-composition question over the training trajectory. They
do not create an unbounded hole.

The main candidate directions are:
\begin{itemize}
  \item \textbf{Tolerance-band commitments with composition guarantees.} Prove
  every step within width $\eta$. Add theorems that bound the end-of-run effect of
  adversarial within-tolerance deviations on the final weights.
  \item \textbf{Cheaper-to-prove surrogate algorithms.} Design algorithms that
  approximate the exact ones. The aim is to greatly reduce the cost of
  zero-knowledge proofs, possibly of approximate facts. Error bounds might be
  proved formally, and perhaps in zero-knowledge as well.
  \item \textbf{Sampled bit-exact backstops.} Use unpredictable bit-exact checks to
  catch a splice between the executed computation and the committed tolerance-band
  trace. The backstop budget sets the miss probability. For a fixed share of
  tampering, this probability is independent of run length. A single-step splice remains a
  $k_{\mathrm{step}}/T$-type event. The drift-composition analysis must turn every
  \emph{consequential} splice into a fixed-fraction inconsistency.
  \item \textbf{Proof-friendly exact dialects for nonlinear operations.} Replace
  kernels that are hard to express as arithmetic constraints with canonical
  deterministic surrogates. Declare them exhaustively in \texttt{arch\_spec}.
  Such operations can then use cheap exact binding. The tolerance machinery
  remains for the dominant linear algebra.
  \item \textbf{Beacon-derived challenge randomness}
  (\Cref{sec:proof-options}). This prevents offline grinding of audit positions,
  not attacks on the proof system's internal challenges. Publicly checkable audits
  also need the registration and commit-before-release rules of \opref{beacons}.
\end{itemize}
These directions remain ongoing work. One point already frames the comparison. A
hybrid protocol verifies a \emph{different object}: tolerance-band execution with
exact backstops. The checks in this paper instead sample bit-exact execution. Thus,
hybrid guarantees need their own security analysis (\opref{hybrid}). The
ingredient-level questions are split across
\opref{float-native}--\opref{field-hardware}.

\section{Weight-smuggling attacks and global training budgets}\label{sec:weight-smuggling}

\emph{Weight-smuggling attacks} aim to import the results of other training without
charging that work against a run's compute cap. Two routes illustrate the problem:
\begin{itemize}
  \item \textbf{Chaining.} Each new run resumes from an earlier run's weights but
  receives a fresh compute allowance.
  \item \textbf{Aggregation.} Several runs stay below their individual caps, then
  their weights are merged.
\end{itemize}
Both routes can combine contributions whose total training cost exceeds any one
run's allowance. Weight averaging already combines compatible fine-tuned
models~\citep{wortsman2022soups,maiti2026souper}. This does not establish a practical
bypass of frontier training caps.

An undeclared weight splice violates the recipe, but sampling may miss it.
Transfers allowed by the recipe pose a different problem: every declared operation
may be correct. Even complete execution checks would then leave the accounting
gap open. Recomputing the prescribed initial weights and checking their full
commitment closes direct substitution at the start. However, crafted data might
steer a freshly initialised model towards weights trained elsewhere. Dataset
distillation compresses much of a dataset's training effect into a few synthetic
examples in studied settings~\citep{wang2018datasetdistillation}. Whether such
methods can efficiently recover chosen weights under strict frontier-training
recipes remains open (\opref{weight-smuggling}).

These attacks make \emph{global training budgets} a more plausible basis for
limiting total training compute than \emph{per-run caps}. Here, ``global'' means one
total across all training runs covered by the regime, regardless of model or
training history. When all contributing work is counted, starting a new run or
merging models cannot create additional allowance. Each training operation counts
towards the same budget, whether its result is reused, merged or discarded. This
reduces reliance on tracing each model's training history.

The advantage depends on coverage. The regime must specify the actors, hardware,
time interval and activities it covers. Imported weights, synthetic-data
preparation and training outside that boundary need explicit treatment. Metering
covered hardware alone cannot bound training elsewhere (\opref{compute-accounting}).
Setting the budget and allocating it among trainers also remain policy choices.

\section{Conclusion}\label{sec:conclusion}
We argued that frontier dense pre-training admits a tractable verification architecture
combining three complementary anchors (a committed training specification, inter-node
network observations, and on-the-fly Merkle commitments of intermediate computation)
verified through a succinct proof layer with native BF16/FP32 precompiles. The
architecture sidesteps the scale and arithmetic limits of the existing ZK-ML paradigm by
verifying bit-exact conformance to the declared BF16/FP32 execution semantics rather than
a fixed-point approximation of the model.
The proof layer can use a generic zkVM (our reference design), custom circuits,
or direct interactive audits. Coverage, challenge randomness, privacy, and the
proving schedule are separate design choices (\Cref{sec:proof-options}). The sampling-based variants carry the
explicit detection-probability guarantees of \Cref{sec:sampling-guarantees}: at
negligible challenge cost, they catch bulk and systematic cheating with high probability.

However, minimal targeted deviations limit this assurance and
motivate the hybrid protocols under investigation (\Cref{sec:hybrid}). We list open problems that seem relevant to us in \Cref{app:open-problems}.
As explained in \Cref{sec:weight-smuggling}, it also seems likely that malicious provers can circumvent per-training-run FLOPs caps, by combining several training runs, either sequentially or in parallel, in ways that are hard or impossible to detect, especially with purely sampling-based procedures. Thus, \emph{we argue that verification regimes with global training budgets for each actor are more robust}. Finally, given the time scales, deploying simpler solutions for potential "emergency deals", relying on e.g. inference-only verification, auditing (with humans or LLMs), seems worthwhile as well.


\bibliographystyle{plain}
\bibliography{sample}


\appendix
\crefalias{section}{appendix}
\crefalias{subsection}{appendix}
\crefalias{subsubsection}{appendix}

\section{Disclosure of AI use}\label{app:ai-use}
In this work, we used LLMs (Claude and GPT) for finding references, drafting proofs and parts of the text, proofreading, help with implementation.

\section{Open problems}\label{app:open-problems}

This appendix lists the research and engineering problems that remain after
specifying the architecture in \Cref{sec:proposed-solution}. It complements and expands the public
roadmap at \texttt{gpaipolicylab.org/verification}.

Each problem is a research question with a named target. It can be attacked independently.
Together, they invite contributions from cryptography, systems, hardware, and
machine learning.

\subsection{Outlook} \label{sec:future-work}

These problems separate the architecture from a fielded verification primitive.
Three items form the critical path:
a deterministic-by-construction attention kernel (\opref{det-attention}), a
canonical wire-to-tensor mapping for NCCL (\opref{wire-to-tensor}), and a
zero-knowledge proof of backpropagation at non-trivial scale (\opref{zk-backprop}).
The first two are engineering problems with measurable success criteria.
The third is the roadmap's highest-risk milestone. A fundamental obstacle would
require rethinking the protocol, not merely refining its implementation.

A second group asks what a verified run establishes. These questions are not
about cost or engineering. \opref{reachability} asks which final weights a
protocol-conformant run can reach when the trainer authors the inputs.
\opref{compute-accounting} asks what can be certified about the \emph{quantity}
of compute, as distinct from following a recipe. These questions matter because
the architecture's soundness guarantees concern a computation. The motivating
governance frameworks instead draw conclusions about a model. No amount of
cryptography supplies the step between those claims.

The estimated 36-month deployment envelope depends on coordinated progress
across these fronts.

On the proof side, one number dominates: the MAC precompile's constraint count
(\opref{mac-constraints}). An algebraic reduction (\opref{algebraic-reduction})
is the structural alternative. Two problems dominate the systems side: the attention-backward
determinism tax (\opref{det-attention}) and portability beyond one
vendor's silicon (\opref{precompile-portability}).

The network anchor's independence rests on
\opref{tap-hardware}--\opref{intranode-witness}. The limits in \opref{sdc}
constrain what any full-coverage tier can accept. The other groups extend the
base protocol (\opref{moe}--\opref{multisite}) and connect it to real training
code (\opref{converter}). They also place deployments within the choices in
\Cref{sec:proof-options} (\opref{beacons}--\opref{transferability}). Finally,
they draw more from the anchor than the base protocol consumes
(\opref{anchor-exploitation}--\opref{compute-accounting})
and price the underlying algebra (\opref{float-native}--\opref{codesign}).

\subsection{Proof foundations}

\opitem{algebraic-reduction}{Algebraic reductions for floating-point GEMM verification.}
Per-element sampling is the current strategy for BF16/FP32 GEMMs. Exact
Freivalds and sumcheck fail because floating-point addition is non-associative.
\emph{Question:} is per-element sampling provably optimal for floating-point
matrix-multiplication verification? Or can an algebraic shortcut exploit
structure, such as the accumulation-tree topology declared in the precompile?
Sub-problems include:
\begin{itemize}
  \item \textbf{Approximate reductions.} Recent work on sumcheck with bounded
  approximation error~\citep{bitan2025approxsumcheck} suggests another route.
  An approximate Freivalds test might absorb rounding differences while its
  soundness degrades gracefully. The remaining obstacles are polynomial
  commitments that preserve soundness under approximation and a tight error
  bound $\delta$ that separates rounding noise from adversarial deviation.

  \item \textbf{Separate the two relaxation axes.} One axis relaxes the
  \emph{precision} that binds each output entry. Examples include tolerance
  bands and approximate sumcheck. The other relaxes the \emph{coverage} of
  entries bound at all. Examples include sampling, sketches, and aggregate
  norms. The detection floors in \Cref{sec:sampling-guarantees} are inherent to
  per-entry coverage relaxation. An entry that no check reads can absorb an
  arbitrary deviation. Precision relaxation instead degrades gracefully within
  each step. Its aggregate behaviour over a run is the drift-composition
  question in \opref{hybrid}.

  \item \textbf{Is the deviation fraction the right parameter?} Every bound in
  \Cref{sec:sampling-guarantees} applies to a deviation class. The current class
  requires changes to at least a fraction $f$ of one challenged output's entries.
  This is a \emph{syntactic} measure chosen for tractability. It was not chosen
  because an adversary's payoff is a function of it. The guarantee matters
  only if profitable deviations fall inside this class. The sensitivity analysis
  must establish that claim or find a natural attack outside the class.

  This analysis bounds the smallest $f$ that a rational adversary can exploit,
  thereby calibrating the sample budgets used in \Cref{appendix:proof-cost}.
  This parameterisation is especially likely to be wrong because a
  deviation can alter few entries yet have a large effect.
\end{itemize}
An algebraic reduction could cut the dominant proof cost by an order of magnitude or more.
The reduction's own constraint cost determines the size of that saving.

\opitem{mac-constraints}{Constraint minimisation for the MAC precompile.}
The BF16/FP32 MAC precompile costs an estimated ${\sim}90$ constraints per
operation over the Baby Bear field. It accounts for roughly 98\% of the proof
budget at frontier scale. \emph{Question:} can custom gates, lookup arguments,
or alternative field representations reduce the cost to 30--50 constraints per
operation? Sub-problems include:
\begin{itemize}
  \item \textbf{Measure the baseline first.} The ${\sim}90$ estimate comes from
  an IEEE~754 integer decomposition spanning 70--110 constraints. The model uses
  a deliberately wider 50--150 implementation envelope
  (\Cref{appx:proof-cost:model}). A factor-two error in this baseline swamps a
  factor-two optimisation.

  \item \textbf{Pin the cost model.} A ``constraint'' is not a portable unit.
  Counts are not comparable across proof-system families, such as AIR/STARK,
  R1CS, and Plonkish, or across degree bounds. Thus, 30--50 is meaningful only
  for a declared family and degree budget. Cross-system claims also need a
  stated conversion.

  \item \textbf{Attack the rounding chain specifically.} Mantissa multiplication
  and exponent addition are algebraic and cheap. Cost instead concentrates in
  RNE rounding: exponent alignment and the guard, round, and sticky bits.

  \item \textbf{Amortise over the dot product.} MACs occur in length-$d$
  accumulation chains, not in isolation. It remains open whether a length-$d$
  chain relation can cost less than $d$ independent MAC relations. Any such relation
  must still pin the accumulation order declared by the precompile. This is the
  structural question in \opref{algebraic-reduction}, viewed from the constraint
  side rather than the protocol side.

  \item \textbf{Separate what is field-dependent.} Some cost may follow from the
  field rather than the relation. That boundary belongs to
  \opref{field-hardware}, so the two items should not count the same cost twice.
\end{itemize}

\opitem{zk-backprop}{Zero-knowledge proof of backpropagation.}
To our knowledge, as of August 2026, no zero-knowledge system has demonstrated
bit-exact verification of the gradients for every parameter tensor in a
transformer backward pass. This claim assumes a declared IEEE~754 BF16/FP32
operation sequence. It holds even at this problem's $10^{6}$-parameter milestone.

Existing demonstrations cover simple models such as logistic
regression~\citep{garg2023zktraining},
quantized fixed-point training at ${\sim}10$M
parameters~\citep{abbaszadeh2024kaizen,sun2024zkdl}, quantized LoRA
fine-tuning~\citep{liao2026verilora}, or forward inference only. Ray and
El~Jazmi~\citep{ray2025backprop} report an \texttt{f32} MLP
backpropagation/SGD proof branch in a zkVM. However, their published description
does not specify the checked floating-point relation at the operation level.
What the branch establishes therefore remains unresolved.

The backward pass reuses the same primitives: MACs, Merkle paths, and lookups.
Yet no complete forward-plus-backward circuit has been exercised at scale in a
zkVM.
\emph{Question:} produce either (a) a functional proof for a transformer
with at least $10^6$ parameters and bit-exact gradient agreement with GPU
execution, or (b) an identified fundamental obstacle. Sub-problems include:
\begin{itemize}
  \item \textbf{Establish the field-arithmetic baseline precisely.} Training
  proofs over finite-field proxies have progressed further. Record precisely
  what they have demonstrated and at what scale. This keeps the negative claim
  checkable and separates the float-semantics gap from the scale gap.

  \item \textbf{Locate the risk.} Any obstacle is more likely to lie in what
  backward adds. These additions include transposed GEMMs against differently
  committed operands and accumulation into gradient buffers shared across
  microbatches. They also include the optimiser-state transition. Isolating the
  hard component is more valuable than one monolithic scaling attempt.

  \item \textbf{Witness size may bind before constraint count.} Backward needs
  the forward activations. Retention and witness volume may therefore become the
  first wall, rather than proving throughput. The commitment structure in
  \Cref{appx:proof-cost:commitments} determines where that wall lies.

  \item \textbf{Pin what ``bit-exact gradient match'' means.} Which device,
  kernel, and reduction order define the reference? Without those choices, the
  success criterion is not testable. The required reference is exactly what
  \opref{det-attention} and \opref{precompile-portability} provide.

  \item \textbf{Say in advance what a fundamental obstacle would look like.}
  A milestone with an unspecified failure mode cannot be falsified; it can only
  be postponed. Candidate forms include a superlinear blow-up in the backward relation.
  Another is an arithmetisation of gradient accumulation with no sub-quadratic
  form.
\end{itemize}

\opitem{reachability}{Reachability of the final weights under trainer-authored inputs.}
The protocol binds the executed computation to a declared recipe. It does not
constrain the \emph{inputs} that the trainer authors. These include dataset
content, curriculum, batch composition, learning-rate and regularisation
schedules, and auxiliary losses.

A trainer that already holds target weights therefore faces a \emph{control}
problem, not a learning problem. The task is to steer the declared run towards a
known destination through whichever inputs remain free. There is no reason that
navigating to a known target should cost what finding that target cost.
\emph{Question:} characterise the set of reachable final weights under the
protocol's constraints, as a function of which inputs are pinned. Sub-problems
include:
\begin{itemize}
  \item \textbf{Control authority, per channel.} Which trainer-authored inputs
  provide useful authority towards a known target, and how much? The dataset
  channel clearly has decisive authority over the resulting model's
  \emph{capabilities}. Training on model-generated data is standard industrial
  practice. However, to our knowledge, no one has measured whether any channel
  provides authority towards a \emph{specific} target. This includes the dataset
  channel.

  \item \textbf{Capacity as a coarse upper bound.} The entropy of
  trainer-authored inputs bounds what they can transfer into the run. The dataset
  dominates that budget by many orders of magnitude. Constraining low-entropy
  inputs therefore does not materially reduce the budget. This includes an
  auditor-supplied initialisation seed, although that seed does close its own
  channel. The bound is loose only in the useful direction. Distillation is
  efficient precisely because it is not literal encoding.

  \item \textbf{Which pins actually shrink the set.} The obvious candidate is
  restricting training data to a public corpus that an auditor can check.
  Research must establish whether this shrinks or closes the reachable set.
  It must also ask whether ``this text was not model-generated'' is decidable at
  the required scale. The answers determine whether data provenance is a
  solution or merely a delaying tactic.

  \item \textbf{What survives a non-singleton reachable set.} If a conformant run
  certifies membership in a large set of possible outcomes, its certificate
  describes the \emph{process}, not the artefact. The
  governance claims that remain supportable under this reading should be stated
  explicitly, rather than left to the reader.
\end{itemize}

\opitem{weight-smuggling}{Weight-smuggling attacks that look like regular training.}
The chaining and aggregation attacks in \Cref{sec:weight-smuggling} could exploit
four entry points:
\begin{itemize}
  \item \textbf{At initialisation.} Direct substitution of the initial weights
  can be closed. Derive the weights from an unpredictable seed, then recompute
  their full commitment. This becomes possible once the V8 upgrade specifies the
  initialiser semantics (\Cref{appendix:formalization}).

  \item \textbf{Through the data.} One possible attack uses crafted data to move
  the model towards a chosen target during declared training. Dataset distillation
  shows that a few synthetic examples can reproduce much of a full dataset's
  effect in studied settings~\citep{wang2018datasetdistillation}. It does not show
  that crafted data can steer any model to any target.

  A low-rank update can also have a much smaller description than a complete
  weight update~\citep{hu2022lora}. This does not show that an arbitrary target
  difference is low rank. Low rank also differs from coordinate sparsity. Even a
  rank-one update can change nearly every coordinate.

  \item \textbf{Mid-run.} One tampered operation can splice in a payload. This is
  sparse across operations, even if the payload changes many coordinates. Under
  the idealised single-operation challenge in \Cref{sec:sampling-guarantees}, each
  sampled step chooses one of $N_{\mathrm{ops}}$ operations uniformly. For one
  changed operation in one step, the detection probability is about
  \[
    \frac{k_{\mathrm{step}}}{T}\,
    \frac{1}{N_{\mathrm{ops}}}\,
    \bigl[1-(1-f)^k\bigr].
  \]
  Here, the auditor samples $k_{\mathrm{step}}$ of $T$ steps and then checks $k$
  entries of the selected operation. The changed-entry share is $f$.

  This probability is about $k_{\mathrm{step}}/(T N_{\mathrm{ops}})$ only when
  the entry sampler almost surely detects the changed entries after selecting that
  operation. Sampling optimiser-update positions in every step would catch
  coordinate-dense splices.
  It would not reliably catch edits confined to a few coordinates
  (\Cref{sec:sampling-guarantees}).

  \item \textbf{Through merging.} Some distributed training methods average the
  weights of separately trained replicas~\citep{douillard2024diloco}. A declared
  averaging step could also absorb a replica trained off the books.
\end{itemize}
\emph{Question:} for a stated attack goal and accounting boundary, which smuggling
routes remain open against a bit-exact verified run? What does each route cost the
attacker? Sub-problems include:
\begin{itemize}
  \item \textbf{Speed of steering.} Under a declared learning-rate schedule, how
  many declared steps are needed to reach a chosen target through crafted data?
  \opref{reachability} asks which targets are reachable at all. This question asks
  how quickly and visibly they can be reached.

  \item \textbf{Detectable signatures.} Do smuggling runs leave traces in values
  that the protocol already commits to? Examples include loss curves, gradient
  norms, update statistics, and data statistics. Can an attacker hide these traces?

  \item \textbf{Recipe restrictions.} Which restrictions close each route?
  Candidates include learning-rate bounds, bans on averaging with undeclared
  replicas, and training-data provenance requirements. What would these
  restrictions cost honest trainers?

  \item \textbf{Accounting unit.} Could a global budget across all covered
  training runs be enforced more reliably than per-run caps? What coverage would
  prevent imported training from escaping this budget (\opref{compute-accounting})?
\end{itemize}

\subsection{Deterministic execution on hardware}

\opitem{det-attention}{Deterministic attention backward kernels with ${<}5\%$ overhead.}
Our end-to-end measurements show that deterministic mode is dominated by
attention backward. Its step-time tax is $1.6$--$8.2\%$ and varies with sequence
length. cuBLAS GEMMs contribute ${\sim}0.8\%$, while NCCL contributes
${\sim}0$ (\Cref{app:determinism-details}).
Nearly the entire tax therefore falls on the attention backward kernel.
At attention backward's typical share
of step time, the measured tax implies
a kernel-level overhead of tens of percent. DASH~\citep{qiang2026dash} achieves
1.28$\times$ over baseline deterministic mode.

\emph{Question:} can attention backward be deterministic \emph{by construction}?
Possible methods include fixed warp-level reduction trees or an algebraic
reformulation. The target is under 5\% overhead for sequence lengths $\geq 8$K.
Sub-problems include:
\begin{itemize}
  \item \textbf{Decompose the tax.} Three sources contribute separately: atomic
  accumulation into shared gradient buffers, split-$K$ work decomposition, and
  schedule-dependent reduction order. Only a decomposition shows which source is
  worth attacking. DASH performs this isolation for FlashAttention-3. It
  attributes the loss to poor scheduling of the compute and gradient-reduction
  phases, not to determinism itself~\citep{qiang2026dash}. It remains open whether
  this finding covers our end-to-end band. That band includes FlashAttention-2
  and regimes that DASH does not cover.

  \item \textbf{By construction versus by serialisation.} Today's deterministic
  modes achieve reproducibility by constraining the schedule. Fixed warp-level
  reduction trees would instead remove order dependence. So would fixed-point or
  exactly-rounded, order-independent accumulation.

  \item \textbf{How much determinism is actually required.} Within one committed
  run on pinned hardware, the protocol needs bit-exact reproducibility. Bit-exact
  agreement across vendors or silicon generations is strictly harder. That is
  \opref{precompile-portability}'s problem, not this one. Conflating the two
  properties overstates both the requirement and its cost.

  \item \textbf{Interaction with recomputation.} Activation checkpointing
  recomputes the forward pass during backward. Determinism must therefore hold
  between the original and recomputed executions, not merely within one.

  \item \textbf{Baseline discipline.} Our end-to-end tax uses a non-deterministic
  baseline. DASH's $1.28\times$~\citep{qiang2026dash} uses baseline deterministic
  mode. A target of ``under 5\%'' must pin its own reference. Otherwise, the target
  is not falsifiable.
\end{itemize}
This is the architecture's single largest training-side cost for an honest
trainer.

\opitem{precompile-portability}{Multi-architecture precompile specification.}
Each GPU---H100, MI300X, or Gaudi3---has its own accumulation-tree topology,
rounding mode, and subnormal handling. Each currently needs a hand-engineered
precompile. \emph{Question:} can a parameterised family of precompiles be defined,
instantiated automatically from a compact hardware descriptor, and validated
against silicon? Sub-problems include:
\begin{itemize}
  \item \textbf{A specification language for vendor numerics.} It must express
  the vendor differences above. It must also be restrictive enough to compile a
  descriptor to constraints automatically, rather than by hand.

  \item \textbf{Validation methodology.} How many test vectors are enough to show
  that a specification matches silicon, and how should they be chosen? Random
  vectors are weak against the corner cases that matter. These include
  subnormals, ties-to-even boundaries, and overflow. The real question is how to
  generate tests adversarially or direct them towards coverage.

  \item \textbf{Disagreement between silicon and its documentation.} Vendor
  numerics are under-specified. When documentation and hardware disagree, the
  proof must match the hardware. A validated descriptor is therefore an
  empirical artefact with provenance and an expiry. It is not a transcription of
  a datasheet.

  \item \textbf{Mid-run hardware heterogeneity.} Frontier runs continuously
  replace failed accelerators, sometimes with another SKU or firmware revision.
  Suppose a run's \texttt{arch\_spec} pins one numeric descriptor. Replacements
  must then match it bit-exactly. Alternatively, the specification can allow a
  declared set of descriptors and bind which one executed each step. This is a
  commitment-structure question, not only a numerical one.

  \item \textbf{What validation is worth without vendor cooperation.} Black-box
  characterisation can show agreement on tested inputs. It cannot rule out
  input-dependent behaviour elsewhere. Whether vendor disclosure is needed to
  close that gap determines how much of this problem is technical and how much is
  institutional.
\end{itemize}
Cross-architecture portability determines whether trainers on non-NVIDIA silicon
can use verification. It therefore determines whether verification can be
mandated without prescribing a vendor.

\subsection{Hardware trust anchors}

\opitem{tap-hardware}{Open-hardware network TAP at line rate.}
Tier~1 of \Cref{network-tap} requires a passive, auditable device. It must hash
inter-node traffic at 400+ Gbps. Commercial TAPs already exist. However, an
open-hardware, formally verified device would establish a new trust primitive for AI
governance.
\emph{Question:} can field-programmable gate array (FPGA) fabric stream SHA-256 at
400+ Gbps while meeting two further conditions? The register-transfer-level (RTL)
logic must be formally verified. The physical packaging must be tamper-evident.
Sub-problems include:
\begin{itemize}
  \item \textbf{Where the throughput actually binds.} Line-rate hashing is the
  headline requirement. However, framing may become the bottleneck first. Framing
  means parsing InfiniBand or RoCE headers and removing volatile fields before
  hashing (\opref{wire-to-tensor}). Unlike the hash core, this work depends on the
  protocol.
  \item \textbf{What property the RTL verification establishes.} Proving the hash
  core's functional correctness is the easy half. The governance-relevant property
  is \emph{passivity}. The device cannot inject, modify, or drop traffic. It cannot
  exfiltrate payloads. Passivity concerns RTL, architecture, and physical
  design to the same degree. Formally stating passivity is part of the problem.
  \item \textbf{Tamper-evidence and who inspects it.} A tamper-evident enclosure is
  only as good as its inspection regime. The regime must define the inspector,
  frequency, authority, and response to a positive finding. The device and the
  institution must be designed together.
  \item \textbf{What open hardware buys over an audited commercial TAP.} An open
  design removes reliance on one vendor's attestations. However, the device must
  still be fabricated somewhere. Open hardware therefore moves supply-chain trust
  instead of eliminating it. The engineering case must state this trade explicitly
  and identify which threat models actually improve.
  \item \textbf{A moving target.} 800\,Gbps links are already being deployed. A
  design that exhausts its performance margin at 400\,Gbps will be obsolete on
  arrival. Its scaling path therefore matters as much as its point design.
\end{itemize}
This is the least cryptographic item on the list.
It is still worth attempting because the resulting primitive would outlive this
protocol.

\opitem{wire-to-tensor}{Wire-to-tensor mapping for the network anchor.}
The network anchor (\Cref{network-tap}) is useful only if its wire observations
match the trainer's tensor commitments. This requires a canonical mapping from
logical tensors to NCCL's on-wire byte sequence. That mapping is not a stable
public interface today.

NCCL's byte layout depends on runtime settings. These include
\texttt{NCCL\_BUFFSIZE}, \texttt{NCCL\_NTHREADS}, and the selected algorithm.
Ring-allreduce states on the wire are partial reductions, not tensor slices. The
anchor must also parse InfiniBand or RDMA-over-Converged-Ethernet (RoCE) framing.
Before hashing, it must remove the volatile PSN and ICRC fields.

\emph{Question:} for a stated threat model and adoption constraint, which routes
meet both the required physical-wire binding and the deployment constraint? Three
candidates are available:
\begin{enumerate}
  \item \textbf{Patched NCCL with a documented canonical format.} A forked NCCL
  defines a version-locked relation from committed tensors, collective state, and
  runtime configuration to transmitted payloads and offsets. This is the cleanest
  path and supports open-source verification. It also needs continuing maintenance
  as NCCL evolves.

  If the documented semantics match actual transmission, the anchor can bind
  tensor claims to physical wire observations. This route trusts the independent
  anchor, the audited patch, and evidence that the deployed patch produced the
  observed format.

  \item \textbf{Canonicalisation shim at the NCCL-to-transport boundary.} A signed,
  auditable library sits between NCCL and the network interface controller (NIC).
  It emits a canonical copy of each outgoing payload for the anchor to hash. This
  adds memory traffic but avoids NCCL's internals.

  This route matches physical-wire assurance only if an independent mechanism
  binds the shim's record to NIC transmission. That mechanism could sit in the NIC
  or the anchor. Without it, the shim and trainer-host software add a software
  trust root. A signature alone does not prove what reached the wire.

  \item \textbf{Vendor-supported ``verifiable NCCL'' mode.} A first-class setting
  provides documented wire semantics. This path needs neither a fork nor a shim.
  It does require vendor cooperation.

  If those semantics match actual transmission, the anchor can again bind tensor
  claims to physical wire observations. This route trusts the independent anchor,
  the vendor implementation, and evidence that the mode governed the observed
  transmission.
\end{enumerate}

The three routes therefore do not automatically provide equal assurance. They
also differ in deployment effort, maintenance burden, and required institutional
leverage.

Three subsidiary questions remain:
\begin{itemize}
  \item \textbf{Queue-pair mapping auditability.} During initialisation,
  trainer-host software maps NCCL traffic to queue pairs (QPs). How can the
  auditor check this mapping without trusting the trainer's host? The Tier~1
  anchor or Tier~2 SmartNIC firmware could publish an independent view. The run
  is rejected if that view differs from the trainer's view.

  \item \textbf{Tier~2 attestation protocol.} NVIDIA BlueField's DOCA software
  framework provides signing and attestation. Its roots are secure boot and a
  device identity key. The end-to-end protocol remains unspecified. It must cover
  continuous commitment streaming, key rotation, and revocation. A standardised
  reference design would help.

  \item \textbf{Chunk-boundary alignment.} BF16 elements use 2 bytes. Merkle leaves
  use tens of kilobytes. If an element crosses a leaf boundary, checking it needs
  two path openings instead of one. Every candidate must therefore specify its
  alignment guarantees.
\end{itemize}

\opitem{intranode-witness}{Intra-node physical witness.}
Neither network-anchor tier observes intra-node traffic over NVLink. Therefore,
hybrid FSDP+TP training has no physical witness for its tensor-parallel layer.
\emph{Question:} can GPU-resident trusted execution provide continuous, attested
hashing of intra-node collective payloads at the required bandwidth? Examples
include NVIDIA Confidential Compute or a successor mechanism. Sub-problems include:
\begin{itemize}
  \item \textbf{Whether continuous hashing is feasible at all.} NVLink bandwidth
  exceeds inter-node rates by roughly an order of magnitude. A primitive that works
  at the network boundary therefore may not scale here. Current Confidential
  Compute exposes neither the interface nor the throughput this task needs.
  \item \textbf{The minimal primitive.} The protocol may not need full payload
  hashing. A counter, a sampled hash over challenge-selected windows, or an attested
  summary may suffice. The downstream protocol already uses sampling. A witness
  that is itself sampled is therefore not obviously weaker. Finding the weakest
  adequate primitive would redirect the hardware request from ``attest everything''
  to something a vendor might plausibly ship.
  \item \textbf{The trust profile it buys.} A GPU-resident trusted execution
  environment roots the intra-node witness in the accelerator vendor. That is the
  same vendor whose execution the anchor aims to check independently. It also
  repeats this paper's concentration concern about hardware-first approaches. A
  witness that is not independent of the observed party is weaker than the network
  anchor. The protocol should state which guarantees survive this substitution.
  \item \textbf{Avoiding the problem by placement.} Restricting tensor parallelism
  to inter-node links would put its traffic within the existing anchor's view. This
  restriction would cost throughput. Whether that cost is below the hardware
  programme's cost is a co-design question (\opref{codesign}). It may be the cheaper
  answer.
\end{itemize}
Until one of these routes is available, the anchor's independence claim covers only
inter-node traffic.

\opitem{sdc}{Distinguishing SDC from adversarial deviation.}
On frontier hardware, silent data corruption (SDC) produces bit flips. From the
changed value alone, one bit flip is indistinguishable from an intentional change.
Fleet-scale CPU studies show meaningful prevalence across the device
population~\citep{dixit2021sdc,hochschild2021cores}. They find roughly a few
defective cores per several thousand machines. They do not report an absolute
event rate.

A rate of roughly one event per $10^{4}$--$10^{5}$ GPU-hours is a working
assumption. It is not an interval inherited from measurement. It adds an error
frequency per defective device to the measured prevalence. That frequency varies
``by many orders of magnitude'' across defective cores
~\citep{hochschild2021cores}. Applying CPU measurements to accelerators adds a
further assumption. This research programme should test both assumptions on
current GPUs.

Consider a 405B-class run such as Llama~3.1~405B. Its report gives
${\sim}3.8\times10^{25}$ FLOPs on $16{,}384$ H100s
~\citep{dubey2024llama3}. At realistic utilisation, this is about
$3\times10^{7}$ GPU-hours. Under the assumed rate band, the run would have about
$300$--$3{,}000$ SDC events. The arithmetic is
$3\times10^{7}/10^{5}=300$ to $3\times10^{7}/10^{4}=3{,}000$. The low end uses
the optimistic rate of one event per $10^{5}$ GPU-hours.

A sophisticated adversary could make structured deviations look statistically
natural. There is a basic limit to any test based only on those observations.
Suppose honest observations follow distribution $P$. A test that accepts $P$ with
some probability accepts a distribution-matching attacker with the same
probability. No observation-only test can distinguish the two.

Statistical separation therefore needs a restricted attacker that cannot match
$P$, or side information that the attacker cannot reproduce. An impossibility
result for the unrestricted observation-only setting is a valid answer.

\emph{Question:} under which attacker restrictions or side information can tests
separate SDC from adversarial injection? Candidate tests must cover random,
memoryless patterns. They must also cover correlations with known failure modes,
including DRAM row faults and thermal events.

Empirical work on current GPUs is a prerequisite. It should measure absolute
events per GPU-hour, defective-device prevalence, and error frequency per
defective device. It should also characterise the observed error patterns and
their correlations with known failure modes.

The affordability of full coverage with bit-exact checking remains open. It
depends on measured GPU rates, how faults propagate into bound values, and the cost
of handling each detected mismatch. Under the sampled protocol in
\Cref{sec:sampling-guarantees}, an isolated mismatch distributed across checkable
values is rarely sampled. Measured fault footprints are needed to calculate the
honest-run rejection rate.

By contrast, some designs bind every coefficient of every step. The weak tier of
\opref{hybrid} is one example. Such designs provide full coverage, though at
relaxed precision. The working rate predicts $300$--$3{,}000$ raw events. A
full-coverage, bit-exact relation exposes the subset that changes bound values.
Masking may reduce this number, while propagation may create several mismatches
from one event. Each detected mismatch needs one of three responses:
\begin{itemize}
  \item roll back to a clean checkpoint, then deterministically re-execute and
  re-commit the affected suffix, paying the re-proving cost;
  \item admit a tolerance width; or
  \item allow a bounded number of unexplained mismatches.
\end{itemize}

Which response is cheapest remains open. The latter two admit deviations that an
adversary could use. Re-execution instead preserves bit-exactness but incurs its
own cost. If a design uses an allowance, hardware faults set a floor on accepted
deviation. Soundness sets a ceiling. A viable design needs a non-empty gap between
them.

\emph{Question:} how should an allowance be sized? The two forms have different
profiles:
\begin{itemize}
  \item A tolerance band is wide in adversary-choosable bits. However, it caps the
  magnitude of each coordinate's change.
  \item A mismatch allowance is narrow in bits. However, it does not itself bound
  the change's magnitude. An exponent-bit flip can create a very large or non-finite
  value.
\end{itemize}

Under a stated attacker restriction or source of side information, do honest runs
pass while deviations above a predeclared impact threshold fail?

\subsection{Software-only variants}
While hardware-based trust anchors give stronger confidence, they are expected to take longer to develop. Thus, considering software-only solutions can be of interest.

\opitem{salts-for-liveness}{Salting for liveness evidence.} Adding more interactivity to the protocol can help ensure the declared computation is running live, instead of having been run in the past, without having to rely on network TAPs. The verifier could periodically send randomly generated strings, to be used as seeds for the training process or keys for hashing operations. A formal security analysis of such practices could help make sure it is robust.

\opitem{timing-for-location}{Timing-based challenges to pin down locations.} With physical anchors, it is straightforward to determine the location of compute clusters. Without them, one might send time-sensitive challenges to identified clusters, from various locations, and deduce approximate location from the latency data~\cite{gueye2004geolocation,abdou2018securegeolocation}.

\subsection{Protocol extensions}

\opitem{moe}{Mixture-of-experts verification.}
The base protocol assumes that \texttt{arch\_spec} determines the computation graph
in advance. Mixture-of-experts (MoE) training relaxes this assumption. Routing is
data-dependent. At runtime, the router network decides which expert processes each
token.
\emph{Question:} extend the protocol to data-dependent routing at frontier MoE
scale. The extension's shape is clear and appears below. Its cost remains open, as
noted for each element:
\begin{itemize}
  \item \textbf{Routing commitment.} A per-step Merkle root
  $R(\mathtt{routing}_t)$ must join the hash chain. It commits to the top-$K$ experts
  selected for each token. Without this root, the proof does not bind the routing
  decisions.
  \item \textbf{Top-$K$ selection verification.} Selecting the top $K$ BF16 scores
  is non-smooth and data-dependent. Verification needs either a dedicated
  precompile or a decomposition into pairwise comparisons. Pairwise comparisons
  are cheap per element, but scale with $N \log K$ per token. Here, $N$ is the
  total number of experts and $K$ is the number selected.
  \item \textbf{Deterministic tie-breaking.} Ties between expert scores are rare in
  BF16, but possible. A documented rule, such as expert ID, must resolve them. The
  rule must appear in \texttt{arch\_spec}.
  \item \textbf{All-to-all canonicalisation.} Expert parallelism uses all-to-all
  communication to send tokens to their selected experts. Its wire byte layout
  depends on the committed routing. This is a harder version of
  \opref{wire-to-tensor}: the mapping is data-dependent, not merely
  configuration-dependent.
  \item \textbf{Load-balancing auxiliary losses.} Importance and load losses must
  appear in \texttt{arch\_spec}, so the proof binds them. Existing precompiles for
  softmax, log, and division already cover their operations.
\end{itemize}
\emph{Framing for cost arguments.} MoE proof cost per step scales with active
compute per token, not total parameters. With $K$ active experts per token, the cost
is approximately $K$ times one comparable dense model's cost, plus the routing
overhead above. Here, the comparable dense model has the dimensions of one expert.
This scaling arises because each sampled output entry requires checking all $K$
active experts. A 1T MoE with 32B active per token, such as Kimi~K2
with $K{=}8$, therefore costs substantially less per step than a hypothetical 1T
dense model. However, it costs more than a 32B dense model.
Sparse architectures account for an increasing share of frontier capability.
Therefore, a dense-only verification architecture covers a shrinking share of the
runs that a threshold regime is meant to reach.

\opitem{rl}{Reinforcement-learning post-training.}
Post-training creates two distinct verification problems. Fixed-dataset preference
optimisation, such as standard direct preference optimisation (DPO), consumes
committed preference data. Online methods such as proximal policy optimisation
(PPO) and group relative policy optimisation (GRPO) generate model-dependent
rollouts. They may use human feedback (RLHF), AI feedback (RLAIF), or deterministic
rules.
\emph{Question:} extend the protocol to both settings without treating them as one
execution model. Fixed-data methods require dataset and loss binding. Online methods
require a finite relation that checks how the model generated and selected its own
training data. Sub-problems include:
\begin{itemize}
  \item \textbf{Fixed-data preference optimisation.} Commit the preference dataset,
  ordering, loss, reference policy, and any randomness.
  \item \textbf{Sampler-state commitment.} Commit the random-number-generator seeds,
  sampling hyperparameters, tokenizer, sampler implementation, numerical dialect,
  and state-consumption order. These inputs bind each draw only when the resulting
  sampler semantics are single-valued.
  \item \textbf{Rollout verification.} A fully specified synchronous sampler can make
  completions deterministic functions of the committed policy and sampler state.
  Best-of-$N$ and rejection sampling remain deterministic when candidate generation,
  stopping, and selection rules are fully specified. Asynchronous actor merges,
  replay buffers, external environments, and human judgements add further state.
  The proof must check a frozen transcript against a declared finite execution
  relation. It must establish that the committed policy generated the
  policy-originated events, not merely that the trainer committed to them. External
  events must satisfy declared environment or provenance relations.
  \item \textbf{Reference-model binding.} PPO and RLHF often use a frozen reference policy
  for Kullback--Leibler (KL) regularisation. This policy is usually the pre-RL
  checkpoint, which the pre-training chain can verify. The policy's
  \texttt{arch\_spec} should reference that commitment.
  \item \textbf{Reward commitment.} A rule-based reward may use test-case pass/fail
  for code or formal checking for mathematics. The specification must pin the rule,
  toolchain, tests, timeouts, and execution environment. A trained reward model needs
  its own $h_{\text{commit}}$. Human judgements need committed provenance and an
  admission rule. The policy's \texttt{arch\_spec} should reference these objects.
  \item \textbf{PPO clipping.} The clipped objective creates data-dependent gradient
  masking. Clipping depends on per-token probability ratios. Checking the condition
  for each sampled token adds a per-sample condition, structurally similar to
  routing verification.
\end{itemize}
\emph{Recommended first target.} Start with synchronous, rule-based RL for
mathematics or code in the style of DeepSeek-R1. Pin the sampler, execution
environment, and reward checker. This removes the learned reward-model chain while
retaining the central rollout-generation problem.
Without this extension, the protocol does not cover the computation that produces
post-training updates.

\opitem{multisite}{Multi-site training verification.}
Frontier training increasingly spans multiple datacentres connected by wide-area
network (WAN) links. The anchor model assumes an auditor-controlled observation
point on every inter-node link. Realistic deployments divide link control across
administrative boundaries.
\emph{Question:} how should the protocol adapt when different parties control some
network segments? Sub-problems include:
\begin{itemize}
  \item \textbf{Cross-anchor commitment consistency.} Multiple anchors must produce
  one coherent view of a single run. They must agree on what each anchor observed
  and reconcile their partial views into one chain.
  \item \textbf{Clock skew and latency heterogeneity.} WAN links break the
  near-synchronous assumptions that make a single-site anchor chain easy to order.
  The anchor's time-binding uses (\opref{anchor-exploitation}) depend on how tightly
  events can be ordered across sites.
  \item \textbf{Byzantine-robust aggregation.} One site's anchor may be operated by
  a party that is also the adversary. Per-site hashes must combine so that a corrupt
  site cannot forge agreement. The protocol must fall back to a stated weaker
  guarantee instead of failing open.
  \item \textbf{Incentive alignment across organisations.} Several parties may
  share one run. It is not obvious who is the trainer or auditor. Nor is it obvious
  who pays for a rejected proof. The base protocol assumes singular
  roles, but this deployment does not.
  \item \textbf{Whether multi-site helps rather than hurts.} Inter-site links are
  fewer, slower, and already instrumented. These properties make them unusually
  natural observation points. It is worth establishing whether geographically
  distributed training is easier to anchor than one site with fast internal
  fabric. If so, this problem
  partly reverses. The difficulty in \opref{intranode-witness} then becomes the real
  constraint.
\end{itemize}
An architecture that assumes single-site link control addresses a shrinking share
of the runs it is meant to govern.

\subsection{Protocol tooling}

\opitem{converter}{Converter from training code to \texttt{arch\_spec}.}
The protocol uses \texttt{arch\_spec} as the canonical description of the training
computation. Real trainers instead write PyTorch or JAX code. A converter must bridge
these formats for a declared source dialect. Its pipeline needs:
\begin{itemize}
  \item forward-graph capture with \texttt{torch.export} or \texttt{torch.fx};
  \item backward-graph capture with \texttt{functorch};
  \item mapping ATen operations to declared \texttt{OpSpec} variants;
  \item capture of optimiser steps, randomness, distributed annotations, and
  effects; and
  \item versioned serialisation.
\end{itemize}
\emph{Question:} can a bounded converter meet all three requirements below?
\begin{enumerate}[label=(\alph*)]
  \item It preserves training semantics for a pinned source dialect and supported
  operation set.
  \item Code related by a declared set of supported refactorings produces the same
  deterministic \texttt{arch\_spec}.
  \item Unsupported or untraced behaviour causes an explicit error instead of
  silently dropping structure.
\end{enumerate}
Sub-problems include:
\begin{itemize}
  \item \textbf{Measuring coverage.} Coverage needs a versioned denominator. Build a
  declared corpus of representative frontier training programs and record every
  operation they reach. The corpus should include fused and vendor-specific kernels.
  Coverage can then be measured rather than asserted.
  \item \textbf{Capture fidelity.} Graph coverage does not prove that graph capture
  preserved the whole program. The contract must cover Python control flow,
  mutation, data loading, custom kernels, randomness, and distributed side effects.
  Uncaptured behaviour must make conversion fail.
  \item \textbf{Bounded canonicalisation.} Program equivalence is undecidable in
  general. The practical target is therefore a normal form stable under a declared
  set of common refactorings. A compiler pass must reach that form deterministically.
  \item \textbf{Failing loudly by construction.} Use a whitelist discipline for the
  captured graph. Unknown operations are errors, never no-ops. The converter must
  also establish that every captured node has a translation.
  \item \textbf{Longevity.} Framework internals change faster than governance
  timelines. Freeze the \texttt{arch\_spec} semantics and serialisation independently
  of the producing framework. The commitment should also record versioned converter
  provenance.
\end{itemize}
Two correctness contracts must remain separate. Relative to a committed
\texttt{arch\_spec}, the proof system checks only the relation declared by that
specification. A converter bug cannot make an execution outside that relation pass,
although it may reject honest code or mistranslate its intended computation.
If governance instead claims that approved source code ran, mistranslation can yield
a valid proof of the wrong specification. Source-to-spec fidelity is then a separate
end-to-end assurance obligation. ZKML~\citep{chen2024zkml} provides a precedent for
compiling inference models from TensorFlow into Halo2 circuits. Training also
requires backward capture, optimiser state, randomness, and distributed annotations.

\subsection{Proof-layer design space: randomness, transferability, and hybrid protocols}

\opitem{beacons}{Beacon-based challenge derivation.}
This comparison is conditional on enforcement infrastructure, which is itself part
of the problem. Under that condition, \Cref{sec:proof-options} identifies public
randomness beacons as preserving transferability while preventing offline challenge
grinding. The challenges are unpredictable to the trainer before commitment freeze.
They are publicly verifiable afterwards.
\emph{Question:} design and analyse the security of the sampling protocol in
\Cref{sec:sampling-guarantees} when challenges come from a beacon. Sub-problems
include:
\begin{itemize}
  \item \textbf{Which messages can be beacon-timed at all.} The commitment must be
  provably frozen before the relevant beacon output. Post-freeze challenge indices
  may combine that output with the frozen transcript. However, some points are
  determined by values the trainer has already committed. Such an opening point is
  not a free challenge, and no beacon can make it unpredictable. Any design must
  first separate these two kinds of point.
  \item \textbf{The effective grinding budget.} Establish the budget when retries
  are publicly countable instead of free offline hash queries. Then recalculate the
  sample-count budgets in \Cref{tab:sample-count}, relative to a query-bounded
  Fiat--Shamir analysis.
  \item \textbf{Registration infrastructure.} Determine how to enforce
  commit-before-release ordering and count attempts. Timestamping alone does not
  bound the number of admitted attempts. The required role is therefore metering.
  The network anchor could provide it (\opref{anchor-exploitation}).
  \item \textbf{Liveness.} Match beacon cadence to the audit schedule without
  delaying training's critical path. Threshold-cryptography beacons, for example,
  operate on a cadence of seconds.
\end{itemize}
The candidate families are threshold-cryptography beacons~\citep{syta2017scalable},
verifiable delay functions~\citep{boneh2018vdf}, and TEE-attested generators. They
differ in cadence, liveness, and trust assumptions. A deployment-grade
recommendation must compare them under this paper's adversary model.

\opitem{hybrid}{Hybrid tolerance/sampling protocols.}
\Cref{sec:hybrid} sketches a two-tier construction. The weak tier binds every
coefficient of every step to a declared tolerance band. This gives full coverage at
relaxed precision. Random, unpredictable steps may also receive bit-exact checks,
exact anchors, or both.
\emph{Question:} define one weak execution relation and one frontier operating
point. Then formalise the combined construction, prove a useful guarantee, and
measure its end-to-end cost. Sub-problems include:
\begin{itemize}
  \item \textbf{Weak-tier relation.} Specify what the tolerance constrains: operation
  outputs, state transitions, or a surrogate computation. Also specify whether the
  bound is absolute, relative, or exponent-dependent. These choices determine both
  proving cost and the admitted deviations.
  \item \textbf{Drift composition.} Prove a theorem that bounds the final-weight
  effect of adversarially correlated deviations within the tolerance. Predeclare
  the dynamics class, baseline trajectory, norm, probability model, and largest
  useful displacement. A positive result must remain below that threshold and
  change a stated design or governance decision. A matching impossibility or lower
  bound is also informative.
  \item \textbf{Residual freedom and channel capacity.} Displacement does not
  capture every threat. Every design admits some deviations that pass all checks.
  First measure their raw freedom as the log-cardinality of accepted traces relative
  to an honest baseline. Call this quantity channel capacity only after specifying
  an adversary, encoding, and decoder.

  For a declared operating point $(T,d,N,b,\eta/\mathrm{ulp})$, a width-$\eta$
  tolerance band has the following order-of-magnitude independent-coordinate count:
  \[
    T d\log_{2}(\eta/\mathrm{ulp}) \quad \text{bits per run}.
  \]
  Here, $\mathrm{ulp}$ is the local gap between adjacent representable floats.
  Correlations and varying float spacing may reduce this raw count.

  A bit-exact tier allowing at most $N$ unexplained mismatches admits approximately
  \[
    N\bigl(\log_{2}(Td)+b\bigr) \quad \text{bits per run}.
  \]
  These bits encode each mismatch's position and replacement value. The ratio of
  the two counts depends on all five operating-point parameters. Any reported
  comparison must publish them and a sensitivity analysis.

  Minimising raw residual freedom is conservative in the intended direction. Its
  operational value to an adversary remains an ML question outside this roadmap's
  scope.
  \item \textbf{Sizing the tolerance.} Bit-exact determinism is achievable
  (\Cref{enforce-determinism}), so $\eta$ need not accommodate honest
  nondeterminism. Proving cost may still justify a non-zero $\eta$. Measured fault
  rates and remediation costs may constrain it further. If rollback and re-execution
  are affordable, $\eta$ need not absorb those faults; see \opref{sdc}.
  \item \textbf{Soundness class and affordability.} State the covered adversary,
  remaining governance consequence, and interaction between exact backstops and
  unsampled tolerated drift. Relate this class to \Cref{sec:sampling-guarantees}.
  Then measure whether the complete protocol is affordable at the declared frontier
  operating point.
\end{itemize}
The ingredient-level questions remain separate problems; see
\opref{float-native}--\opref{field-hardware}. The authors are investigating this
construction in companion work.

\opitem{transferability}{Transferability requirements of governance settings.}
\Cref{sec:proof-options} separates public verification of sampled checks from
confidence in how those checks were selected. Auditor-chosen seeds require
non-collusion even when later disclosed. Transcript-hash and beacon-based
selection instead require the grinding and registration safeguards stated there.
\emph{Question:} map concrete governance settings to their minimum transferability
requirements. Sub-problems include:
\begin{itemize}
  \item \textbf{Which regimes need which grade.} Some settings can use
  designated-verifier assurance, such as a trusted national auditor reporting its
  conclusion. Others require publicly checkable artefacts. Examples include mutual
  verification between non-cooperating states. Another is public accountability
  when the regulator is not universally trusted.
  \item \textbf{Multi-auditor challenge randomness.} Study settings where a quorum
  of mutually distrusting parties supplies the challenge randomness. No single
  auditor could then steer or suppress an audit.
  \item \textbf{Dispute resolution.} If a transcript is contested, what evidence
  can a rejecting auditor show a third party without breaking zero-knowledge? A
  scheme whose failures cannot be demonstrated has limited governance value.
  Demonstrating failure is distinct from the accept-path guarantees for which the
  protocol was designed.
  \item \textbf{Retention obligations.} Transferability applies across time as well
  as across parties. An artefact may be checkable only while the trainer retains
  private material. Such an artefact is transferable in principle, but not in
  practice. Each regime therefore creates a storage obligation and a checking
  deadline.
  \item \textbf{Verifying the verifier.} When a treaty body lacks independent
  technical capacity, it delegates verification in practice even when the
  artefact is publicly checkable in principle. Whether that delegation can be made
  accountable, and by whom, determines the regime's real trust footprint.
  \item \textbf{Serving two regimes at once.} Can one deployment place a
  transferable, beacon-timed backstop over cheaper designated-verifier day-to-day
  audits? If so, what guarantee does the combination provide?
\end{itemize}
Over-specifying transferability buys
unaffordable machinery. Under-specifying it yields artefacts that convince nobody
the regime needed to convince.

\subsection{Network-anchor exploitation and compute accounting}

\opitem{anchor-exploitation}{Fuller exploitation of the network anchor.}
The network anchor is the protocol's only \emph{physically independent} witness.
Every other trust anchor is a trainer-side commitment. Yet the base protocol uses
the anchor for only three purposes:
\begin{itemize}
  \item reconciling sampled wire-bound tensors with the trainer's commitments
  (\Cref{appx:proof-cost:anchor-bindings});
  \item providing an aggregate collective count for the step-count check
  (\Cref{appx:proof-cost:defense-depth}); and
  \item providing the genesis-time traffic tag $h_{\mathrm{gen}}$ from
  \Cref{appendix:formalization}.
\end{itemize}
In fact, the anchor observes all inter-node traffic during the run. It sees
collective counts, sizes, endpoints, cadence, and wall-clock timing. Most of this
signal is currently discarded.
\emph{Question:} how much more verification power can the network anchor provide,
and at what privacy cost? Candidate directions include:
\begin{itemize}
  \item \textbf{Traffic-shape fingerprinting.} The (size, endpoint, cadence)
  profile of collectives fingerprints the executed parallelism topology and training
  regime. Can the profile continuously and cheaply certify the topology and
  per-step collective schedule declared in \texttt{arch\_spec}, without opening any
  payload?

  The current step-count argument instead compares aggregate collective counts. It
  cannot partition traffic into declared and undeclared collectives. A shape-level
  account of \emph{all} observed traffic would constrain undeclared co-located compute
  much more tightly.
  \item \textbf{Physical time-binding.} Anchor observations have wall-clock
  timestamps, binding the committed run to physical time. A training log fabricated
  later cannot have a contemporaneous anchor trace. The observations also lower-bound
  the cluster's elapsed occupancy. This provides an entry point to
  \opref{compute-accounting}.
  \item \textbf{The anchor as protocol infrastructure.} The anchor chain is an
  append-only, independently timestamped serialisation of the trainer's commitments.
  This is precisely the commit-before-release ordering and attempt-counting
  infrastructure required for beacon-based challenge derivation (\opref{beacons}).
  TAP-bound checkpoints are
  also natural candidates for the random exact anchors in hybrid protocols
  (\opref{hybrid}). Which minimal TAP upgrades unlock these roles while preserving
  Tier-1 auditability? Candidate upgrades are counting, timestamping, and signing.
  \item \textbf{On-path computation at Tier 2.} A programmable SmartNIC anchor can
  compute streaming payload statistics or sketches beyond raw hashing. Which cheap
  on-path computations provide the most verification power per unit of added trust
  in the device?
  \item \textbf{The privacy trade.} Each signal also leaks information. Traffic
  shape reveals layer counts and parallelism degrees. It may also reveal
  architectural dimensions that the private \texttt{arch\_spec} should protect.
  The problem includes quantifying the trade between leakage and verification. It
  also includes deciding which signals a zero-knowledge proof can consume instead
  of revealing them in the clear.
\end{itemize}
Every unchecked commitment is essentially useless.

\opitem{compute-accounting}{Compute accounting as a distinct certified object.}
The preceding machinery certifies that a computation followed a declared recipe.
Compute monitoring asks a coarser question: how much compute a facility used,
regardless of what it
computed~\citep{shavit2023chinchilla,petrie2025flexheg,petrie2025guaranteeable,baker2025verifying}.
This protocol may supply some accounting evidence as a by-product.

Counting compute needs no correctness proof, bit-exact determinism, or
\texttt{arch\_spec}. Whether it needs zero-knowledge depends on the regime. Total
training compute is commercially sensitive. A threshold may therefore require only
proof that a bound holds. Regulatory thresholds are ceilings, so the useful object
is an upper bound on the relevant compute.

The protocol already contains a simple primitive. Its step-count check compares the
anchor's observed collective count with the declared chain length. It uses no
cryptography (\Cref{appx:proof-cost:defense-depth}).

\emph{Question:} what compute quantity can the protocol certify, at which accounting
boundary, and with what error and trust assumptions? A useful certificate should
name the actor or facility, interval, covered hardware, included activities, and
direction of every error. Sub-problems include:
\begin{itemize}
  \item \textbf{Metering and bound direction.} Attested performance counters, power
  draw, and anchor-observed occupancy differ in granularity, tamper evidence, and
  trusted-silicon requirements. Observed activity usually lower-bounds the activity
  that occurred. Upper-bounding total activity also requires exhaustive coverage and
  either trusted counters or a justified maximum-rate model.

  \item \textbf{Attribution and subtraction.} Suppose policy defines training
  compute as total attributable compute minus attributable inference. Then
  \[
    C_{\mathrm{train}} \leq U_{\mathrm{total}}-L_{\mathrm{inference}}
  \]
  requires an upper bound on total compute and a lower bound on inference compute.
  Verifying some inference runs does not show that every excluded operation was
  inference. The accounting rule must classify shared infrastructure, synthetic-data
  generation, distillation, and unclassified work. Inference classification may
  still be tractable because it needs no backward pass, optimiser state, or
  cross-step commitment chain.

  \item \textbf{Coverage.} Metering bounds only \emph{metered} silicon. Undeclared or
  off-premises capacity lies outside that bound. This moves part of the problem from
  cryptography to supply-chain coverage. Export controls and chip registries may
  help, but they do not make coverage automatic.

  \item \textbf{Interpretation.} The certificate bounds compute within its declared
  boundary. It does not directly bound any model's capability. It covers undeclared
  preparation compute only when that work falls inside the boundary. It remains
  silent when a declared recipe produces more capability than policymakers expected
  (\Cref{appx:proof-cost:defense-depth} and \opref{reachability}).

  \item \textbf{Assurance and cost.} Must an actor-budget claim match the soundness
  of a recipe-conformance proof? A cheaper attestation may suffice for some decisions.
  The comparison must measure its full deployment cost and residual coverage risk.
\end{itemize}
A favourable result could reduce a hidden-compute channel within a stated coverage
boundary. Whether it closes that channel, or costs much less than recipe conformance,
remains to be shown. It would still be a different governance product.

\subsection{Hybrid ingredients and proving-cost levers}

The hybrid construction in \opref{hybrid} needs a cheap tolerance-band tier.
Every design option in \Cref{sec:proof-options} that wraps checks in cryptographic
proofs pays for the algebra used to express them. The ingredient-level
questions are \opref{float-native}--\opref{field-hardware}.
The co-design lever above them is \opref{codesign}. Progress on any one improves overall cost, whichever protocol
prevails.

\opitem{float-native}{Float-native approximate matmul verification on tensor cores.}
Proof-producing options in \Cref{sec:proof-options} spend a large share of their
proving cost translating floating-point arithmetic into proof-friendly algebra.
A randomised check could instead run in native floating point on the training
hardware's fast path (tensor cores). It would avoid that translation for the checks it covers.
Such a check is a candidate hybrid ingredient for \opref{hybrid}. Whether it fits
remains part of the open ``object honesty'' question below.

\emph{Question:} Design a randomised check for approximate floating-point matrix
products on reduced-precision tensor-core hardware. It must retain a sound error
analysis. Sub-problems include:
\begin{itemize}
  \item \textbf{Challenge representability.} Challenge values must be exactly
  representable in the reduced format. Otherwise, distinct challenges collapse
  to one machine value. This weakens per-challenge soundness and forces more
  repetitions. Sign-vector challenges $\{\pm 1\}$ are candidates because every
  format represents them exactly.

  \item \textbf{Semantics pinning.} Fast matrix units do not fully specify reduction
  order, rounding, or subnormal handling. The verified relation must take one of
  two routes. The relation
  can cover every admissible reduction order with an order-independent error
  envelope. That route requires a wider tolerance.
  Alternatively, it can pin the kernel in \texttt{arch\_spec}. Specifying an
  existing kernel's semantics belongs here. Changing the training kernel or
  datapath belongs to \opref{codesign}.

  \item \textbf{The tolerance/soundness tension.} A band wide enough for honest
  rounding also permits fraud of the same magnitude. In a random linear
  combination, a one-entry deviation contributes only a ${\sim}1/d$ fraction of
  the worst-case honest rounding-noise floor. Here, $d$ is the reduction length.
  Aggregate checks must account for this dilution.

  \item \textbf{Object honesty.} An approximate check certifies closeness in an
  aggregate norm. It does not certify the per-coordinate tolerance band that
  \opref{hybrid}'s weak tier binds. Can the aggregate guarantee be lifted to that
  tier? Alternatively, can the hybrid construction use aggregate weak objects?
  This is the same currency question raised by \opref{hybrid}.

  \item \textbf{Trust locus and binding.} A native check on the trainer's own
  hardware proves nothing by itself. The problem statement must name the runner
  and bind the operands and result to the commitments.
  Possible routes include independent auditor re-execution over authenticated
  openings, an attested device, or proof that committed values satisfy the check.
  Cost analysis must cover the chosen route, not only the tensor-core arithmetic.
\end{itemize}
Approximate sumcheck~\citep{bitan2025approxsumcheck} is a natural protocol-side
starting point. A broader co-design question appears open. Which verification
relations, if any, preserve both their challenge distribution and rounded
computation graph on native reduced-precision hardware?

\opitem{weak-ozaki}{Weakened, non-bit-exact slice decompositions (Ozaki-style).}
Exact Ozaki-style slice decompositions~\citep{ozaki2012errorfree} rewrite a
high-precision matrix product as a sum of exact small-integer slice products.
Each product is cheap to verify exactly in proof-friendly algebra. However, their
number grows quickly with the covered mantissa width and exponent range. A
weakened variant exactly binds only the leading products. It puts the tail in a
committed, range-checked residual window. This yields a tunable per-coordinate
error bound and is a concrete candidate for \opref{hybrid}'s tolerance-band tier.

\emph{Question:} Characterise the soundness and cost of truncated
slice-decomposition commitments. Sub-problems include:
\begin{itemize}
  \item \textbf{Window denomination.} The auditor must be able to pin the residual
  window's scale. For example, the scale could be the unit in the last place of a
  committed output. The prover must never choose the scale. Otherwise, scale
  inflation silently widens the window until it is vacuous.

  \item \textbf{Slice economics.} How few slice products can be bound before the
  window becomes uninformative? Can slice-count-efficient variants preserve
  soundness at an acceptable proof cost? One notable example is the
  modular/Chinese-remainder-theorem (CRT) based ``Ozaki
  Scheme~II''~\citep{ozaki2025scheme2}. Its slice count grows roughly linearly,
  rather than quadratically, with the precision covered. The proof cost must
  include checking the modular reconstruction itself.

  \item \textbf{Payload coupling.} Does narrowing the window actually reduce the
  adversary's end-of-run payload? Optimiser normalisation can amplify small
  per-step deviations. The window bounds each coordinate's deviation magnitude.
  Yet it leaves the number of adversary-chosen bits nearly unchanged. This is
  therefore part of \opref{hybrid}'s currency question, not only its
  drift-composition sub-problem.

  \item \textbf{A possible cost floor.} Full-coverage weak objects may require
  per-matmul work proportional to the output size. The per-coordinate width would
  then be the only free parameter. Is this floor a provable lower bound? Or is it
  an artefact of the constructions tried so far?
\end{itemize}
This is the list's most concrete candidate for a cheap full-coverage tier. Its
cost floor is also the closest to being settled either way.

\opitem{field-hardware}{Optimised kernels and hardware for finite-field operations.}
For every layer that wraps its checks in cryptographic proofs, proving throughput
is ultimately bounded by the cost of finite-field operations. Current hardware is
instead optimised for reduced-precision floating point. The resulting performance
gap has two factors.
The first is the number of field operations required per native FLOP\@. Work on the
verified relation attacks this factor through \opref{algebraic-reduction},
\opref{mac-constraints}, \opref{float-native}, and \opref{weak-ozaki}. The second
is the \emph{silicon-parity} factor between field arithmetic and native tensor-core
arithmetic on the same device. Reducing this second factor lowers the overhead of
every proof-producing instantiation of \Cref{sec:proof-options}.

\emph{Question:} How far can the silicon-parity factor be reduced on commodity and
dedicated hardware? Sub-problems include:
\begin{itemize}
  \item \textbf{Commodity kernels, and the limb floor.} Small proof fields
  (${\sim}31$-bit) can use integer tensor-core paths through limb decomposition.
  This method splits each field element across several machine-word \emph{limbs}.
  At a fixed soundness target, three choices must jointly provide enough soundness
  bits: the field, extension degree, and repetition count. This requirement sets a
  minimum limb count for the multiply datapath. The floor depends on the protocol
  family and its repetition/extension trade-offs. How closely can optimised
  kernels approach it?

  \item \textbf{Circumventing the floor by field choice.} Field selection might
  bypass the floor without kernel engineering. One candidate is a
  small-characteristic prime matched to native significand widths. Another is a
  regular-representation encoding. There, field multiplication becomes a dense
  small matrix product that matrix units can execute. Binary-field routes may
  instead use native instructions such as carryless multiply and Intel GFNI\@.
  Binius~\citep{diamond2025binius} is the developed instance of this approach.
  It has end-to-end pricing against CPU provers. On a Keccak-$f$ proving task, the
  result is roughly two- to fourfold over Plonky3. However, to our knowledge,
  neither Binius nor the other candidates have been compared against native
  tensor-core arithmetic's silicon-parity baseline.

  \item \textbf{The non-matmul fraction.} Only the matmul-shaped share of a prover
  benefits from tensor mapping. Hashing and bandwidth-bound streaming remain on
  the general-purpose pipeline. So does converting non-algebraic operations such
  as rounding into algebraic constraints. Can these parts be restructured for
  dense-silicon-friendly execution? Or do they structurally cap whole-prover
  speedup?

  \item \textbf{Dedicated silicon.} A proving accelerator could combine native
  small-field multiply-accumulate at tensor-core density with fused hash and
  transcript pipelines. What would it cost to build and deploy? What timeline and
  fleet-level overhead would it require? Prover hardware requires no attestation
  because soundness here is cryptographic. Its trust profile is therefore
  categorically lighter than TEE-based verification hardware.

  \item \textbf{Benchmarking discipline.} Kernel-level speedups do not automatically
  translate into end-to-end prover throughput. Computing that throughput is not
  necessarily easy. The community also lacks agreed whole-prover acceptance
  benchmarks. Such benchmarks cover time, energy, area, and cost against pinned
  baselines. Without them, overhead claims cannot be meaningfully compared.
\end{itemize}

\opitem{codesign}{Verification-aware model and datapath co-design.}
For a fixed proof backend, matrix-product verification cost may depend on tensor
shape and kernel materialisation. A candidate cost model separates boundary-sized
work, $mk+kn+mn$, from MAC-volume work, $mkn$. The balance is specific to the
backend; shape is not an intrinsic verification property.

Small attention-head dimensions provide a controlled test. For a per-head product
with sequence length $L$ and head dimension $h$, the boundary-to-volume ratio is
$2/L+1/h$. Changing $h$ and the number of heads can preserve model width and
attention MAC volume.

\emph{Question:} can this controlled change reduce whole-step verification cost
without a material capability loss? First, pin one proof backend, hardware stack,
training step, and kernel layout. Fit or derive its shape-sensitive cost model, then
test that model on held-out tensor shapes.

Next, train matched variants that change only the permitted head configuration.
Hold model width, parameter budget, training tokens, data, optimiser, and target
native FLOPs fixed. Measure whole-step prover time and energy, native training
slowdown, and a predeclared capability suite.

Use the declared architecture as the baseline. Success requires a Pareto improvement
in the predeclared whole-step cost metrics within the capability margin. The proposal
fails for this design space if the cost model does not generalise or no variant
improves the measured frontier. Datapath changes can use the same discipline, but
should be tested separately from model changes.

Only after such a result exists does the policy question become concrete: whether a
governance regime should require or incentivise the verified design.

\opitem{surrogate}{Surrogate algorithms and formal proofs for error bounds.}
Many zero-knowledge inference protocols use integer rather than floating-point
arithmetic. This better suits finite-field protocols, but it introduces approximation
error relative to mainstream floating-point implementations. The same idea could
produce training algorithms that are cheaper to prove.

A concrete programme should begin with one pinned reference transition and one
proof-friendly surrogate. Both must include the optimiser state and numerical
semantics. A machine-checked theorem should bound their difference over a declared
training horizon, norm, and certified state region. Any claimed endpoint must have
its own link from state error to the relevant loss or model property.

The error threshold must be useful rather than merely finite. A candidate succeeds
only if its certified bound stays below that threshold and its complete proof costs
less than proving the pinned reference transition. The comparison should include
range checks, stability certificates, native slowdown, and a predeclared capability
margin. Polynomial replacements for nonlinear operations, such as softmax, are one
candidate rather than the programme itself.

The theorem must also apply to the executed run. A training certificate should bind
the implementation, batches, parameters, optimiser state, and numerical semantics to
the formal transition. It should also bind any range or stability certificates that
discharge the theorem's hypotheses. Without this link, a correct theorem can remain
irrelevant to the certified execution.

Lean or Rocq can machine-check the approximation and composition arguments. The
formal statement should expose all permitted assumptions and dependencies. Useful
approximation libraries may shorten these proofs and standardise their reviewed
components.

Proof privacy is a separate, optional layer. zkPi demonstrates a zkSNARK for
possession of a Lean proof of a public statement without revealing the
proof~\citep{laufer2024zkpi}. It does not establish that the error bound is useful,
that its hypotheses hold, or that execution matches the formal model.

Machine checking still inherits bugs in the proof assistant's kernel, elaborator,
parser, libraries, and surrounding checker. Independent scripts, human review, and
LLM-assisted review can inspect proof terms, statements, assumptions, imports, and
dependency versions. They may expose suspicious attempts to exploit such bugs or
hide specification gaps. These checks are heuristic defense in depth. They add no
formal or cryptographic guarantee and do not replace machine checking. To our
knowledge, the suggestion to use LLM auditors is due to Luke Marks.

This review tooling could also support inference verification. Its value should be
measured through bug-finding tests, including seeded proof and specification errors.

\section{The zkVM instantiation: details}\label{app:zkvm-details}

This appendix details the zkVM instantiation of the proof layer --- the reference
design of this paper (\Cref{sec:proof-options} compares it against bespoke-SNARK
and beacon-based interactive alternatives). It expands the condensed description of
\Cref{architecture}: the zkVM requirements and the hint-and-verify cost model, then
the full precompile catalogue.

\subsection{The zero-knowledge virtual machine}\label{zkVM}

The zkVM produces a succinct, non-interactive proof of correct execution of a guest
program. We require three capabilities: application-defined precompiles (native
constraint circuits callable from the guest), GPU-accelerated proof generation, and
continuations with recursive composition\footnote{We use
\href{https://risczero.com/}{\texttt{RISC Zero}} as our reference implementation;
the architecture is not coupled to this choice.}. The zkVM operates on a
\emph{hint-and-verify} model: the host (trainer's GPU) runs the full computation
outside the proof and provides intermediate values as hints; the guest (proof
checker inside the zkVM) does not re-execute, but verifies hints by calling
precompiles against Merkle-committed values. Host computation contributes zero
constraints; only the guest's verification logic does. This reduces cost by roughly
five orders of magnitude versus full re-execution.\footnote{Reduction factor
$(m\cdot n / k)\cdot(c_{\mathrm{sw}}/c_{\mathrm{pre}})$, with
$c_{\mathrm{sw}}\approx 1{,}000$ RISC-V cycles per software-emulated MAC and
$c_{\mathrm{pre}}\approx 90$ constraints per native MAC; typically
$10^{4}$--$10^{6}$.} Hint-based witness computation is standard in ZK-ML inference
systems~\citep{kang2022scaling, chen2024zkml, sun2024zkllm}, and proof-of-training
constructions exist for smaller model classes~\citep{garg2023zktraining,abbaszadeh2024kaizen}.
The design extends the same pattern to the backward pass:
transposed GEMMs via the MAC precompile, activation derivatives via the BF16 lookup
(BF16 $\mathrm{GELU}'(x)$ is also a $2^{16}$-entry table), and Adam's sqrt and
division via FP32 nonlinear precompiles --- the step that would make verifying
\emph{training}, not just inference, possible, and which this paper specifies but
does not yet demonstrate (\opref{zk-backprop}). We provide a complete walkthrough in
\Cref{appendix:toy-example}.

\subsection{The precompiles}\label{precompiles}
Precompiles are native constraint circuits hardwired into the zkVM's proof system. They
implement specific numerical operations as compact constraint sets over the zkVM's native
field $\mathbb{F}_p$. This approach differs subtly from prior ZK-ML
(cf.~\Cref{tab:related-work}). Prior work converts the \emph{model's computation} to
finite-field arithmetic, executes GEMMs over $\mathbb{F}_p$ in fixed-point, and thereby
proves a different computation from the one performed by the GPU. Our precompiles take
the opposite approach: the GPU executes actual BF16/FP32, while the proof system verifies
\emph{individual operations} by encoding IEEE~754 semantics as $\mathbb{F}_p$
constraints. Each floating-point value is decomposed into sign, exponent, and mantissa
(integers in $\mathbb{F}_p$). Constraints then enforce that the integer relationships
among these components match the IEEE~754 specification for the operation
(multiplication, accumulation, rounding). The proof system runs over
$\mathbb{F}_p$ internally but proves a statement about the floating-point computation
that actually ran on the GPU. Eight precompiles cover a Transformer training step:

\begin{table}[h!]
  \centering
  \small
  \begin{tabular}{clll}
    \toprule
    \# & Precompile & Constraints/op & Used for \\
    \midrule
    1 & BF16/FP32 MAC chain & $\sim$90 & GEMM dot products \\
    2 & BF16 lookup table & $\sim$15 & GELU, SiLU ($2^{16}$ entries) \\
    3 & Merkle path (zkVM-native hash)\footnotemark & $\sim$1{,}500 & Binding hints to commitments \\
    4 & FP32 exp & $\sim$250 & Softmax, cross-entropy \\
    5 & FP32 sqrt/rsqrt & $\sim$200 & RMSNorm, Adam \\
    6 & FP32 div & $\sim$200 & Normalisation, Adam \\
    7 & FP32 log & $\sim$250 & Cross-entropy loss \\
    8 & SHA-256 compression & $\sim$7{,}500 & Network-anchor reconciliation \\
    \bottomrule
  \end{tabular}
  \caption{Precompile catalogue. The MAC chain (1) dominates: ${\approx}98\%$ of constraints
  at Llama~3.1 405B scale. Merkle path (3) is invoked per sampled value, $<0.1\%$
  of budget. SHA-256 (8) is used only when opening paths against the network anchor
  (\Cref{network-tap}). The ${\sim}90$-constraint MAC figure is a decomposition-based
  estimate carried with a $50$--$150$ implementation envelope
  (\Cref{appx:proof-cost:model}); absolute proof-time estimates inherit this
  uncertainty. Full breakdown in \Cref{appendix:proof-cost}.}
  \label{tab:precompiles}
\end{table}
\footnotetext{Concretely Poseidon today; any zkVM-friendly hash with low per-call
constraint cost (Poseidon2, Reinforced Concrete) is a drop-in substitute.}
The dominant precompile is the \textbf{BF16/FP32 MAC chain (precompile 1)}, accounting for ${\approx}98\%$ of constraints at 405B scale. It verifies a single
multiply-accumulate: BF16 multiply producing an FP32 intermediate, accumulated into
an FP32 running sum. The circuit decomposes each BF16 input into
sign/exponent/mantissa, verifies the multiplication via integer arithmetic on the
mantissa, and checks FP32 accumulation under IEEE~754 round-to-nearest-even. Each MAC
costs ${\sim}90$ constraints; a dot product of dimension $d$ requires $d$ sequential
MACs. {\bf Precompile 1} is used for all GEMM verification in forward and backward passes.
We use \emph{interactive per-entry sampling} for GEMM verification, involving {\bf precompile 1}\footnote{Exact verification for GEMM is costly; we make use of randomness and interaction to achieve efficiency, at the cost of non-perfect soundness. See~\Cref{appx:proof-cost:gemm-baseline} for further discussion.}: the
host hints the full result, the auditor selects $k$ random output entries, and the
proof checker recomputes each via the MAC precompile, binding inputs and output to
their committed roots. The detection probabilities achieved by this sampling, as a
function of $k$ and of the deviation fraction, are quantified in
\Cref{sec:sampling-guarantees}. Approximate
sumcheck~\citep{bitan2025approxsumcheck} suggests a path to recovering Freivalds-like
efficiency for FP GEMMs (\Cref{sec:future-work}).
\textbf{Precompile 2} uses a precomputed $2^{16}$-entry exhaustive table for BF16 activations
(GELU, SiLU, derivatives) at ${\sim}15$ constraints per element. {\bf Precompile 3}
verifies Merkle-path membership (Poseidon, ${\sim}1{,}500$ constraints per depth-30
path); every hinted value passes a path check to bind it to the committed root.
{\bf Precompiles 4--7} encode FP32 nonlinearities (exp, log, sqrt/rsqrt, div) used sparsely
in normalisation, softmax, optimiser, and loss, at 200--250 constraints per call
($<1\%$ of total). {\bf Precompile 8} (SHA-256, ${\sim}7{,}500$ constraints/block) is used
exclusively for network-anchor reconciliation (\Cref{network-tap}). The rationale for the
Poseidon~+~SHA-256 dual-hash design is in \Cref{appendix:proof-cost} and
\Cref{app:network-anchor}.

\section{Proof cost and overhead estimation}\label{appendix:proof-cost}

This appendix supports the quantitative claims in \Cref{sec:proposed-solution} with detailed derivations. It covers (i) the zkVM cost model and per-operation constraint counts, (ii) why exact per-entry GEMM verification is infeasible over floats and why Freivalds' algorithm does not directly apply, (iii) the per-layer and per-step proof budget at Llama~3.1~405B scale under the commitment and fusion assumptions used in \Cref{architecture}, (iv) the statistical properties of interactive sampling, (v) the MoE cost story, (vi) the three proof flows (genesis, in-training, ex-ante) with concrete per-flow costs, and (vii) a training-side overhead summary. Caveats and uncertainties are at the end.

\subsection{Cost model}\label{appx:proof-cost:model}

\subsubsection{Mechanics: hint-and-verify}

The zkVM operates as a verification machine rather than an execution machine. For each operation in the training step, the host (trainer's GPU) computes the result and provides it as an unchecked hint. The guest (proof checker running inside the zkVM) verifies the hint against committed Merkle roots and against declared operations via native constraint circuits called precompiles. Host execution is free from the proof's perspective; proving cost is determined entirely by the number of constraints the precompiles generate.

\subsubsection{Per-operation constraint counts}

\Cref{tab:proof-cost-precompiles-detail} refines the precompile catalogue of \Cref{tab:precompiles} to the primitive level: it decomposes the per-path hashing costs into per-invocation primitive costs and states the tree-depth envelope behind the per-path totals.

\begin{table}[H]
  \centering
  \small
  \resizebox{\textwidth}{!}{%
  \begin{tabular}{clll}
    \toprule
    \# & Primitive cost & Per-invocation & Used for \\
    \midrule
    1 & BF16/FP32 MAC & $\sim$90 constraints & GEMM dot products (one per sampled entry times inner dim) \\
    2 & BF16 table lookup & $\sim$15 constraints & Activation functions (GELU, SiLU, their derivatives) \\
    3 & zkVM-native hash compression & $\sim$50 constraints & Merkle-path verification ($\sim$1{,}500 / path at depth 30) \\
    4 & FP32 exp & $\sim$250 constraints & Softmax, cross-entropy loss \\
    5 & FP32 sqrt/rsqrt & $\sim$200 constraints & RMSNorm, Adam \\
    6 & FP32 div & $\sim$200 constraints & Normalisation, Adam \\
    7 & FP32 log & $\sim$250 constraints & Cross-entropy loss \\
    8 & SHA-256 compression & $\sim$7{,}500 constraints & Anchor-path verification ($\sim$150{,}000 / path at depth 20) \\
    \bottomrule
  \end{tabular}%
  }
  \caption{Per-invocation constraint counts for the eight precompiles. The dominant primitive is the BF16/FP32 MAC; Merkle-path verifications are an additive per-sample cost; SHA-256 is invoked only for wire-bound sample reconciliation. Per-path totals assume a tree depth matched to the payload size (zkVM-native Merkle for tensors up to $\sim$$10^9$ elements at depth 30; SHA-256 anchor paths at depth 20, a conservative envelope covering aggregated anchor trees up to ${\sim}1$\,TB at 1\,MB leaves (a single 1\,GB payload alone needs binary depth 10)).}
  \label{tab:proof-cost-precompiles-detail}
\end{table}

The per-MAC count of $\sim$90 is an estimate based on the IEEE~754 integer decomposition: BF16 multiply ($\sim$30--50 constraints for mantissa product and exponent addition), FP32 accumulation with RNE rounding ($\sim$40--60 constraints for exponent alignment, mantissa addition, guard/round/sticky bits). The decomposition spans 70--110; we carry the deliberately wider 50--150 implementation envelope, which affects absolute proof-time numbers by up to ${\sim}2\times$ but not comparative analysis between approaches; a direct measurement of the actual count is part of the planned empirical validation work.

\subsubsection{Proving throughput}

The cluster-side proving throughput assumed throughout is ${\sim}10^6$ constraints per second per GPU (RISC Zero-class benchmark on A100/H100 for constraint-heavy guests), i.e.\ ${\sim}10^9$ constraints per second aggregate on a 1{,}024-GPU cluster. Segment size is $2^{20}$ cycles, segment-level proving is embarrassingly parallel, and recursive composition folds per-segment proofs into a constant-size aggregate ($\sim$200\,KB) independent of the underlying circuit size.

\subsection{Exact per-entry GEMM verification is O(n$^3$); Freivalds does not directly apply}\label{appx:proof-cost:gemm-baseline}
Exact verification of an $m\times d\times n$ GEMM at ${\sim}90$ constraints/MAC costs
$O(mnd)$, exceeding current proof systems by several orders of magnitude at 405B
scale. Freivalds' algorithm reduces this to $O(mn+md+nd)$ but does not directly apply to
BF16/FP32 since floating-point addition is \emph{not} associative
(\Cref{appx:proof-cost:freivalds-fails}). 
\subsubsection{Why the na\"ive verification is cubic}

A GEMM of dimensions $m \times d \times n$ computes $C = A \cdot B$ where each entry $C_{ij}$ is a dot product of length $d$ of BF16 inputs accumulated in FP32 with IEEE~754 RNE rounding. Verifying every entry exactly requires $mn$ dot products, each consisting of $d$ MACs, yielding $mnd$ MAC verifications. At $\sim$90 constraints per MAC this is $\sim$$90mnd$ constraints. For a single Q-projection GEMM of Llama~3.1 405B (taking $m=2{,}048$, $d=n=16{,}384$), the unsampled cost is $\sim$$5 \times 10^{13}$ constraints, which exceeds the capacity of current proof systems.

\subsubsection{Why Freivalds' algorithm does not directly reduce this for floats}\label{appx:proof-cost:freivalds-fails}

Freivalds' algorithm verifies $C = A \cdot B$ by drawing a random vector $r \in \mathbb{F}_p^n$ and checking $A(Br) = Cr$. Over an exact field this reduces verification from $O(n^3)$ to $O(n^2)$ with error probability $\leq 1/|\mathbb{F}_p|$. For native BF16/FP32 arithmetic the algorithm fails, because floating-point addition is not associative. The two computation paths $A(Br)$ and $Cr$ involve different accumulation orders and therefore different rounding sequences, producing different bit-exact outputs even when $C$ is the correct GPU output. The check rejects correct computations.

Converting floats to exact integers and then applying Freivalds over a large field is theoretically available. For example, one could scale BF16 mantissa-exponent pairs to a common exponent and pad them with zeros to a fixed bit-width. This approach faces three obstacles. First, a naive common-exponent encoding produces integers that exceed the trainer's native Baby Bear field ($\sim$$2^{31}$), requiring multi-limb arithmetic with ${\sim}121\times$ overhead per operation. Error-free slice decompositions~\citep{ozaki2012errorfree}, particularly the integer-modular variant of Ozaki Scheme~II~\citep{ozaki2025scheme2}, avoid this wide-integer route. They instead require a slice count that grows with the precision covered and verification of the reconstruction (\opref{weak-ozaki}). Second, the exact integer product does not equal the GPU's rounded output, so an additional $O(n^2)$ rounding-verification pass is needed. Third, the interaction between exact-integer Freivalds and the FP32 accumulation rounding chain that determines the GPU's actual output remains an open research question. Recent theoretical work on sumcheck protocols that tolerate bounded approximation error~\citep{bitan2025approxsumcheck} points to a path for recovering Freivalds-like asymptotics over floats; we discuss this as an open problem in \opref{algebraic-reduction}.

In the remainder of this appendix the baseline assumption is interactive per-element sampling; any future reduction is an optimisation on top.

\subsection{Proof cost per microbatch at Llama~3.1 405B}\label{appx:proof-cost:per-step}

\subsubsection{Commitments and fusion assumptions}\label{appx:proof-cost:commitments}

Per \Cref{constant-hashing}, the trainer commits six Merkle roots per layer during training: layer input, $Q$, $K$, $V$, attention output, and layer output. What is \emph{not} committed during training has cost consequences, so we state this explicitly.

\begin{itemize}
  \item \textbf{Flash Attention internals are truly fused.} The attention score matrix ($B \times \text{heads} \times s \times s$) and the softmax output never materialise in GPU memory; they are streamed through shared memory inside the Flash Attention kernel. These are uncommittable by kernel construction, not by choice. When the auditor samples attention-output entries, the host recomputes each sampled entry by running the full per-query attention over the sequence dimension. Per-sample cost is $\Theta(s \cdot d_{\mathrm{head}})$ MACs, which is materially larger than a single GEMM dot product. To keep the attention contribution bounded, we use a reduced sample count $k_{\mathrm{attn}} = 500$ for attention outputs, relying on the committed boundary roots (Q, K, V, attention output) for the primary consistency guarantee.

  \item \textbf{FFN-block internals are not truly fused.} In Llama-style SwiGLU blocks the gate and up projections materialise in HBM as separate tensors before the elementwise activation and multiply. We exploit this by committing $\mathrm{gate\_out}$ and $\mathrm{up\_out}$ Merkle roots \emph{at challenge time} (rather than during training) when the trainer re-runs the step. The training-time hashing overhead therefore remains at six roots per layer, while the per-challenge proof samples each FFN GEMM independently at the cheaper per-GEMM rate. This is a deliberate design trade: training-time hashing stays cheap, challenge-time response grows by the time to hash two FFN intermediates (about 436\,MB per layer for one 2{,}048-token microbatch, a few milliseconds at the hashing rate of \Cref{appx:proof-cost:hashing}), and the per-sample cost of FFN verification drops by a factor of ${\sim}d_{\mathrm{ff}}$. Because these roots are created only at challenge time, the entry-sampling soundness argument of \Cref{sec:sampling-guarantees} requires a \emph{nested} challenge ordering for them: the step and tensor identifiers are revealed first, the trainer publishes the lazy intermediate roots, and only then is the fresh, domain-separated entry-sampling randomness for those tensors drawn and revealed. Without this second commit-then-reveal round, a trainer that already knows the sampled entry positions could construct challenge-time roots that are correct exactly on the sampled entries, voiding the $(1-f)^k$ bound for the lazy intermediates.
\end{itemize}

\subsubsection{Per-layer constraint budget}\label{appx:proof-cost:per-layer-budget}

We compute the per-layer forward + backward cost at Llama~3.1 405B scale (d$_{\mathrm{model}}=16{,}384$, d$_{\mathrm{ff}}=53{,}248$, $n_{\mathrm{heads}}=128$, $n_{\mathrm{kv\_heads}}=8$ with grouped-query attention (GQA), $d_{\mathrm{head}}=128$, 126 layers, sequence length $s=2{,}048$ per microbatch, $k=4{,}605$ samples per committed GEMM, $k_{\mathrm{attn}}=500$ for attention outputs).

\emph{Scope.} Every figure in this subsection prices one 2{,}048-token microbatch on one
model replica, with one sampled set of entries per committed GEMM. A training step at this
scale processes far more tokens (the Llama~3.1 405B global batch grows from 4M to 16M
tokens), so a step contains thousands of such microbatches. What a full step costs to
prove is discussed, and left open, in \Cref{appx:proof-cost:full-step}.

Each forward GEMM of shape $(m \times d) \times (d \times n)$ induces two backward GEMMs (input gradient and weight gradient) whose inner dimensions are $n$ and $m$ respectively. Per sampled output entry, the proof cost is inner-dimension $\times 90$ constraints.

\begin{table}[H]
  \centering
  \small
  \resizebox{\textwidth}{!}{%
  \begin{tabular}{lrrr}
    \toprule
    Per-layer component & Inner-dim sum (fwd + 2 bwd) & Sampled cost (k = 4{,}605) & Layer contribution \\
    \midrule
    Q projection & $34{,}816$ & $90 \cdot k \cdot 34{,}816$ & $1.44 \times 10^{10}$ \\
    K projection (GQA, smaller n) & $19{,}456$ & $90 \cdot k \cdot 19{,}456$ & $8.06 \times 10^{9}$ \\
    V projection (GQA) & $19{,}456$ & $90 \cdot k \cdot 19{,}456$ & $8.06 \times 10^{9}$ \\
    O projection & $34{,}816$ & $90 \cdot k \cdot 34{,}816$ & $1.44 \times 10^{10}$ \\
    FFN gate projection & $71{,}680$ & $90 \cdot k \cdot 71{,}680$ & $2.97 \times 10^{10}$ \\
    FFN up projection & $71{,}680$ & $90 \cdot k \cdot 71{,}680$ & $2.97 \times 10^{10}$ \\
    FFN down projection & $71{,}680$ & $90 \cdot k \cdot 71{,}680$ & $2.97 \times 10^{10}$ \\
    Flash Attention (fused, $k_{\mathrm{attn}} = 500$) & $c_{\mathrm{attn}} \cdot s \cdot d_{\mathrm{head}}$ per sample, $c_{\mathrm{attn}} {\approx} 4$ & end-to-end recompute & $\sim$$5 \times 10^{10}$ \\
    Elementwise SiLU + multiply (per FFN sample) & $d_{\mathrm{ff}}$ ops $\times$ (15 + 90) & lookup + BF16 mult & $\sim$$2.6 \times 10^{10}$ \\
    Merkle paths (6 trainer-side per sample) & depth 30, zkVM-native hash & $1{,}500$ constraints each & $\sim$$4 \times 10^{7}$ \\
    \midrule
    Per-layer total (fwd + bwd) & & & $\sim$$2.1 \times 10^{11}$ \\
    \bottomrule
  \end{tabular}%
  }
  \caption{Per-layer forward + backward constraint budget at Llama~3.1 405B scale under the commitment and fusion assumptions of \Cref{appx:proof-cost:commitments}. Query/key/value/output (Q/K/V/O) and feed-forward-network (FFN) GEMMs are sampled per-GEMM (trainer-side Merkle roots at layer boundaries and lazy intermediates bind the hints). Flash Attention uses reduced-sample recomputation; elementwise SiLU and multiply contribute per-FFN-sample. SHA-256 anchor-path verification (precompile~8) is zero here because no wire-bound tensors are verified per layer on an intra-node path; it appears only on sampled wire-bound tensors (\Cref{appx:proof-cost:anchor-bindings}). The Merkle-path row assumes six openings per sample; the leaf layout that would make six sufficient is not yet specified, so that row is a lower bound.}
  \label{tab:proof-cost-per-layer}
\end{table}

\subsubsection{Total per microbatch}

Summing 126 layers contributes $126 \cdot 2.1 \times 10^{11} \approx 2.6 \times 10^{13}$ constraints. The LM-head contribution is bounded above by a single GEMM with inner-dim sum $d_{\mathrm{model}} + \text{vocab} + m \approx 146{,}688$, giving $\sim$$6 \times 10^{10}$ constraints (less than 0.3\% of the total).

\textbf{Proof budget per microbatch: $\sim$$2.6 \times 10^{13}$ constraints.} The MAC precompile accounts for roughly 98\% of this budget; Merkle-path verification (precompile~3) contributes well under 0.1\%. SHA-256 anchor-path verification (precompile~8) is invoked only for wire-bound samples, adding a bounded amount per challenged wire-bound tensor (see \Cref{appx:proof-cost:anchor-bindings}).

\subsubsection{Proving time}

At a cluster-wide throughput of ${\sim}10^9$ constraints per second (1{,}024 GPUs at ${\sim}10^6$ constraints per second each), a proof of one microbatch through all 126 layers of Llama~3.1 405B, forward and backward, completes in ${\sim}7.4$ hours. Linear scaling of compute yields $\sim$110 minutes on 4{,}096 GPUs, or $\sim$28 minutes on 16{,}384 GPUs. Per-layer proofs are independent and compose via recursive STARK folding, so the critical path is the most expensive individual layer proof rather than the serial sum, which becomes relevant when proving is parallelised across dedicated proof clusters.

\subsubsection{Comparison with the O(n$^3$) baseline}

Summed over all per-layer GEMMs, the unsampled baseline at Llama~3.1 405B yields ${\sim}2.2 \times 10^{17}$ constraints per microbatch (roughly $10^{4}$-fold the sampled cost; equivalently, $90$ constraints per MAC over the $3N$ MACs per token of a fwd+bwd pass at $N \approx 4\times10^{11}$ parameters and $2{,}048$ tokens). The ${\sim}10^{4}\times$ reduction from interactive sampling at $k=4{,}605$ is what makes proof generation feasible at all at this scale: the unsampled figure would take ${\sim}7$ years per microbatch at the cluster throughput above.

\subsubsection{What these figures do not price: a full training step}\label{appx:proof-cost:full-step}

The figures above do not give the cost of proving a full training step. The gap comes from
the weight-gradient GEMMs. Their inner dimension is the number of tokens summed into the
gradient. Within one microbatch that is 2{,}048, as tabulated. The gradient used by the
optimiser, however, sums over every token of the step. The protocol can commit it in one of
two ways, and it does not yet specify which:
\begin{itemize}
  \item \textbf{Accumulated gradient.} If only the step's accumulated gradient is
  committed, each sampled weight-gradient entry is a dot product over all of the step's
  tokens. At 16M tokens, one sampled set of weight-gradient entries for all 126 layers
  costs about $5.9\times10^{15}$ constraints in multiply-accumulate work alone. That is
  roughly $1.6\times10^{6}$ GPU-hours at the throughput above, or more than 200 times the
  microbatch figure.
  \item \textbf{Per-microbatch gradients.} If each microbatch's partial gradient is
  committed, the figures above remain valid per microbatch. The accumulation across
  microbatches and the all-reduce across replicas then become further objects to commit
  and check. Neither is priced here.
\end{itemize}
Both designs also need the optimiser transition and the reduction order to be specified
and committed (\opref{zk-backprop}). Until the protocol fixes these choices, we give no
per-step or per-run proving cost. The per-microbatch figures are the primary quantities;
any aggregate built from them must state which commitment design it assumes.

\subsection{Interactive sampling: statistical properties}\label{appx:proof-cost:sampling}

\subsubsection{Sample count for target security}

For auditor-chosen samples drawn after Merkle roots have been committed (no grinding), the probability of missing a deviation affecting fraction $f$ of entries is $P_{\mathrm{miss}} = (1 - f)^k$. Solving for target miss probability $\tau$:
\[
  k \geq \frac{\ln(1/\tau)}{-\ln(1-f)}.
\]
Concrete sample counts for target miss probabilities $\tau \in \{10^{-3}, 10^{-6}, 10^{-10}, 10^{-20}, 2^{-128}\}$ and deviation fractions $f \in \{10\%, 1\%, 0.1\%\}$ are tabulated in the main text (\Cref{tab:sample-count} in \Cref{sec:sampling-guarantees}, which also states the step-sampling and end-to-end detection guarantees and their limits); the recommended GEMM default $k = 4{,}605$ achieves $10^{-20}$ miss probability per committed GEMM output at a $1\%$ deviation fraction (fused-attention outputs use the reduced budget $k_{\mathrm{attn}}=500$).

The default $f = 1\%$ is a tuning choice rather than a fundamental threshold. Sampling detects deviations affecting a fraction $f$ of entries with probability $1 - (1-f)^k$, so any target miss probability is reachable by scaling $k$: smaller $f$ simply requires more samples (\Cref{eq:entry-budget}). The default $k = 4{,}605$ corresponds to $10^{-20}$ per-GEMM-output miss probability at $f = 1\%$; the same guarantee at $f = 0.1\%$ requires $k = 46{,}029$. For wire-bound tensors, the network anchor (\Cref{network-tap}) provides an independent commitment that the proof binds against at every sampled entry; a mismatch between the trainer's tensor commitment and the anchor's wire commitment surfaces as a failed anchor-consistency check inside the zkVM proof. Formal sensitivity analysis bounding the smallest $f$ a rational adversary can exploit is an open problem (\opref{algebraic-reduction}).

\subsubsection{Network-anchor bindings}\label{appx:proof-cost:anchor-bindings}

Wire-bound tensors are committed twice: the trainer holds a zkVM-native Merkle root of the tensor value, and the network anchor holds an SHA-256 Merkle root of the wire payload that transported that tensor. When the auditor samples a wire-bound entry, the proof opens both paths and verifies that the element value is consistent with both commitments. Per sampled wire-bound element, the anchor consistency check costs $\mathrm{depth} \cdot \text{SHA-256}_{\mathrm{compress}} + \lceil c / 64 \rceil \cdot \text{SHA-256}_{\mathrm{compress}}$ for a chunk of $c$ bytes ($\sim$2--8 million constraints at $c = 16$--$64$\,KB leaves per \Cref{app:network-anchor}). Anchor checks are invoked only for sampled wire-bound tensors. Their total therefore depends on how many wire-bound tensors a challenge samples, which the challenge mix must state. Each sampled wire-bound tensor with $k=4{,}605$ openings adds about $10^{10}$--$4\times10^{10}$ constraints.

\subsubsection{Defense-in-depth}\label{appx:proof-cost:defense-depth}

The general-purpose security argument against an adaptive adversary composes three layers:

\begin{enumerate}
  \item \textbf{Network anchor.} Catches any deviation that changes wire-bound content, by a deterministic (not statistical) comparison on any sampled wire-bound entry --- given the wire-to-tensor reconciliation of \opref{wire-to-tensor}; selecting the entry remains statistical.
  \item \textbf{On-the-fly Merkle commitments.} Each layer's output is committed during training. Any error in any intermediate that propagates to the layer output changes the committed root relative to an honest recomputation; a challenge at that boundary then exposes it (an internally consistent fake chain is instead caught by the operation-level checks of layer 3).
  \item \textbf{Interactive sampling.} Adds $10^{-20}$ statistical assurance on intermediate GEMMs for deviations affecting $\geq 1\%$ of entries.
\end{enumerate}
Each layer is necessary: (1) alone cannot see intra-node computation; (2) alone does not rule out an adversary who produces internally consistent fake commitments; (3) alone has a detection floor at the deviation fraction. Together they provide complementary coverage.

\paragraph{Compute-threshold attestation: case-by-case analysis.}
The three-layer argument above is the general case against an adversary cheating on arbitrary aspects of the training procedure. For the specific target of compute-threshold attestation (ex-ante binding of total FLOPs, \Cref{ex-ante-attestation}), the adversary's options are narrower. The rational adversary under this target wants to \emph{under}-declare compute: to run more than was claimed in order to stay below a regulatory threshold. The following cheating modes each admit a cheap and direct detection; a further mode, which does not, is discussed after them.

\begin{itemize}
  \item \textbf{Model larger than declared (wider).} The adversary inflates $d_{\mathrm{model}}$, $d_{\mathrm{ff}}$, $n_{\mathrm{heads}}$, or another dimension while publishing $\texttt{arch\_spec}$ for smaller values. The commitment $R(W_0)$ is a Merkle tree whose leaves and depth are determined by the actual weight shapes. An opening at any challenged weight position in the genesis proof must produce a path matching the $\texttt{arch\_spec}$-declared shape. A wider actual weight matrix has leaves at different offsets; the path opening fails. Detection requires a single genesis weight-opening.

  \item \textbf{Model larger than declared (deeper).} The adversary runs $L_{\mathrm{actual}} > L_{\mathrm{declared}}$ layers. Three consequences: (i) the committed per-layer chain terminates at $L_{\mathrm{declared}}$, so the input to the committed LM head is the declared-depth activation, not the actually-deeper activation from which the loss was computed; (ii) in the backward pass, the gradient committed at declared layer $L_{\mathrm{declared}}$ does not equal $\partial L / \partial \mathrm{layer}_{L_{\mathrm{declared}}}$ under $\texttt{arch\_spec}$ because the real backprop chain is longer; (iii) the network anchor observes inter-node communication volume proportional to $L_{\mathrm{actual}}$. Any of these three discrepancies is caught by a single in-training or genesis challenge.

  \item \textbf{Data larger than declared.} The adversary trains on $\alpha$-fold more data than committed in $R(\mathtt{dataset})$. The $\texttt{arch\_spec}$-declared shuffling seed determines which dataset indices appear in batch $t$. A fraction $1 - 1/\alpha$ of steps use indices absent from $R(\mathtt{dataset})$. Opening the layer-0 input commitment $R(\mathtt{mlp}_0\mathtt{\_in})$ at committed positions fails for those steps. Detection probability with $Q_{\mathrm{data}}$ uniform random input-openings:
  \[
    P(\text{detect}) = 1 - \alpha^{-Q_{\mathrm{data}}}.
  \]
  Hiding $\alpha = 10$ data needs $Q_{\mathrm{data}} = 2$ for $99\%$ detection; $\alpha = 2$ needs $Q_{\mathrm{data}} = 7$; $\alpha = 1.1$ needs $Q_{\mathrm{data}} = 49$. Large-scale data smuggling is trivially caught; the detection budget scales with the inverse of the attempted inflation factor.

  \item \textbf{More training steps than declared.} The network anchor (\Cref{network-tap}) observes inter-node collectives continuously for the duration of the run. For training that requires inter-node communication at every step (FSDP, TP across nodes, PP), the anchor's observed collective count matches the actual step count. The trainer's published chain has one entry per declared step. Discrepancy is a single integer comparison and requires no cryptographic proof, assuming the anchor's placement covers every inter-node fabric the run uses (\Cref{network-tap}).
\end{itemize}

\paragraph{The mode without a structural detection: compute smuggled through trainer-authored inputs.}
The cases above share a structure: each forces the executed computation to differ from the declared one, and the difference surfaces at a commitment. A further strategy avoids that structure entirely. The adversary declares, and then honestly executes, a run of exactly the committed architecture, dataset, and length --- but \emph{prepares its inputs using undeclared compute}, so that the verified run arrives at a capability its declared FLOP count would not buy. The cheapest instance is distillation: an undeclared model generates the synthetic corpus or the soft targets that are then committed as the official dataset. A stronger version of the same strategy is available to an adversary that already holds the weights it wishes to reach, which faces a problem of \emph{control} rather than of learning --- steering the trajectory toward a known target using whichever trainer-authored inputs the protocol leaves free (dataset content, curriculum and batch composition, learning-rate and regularisation schedules, auxiliary losses). That the dataset channel carries decisive authority over the resulting model's \emph{capabilities} is not in doubt, since training on model-generated data is standard industrial practice; whether any channel, including the dataset, confers authority toward a \emph{specific} target is, to our knowledge, unmeasured (\opref{reachability}).

None of the detections above fires: commitment shapes match \texttt{arch\_spec}, every opened input lies in the committed dataset, and the anchor's collective count matches the declared step count. The run is bit-exactly correct, and would remain so under a hypothetical full-coverage protocol with no sampling floor at all. What fails is not the verification but the inference from verified run-FLOPs to model capability. Whether it is also under-declaration in the strict sense turns on where the accounting boundary is drawn: Article~51(2)~\citep{euaiact_art51} states the threshold over \emph{cumulative} training compute, and recital~111 of the same Regulation counts the activities intended to enhance capabilities prior to deployment, ``such as pre-training, synthetic data generation and fine-tuning'' --- so the compute the provider spends generating a synthetic corpus falls inside the stated boundary. What the Regulation leaves unaddressed is \emph{attribution}: whether compute spent by a third party --- an upstream generator of the corpus, or the training of the model being distilled from --- counts toward the model under assessment. Settling that attribution question is part of \opref{compute-accounting} rather than something this paper resolves.

We know of no cheap structural detection. The capacity of the channel is the entropy of the trainer-authored inputs, which the dataset dominates by many orders of magnitude ($\sim$$10^{14}$ bits for a frontier corpus against $\sim$$10^{2}$ for an initialisation seed), so measures that constrain the low-entropy inputs --- including auditor-supplied initialisation seeds, which do close the channel they target --- do not materially reduce it.

As \Cref{sec:sampling-guarantees} notes, three cases must be distinguished. The first is
\emph{hidden compute}: the adversary really did spend more than it declared. Moving the
accounting boundary from the run to the facility addresses this problem \emph{to the extent
that the preparation ran on metered silicon}. This move converts a cryptographic problem
into a coverage problem rather than dissolving it (\opref{compute-accounting}). The second
case is \emph{capability drift}: a recipe genuinely costs what it declares but buys more
than the threshold's authors assumed. This is a property of the proxy, because the map
from training compute to capability is not stationary. No accounting boundary bounds
capability drift, and this roadmap does not attempt to close it. In the third case, the
preparation compute may have been spent and declared by \emph{someone else}, as when a
trainer distils from a publicly released model. In that
case, nobody under-declares compute, and a per-model threshold does not reach it. A
verification architecture can be held responsible for the hidden-compute case, and claims
about compute-threshold attestation should be scoped accordingly.

Setting that mode aside, the cases above imply a compute-threshold attestation needs far fewer challenges than general-purpose $\texttt{arch\_spec}$ verification. A conservative budget needs only a genesis proof, a handful of input-data challenges ($Q_{\mathrm{data}} \sim 10$, to rule out ${\geq}2\times$ data smuggling) and the anchor-count check. In the generic worst case, by contrast, the auditor needs $500$--$1{,}000$ deep challenges, each covering at least one whole microbatch (\Cref{appx:proof-cost:full-step}).

The overall cost-analysis tables retain the larger sampling budget as the default for two reasons. Formal sensitivity analysis (\opref{algebraic-reduction}) could yet reveal corner cases that rescale the argument for the four structural detections above. Attestations beyond compute (for instance data-content filters targeting \emph{semantic} smuggling within the committed dataset, or procedural attestations on fine-grained regime compliance) lack the structural detection surface of compute and operate closer to the 500-challenge regime against adaptive adversaries.

\subsection{MoE verification cost}\label{appx:proof-cost:moe}

Per \opref{moe}, the proof cost per microbatch for MoE scales with per-token active compute, not with total parameters. For an MoE with $K$ active experts per token of per-expert dimensions ($d_{\mathrm{model}} \times d_{\mathrm{expert}}$), the per-sample FFN cost is approximately $K$ times the cost of a dense model with the same per-expert dimensions, plus a small routing overhead. Kimi~K2 ($K = 8$, $d_{\mathrm{model}} = 7{,}168$, $d_{\mathrm{expert}} = 2{,}048$) costs substantially less per step than a hypothetical 1T dense model, but more than a 32B dense model because each sampled output requires verifying all $K$ active experts.

Router verification adds a small per-token GEMM of inner dimension $d_{\mathrm{model}}$ plus a top-$K$ selection step (either a dedicated precompile or a decomposition into pairwise comparisons, scaling as $N \log K$ for $N$ total experts). Load-balancing auxiliary losses (importance, load) introduce per-token softmax and division, covered by existing precompiles 4 and 6 at negligible cost relative to the expert GEMMs. All-to-all canonicalisation for expert parallelism is a separate open problem (\opref{wire-to-tensor} and \opref{moe}).

Concrete cost figures for MoE verification depend on sampling-scheme design decisions that are themselves open (\opref{moe}); detailed per-architecture numbers are left to future work once \opref{moe} is resolved.

\subsection{Proof-flow costs}\label{appx:proof-cost:flows}

The protocol produces three proof types (\Cref{sec:verifying}). Their costs differ.

\subsubsection{Genesis proof}

Executed once at initialisation on an auditor-chosen batch of $b$ samples drawn from the committed dataset. The proof covers one full training step on that batch (forward, backward and optimiser update). Its cost is the budget of \Cref{appx:proof-cost:per-layer-budget} for the chosen $b$, plus $b$ dataset-membership Merkle-path openings (one per sampled index), plus the optimiser update, which the table does not price. At $b \leq$ one 2{,}048-token microbatch, the proof completes in about the $7.4$-hour envelope on 1{,}024 GPUs. Reducing $b$ shrinks only the weight-gradient inner dimensions and the membership openings; the forward and input-gradient costs are set by $k$, not by $b$. A reduced-$b$ genesis is therefore cheap, but it is a genesis on $b$ samples, not on a full production batch.

\subsubsection{In-training step proof}\label{appx:proof-cost:in-training}

Executed many times across a training run, each invocation on a sampled step $t$. Four challenge granularities are meaningful in practice; the auditor can mix them according to the threat model.

\begin{itemize}
  \item \textbf{Single-commit challenge} (verify one committed root against an adjacent committed root via a single declared operation, for instance a GEMM verified by the MAC precompile, an activation verified by the BF16 lookup, or a LayerNorm verified by the FP32 nonlinear precompiles): up to $\sim$$2 \times 10^{10}$ constraints, $\sim$$20$\,s on $1{,}024$ GPUs, $\sim$\$$12$ at \$$2$/GPU-hour. GEMM-backed challenges set this headline cost; challenges that exercise cheaper precompiles (table lookups, FP32 nonlinear ops) are essentially free by comparison.
  \item \textbf{Layer-partial challenge} (verify one layer's forward pass against its input and output commitments, or one layer's backward pass against its gradient-boundary commitments; the two are structurally independent sub-proofs): $\sim$$1 \times 10^{11}$ constraints each, $\sim$$1.8$\,min on $1{,}024$ GPUs, $\sim$\$$60$.
  \item \textbf{Depth-$N$ partial chain} ($N$ consecutive forward layers, or $N$ consecutive backward layers, chained via their committed boundary roots): $N$ times the layer-partial cost. Useful when the auditor wants one proof that cross-checks multiple intermediate roots and their boundary consistency, for example an entire pipeline-parallel stage.
  \item \textbf{Microbatch challenge} (all 126 layers, forward plus backward, for one 2{,}048-token microbatch; used for genesis and periodic comprehensive audits): $\sim$$2.6 \times 10^{13}$ constraints, $\sim$$7.4$\,h on $1{,}024$ GPUs, $\sim$\$$15$K. Proving is embarrassingly parallel across segments, so on $16{,}384$ GPUs the same work completes in $\sim$$28$\,min for the same total GPU-hours. A challenge covering a whole training step costs far more and is not yet priced (\Cref{appx:proof-cost:full-step}).
\end{itemize}

All four costs are per 2{,}048-token microbatch. Each assumes that the challenged objects are committed per microbatch, or that the challenged weight-gradient entries sum over only 2{,}048 tokens. We therefore do not total a challenge mix over a full run. Such a total depends on the commitment design of \Cref{appx:proof-cost:full-step}. A mix would still combine many cheap single-commit challenges for broad coverage with a few deep challenges for boundary integrity. Auditors targeting only compute-threshold attestation can run a much smaller budget (see \Cref{appx:proof-cost:defense-depth}).

Per-challenge response time exclusive of proof generation is dominated by the host-side re-run: loading the stored weights from the rolling window (seconds on NVMe) and re-executing the forward + backward pass for the challenged step on a partial replica (minutes to hours depending on challenge depth).

\subsubsection{Ex-ante attestation running invariant}

Ex-ante attestations enforce invariants (e.g.\ $F_{\mathrm{cum}}(t) \leq \mathtt{max\_total\_flops}$) at every step proof. The invariant check is a single addition and a single comparison per step, adding a constant number of constraints (under $100$) to each step proof. For any realistic invariant list the contribution is negligible relative to the per-step MAC budget.

\subsection{Deeper proofs and proof composition}\label{appx:proof-cost:composition}

A depth-1 step proof trusts its input Merkle roots: if the trainer committed fake input roots, the proof still passes (the auditor catches the inconsistency downstream when the fake root does not match the preceding step's output). A depth-$N$ proof starts from $N$ layers earlier and verifies each intermediate commitment along the chain in-proof, at roughly $N$ times the single-layer cost.

\begin{table}[H]
  \centering
  \small
  \resizebox{\textwidth}{!}{%
  \begin{tabular}{lll}
    \toprule
    Challenge & Proving time (1{,}024 GPUs) & What it cross-checks \\
    \midrule
    Single-commit (one declared op) & $\sim$$20$\,s & One committed root against its adjacent committed root \\
    Layer-partial (forward \emph{or} backward of one layer) & $\sim$$1.8$\,min & That layer's input/output (or gradient) boundary \\
    Depth-6 partial chain (forward \emph{or} backward) & $\sim$$11$\,min & One PP stage: intra-stage roots + one PP boundary \\
    Microbatch (all 126 layers, forward + backward) & $\sim$$7.4$\,h & Every intermediate root of one microbatch, every anchor observation \\
    \bottomrule
  \end{tabular}%
  }
  \caption{Proving time at Llama~3.1 405B scale on 1{,}024 GPUs, per 2{,}048-token microbatch. Within a proof, layer-level sub-proofs run in parallel on independent GPU subsets and are folded via recursive composition into a single $\sim$200\,KB aggregate. Linear scaling in cluster size: a microbatch proof on 16{,}384 GPUs completes in $\sim$$28$ minutes for the same total GPU-hours.}
  \label{tab:proof-cost-depths}
\end{table}

A realistic challenge mix combines shallow challenges for broad coverage (many, cheap) with occasional deep challenges that cross-check boundary commitments (few, expensive). The exact mix is an operational decision set by the auditor's risk budget.

\subsection{Training-side hashing overhead}\label{appx:proof-cost:hashing}

\subsubsection{What must be hashed per microbatch}

Six tensors per layer are committed during training (\Cref{constant-hashing}): layer input, $Q$, $K$, $V$, attention output, layer output. At Llama~3.1 405B with $s = 2{,}048$, $m = 2{,}048$, activations are $m \cdot d_{\mathrm{model}} \cdot 2\text{\,B} = 67$\,MB each for the layer input/output and $m \cdot n_{\mathrm{heads}} \cdot d_{\mathrm{head}} \cdot 2$ = $67$\,MB for Q/O, $m \cdot n_{\mathrm{kv\_heads}} \cdot d_{\mathrm{head}} \cdot 2 = 4$\,MB for K/V. Per layer, aggregate tensor bytes to hash are ${\sim}275$\,MB. Over 126 layers the total is ${\sim}35$\,GB per microbatch.

The concurrent GPU stream delivers ${\sim}80$--$120$\,GB/s effective throughput (CUDA hashing kernel, H100), putting concurrent hashing time at ${\sim}300$--$450$\,ms per microbatch. At a typical compute time of $\sim$1\,s per microbatch this is within the overlap window with ${\sim}1$--$3\%$ memory-bandwidth contention on the primary compute stream.

Wire-bound collective payloads are hashed by the network anchor (Tier~1 physical TAP or Tier~2 attested SmartNIC). Anchor hashing happens outside the GPU budget and does not contribute to trainer-side overhead.

\subsubsection{DPU/SmartNIC hashing capabilities (Tier~2 context)}

For Tier~2 deployments the anchor lives in the SmartNIC. Relevant hashing throughput figures:

\begin{table}[H]
  \centering
  \small
  \resizebox{\textwidth}{!}{%
  \begin{tabular}{lrll}
    \toprule
    Device & SHA-256 throughput & Mechanism & Notes \\
    \midrule
    NVIDIA BF3 & $\sim$10 GB/s & Software on 16$\times$ ARM A78 cores & Hardware SHA-256 removed vs BF2 \\
    NVIDIA BF4 (expected) & $\sim$40 GB/s & 64$\times$ Neoverse V2 cores & Announced; throughput TBD \\
    Marvell OCTEON 10 & $\sim$12--25 GB/s & Hardware NITROX V & Commercially available \\
    Dedicated FPGA (Tier~1) & $\sim$100+ GB/s & Pipelined SHA-256 hardware core & Per \opref{tap-hardware}, design target \\
    \bottomrule
  \end{tabular}%
  }
  \caption{Observed or projected SHA-256 throughput for network-anchor-side hashing hardware. BF3 software-only throughput is marginal for cluster-scale inter-node traffic; hardware-accelerated silicon (Marvell OCTEON, BF4, or a dedicated Tier-1 FPGA) is required to sustain line rate at frontier cluster fabrics.}
  \label{tab:proof-cost-dpu-hashing}
\end{table}

\subsubsection{Total training-side overhead}

Combining determinism and hashing costs, the total training-side overhead at Llama~3.1 405B is shown in \Cref{tab:proof-cost-training-overhead}.

\begin{table}[H]
  \centering
  \small
  \resizebox{\textwidth}{!}{%
  \begin{tabular}{lrl}
    \toprule
    Component & Overhead & Confidence \\
    \midrule
    Determinism (attention backward-dominated) & $1.6$--$8.2\%$ & Measured on Llama~7B H100; seq-len-dependent \\
    Concurrent Merkle hashing (6 tensors/layer, GPU stream) & $0.5$--$1.5\%$ & Estimate from 2-pass kernel + optimised target \\
    Weight rolling-window storage (81\,TB) & $<0.1\%$ & Small vs cluster's existing checkpointing storage \\
    Anchor-side hashing & $\sim$0\% (trainer) & Moves to anchor hardware \\
    \midrule
    Total trainer-side overhead & $\sim$$2$--$10\%$ & Sequence-length-dependent \\
    \bottomrule
  \end{tabular}%
  }
  \caption{Training-side overhead at Llama~3.1 405B. Determinism is the dominant cost and is attention-backward-driven, scaling with sequence length. For a \$$100$M training budget the end-to-end trainer-side verification overhead is $\sim$\$$2$--$10$M.}
  \label{tab:proof-cost-training-overhead}
\end{table}

\subsection{Cost summary at Llama~3.1 405B}\label{appx:proof-cost:summary}

\begin{table}[H]
  \centering
  \small
  \resizebox{\textwidth}{!}{%
  \begin{tabular}{lrl}
    \toprule
    Component & Cost at \$100M training budget & \% of training \\
    \midrule
    Determinism tax & \$1.6--8.2M & 1.6--8.2\% \\
    Concurrent Merkle hashing & \$0.5--1.5M & 0.5--1.5\% \\
    Genesis proof (one-time, reduced-batch) & $\sim$\$15K & $\sim$0.015\% \\
    In-training challenges & not totalled; per challenge, from $\sim$\$12 (one operation) to $\sim$\$15K (all layers), per microbatch (\Cref{appx:proof-cost:full-step}) & --- \\
    Weight storage (81\,TB rolling) & $\sim$\$50K/year (annual rate; held for the run's duration) & $<$0.1\% \\
    Ex-ante attestation running cost & negligible & $<$0.001\% \\
    \midrule
    \textbf{Total, excluding in-training challenges} & \textbf{$\sim$\$2--10M} & \textbf{$\sim$2--10\%} \\
    \bottomrule
  \end{tabular}%
  }
  \caption{Verification overhead for a Llama~3.1 405B-scale pre-training run on a \$100M training budget. Of the priced components, the determinism tax dominates; hashing, storage and the genesis proof add $\sim$0.5--1.6\%. Trainer-side component endpoints sum to ${\approx}2.1$--$9.7\%$ (\Cref{tab:proof-cost-training-overhead}). The auditor-side challenge budget is not included: its per-challenge figures price one 2{,}048-token microbatch, and a total over a run depends on how gradients are committed (\Cref{appx:proof-cost:full-step}).}
  \label{tab:proof-cost-summary}
\end{table}

\subsection{Caveats and open questions}\label{appx:proof-cost:caveats}

\begin{enumerate}
  \item \textbf{Proving throughput ($\sim$$10^6$ constraints/s/GPU)} is approximate and may be optimistic for constraint-heavy FP precompiles. A direct measurement is part of the planned empirical validation work. Relative comparisons between approaches are robust to this assumption.
  \item \textbf{Per-MAC constraint count ($\sim$$90$)} has a range of $50$--$150$ depending on implementation. Absolute proof times scale linearly with this; the relative structure of the breakdown does not.
  \item \textbf{Non-linear lookup tables} (BF16 GELU, SiLU) must match the GPU kernel output at every input. Generation and validation of these tables is routine engineering work.
  \item \textbf{Sampling security at $f < 1\%$} is a defense-in-depth argument, not a hard theorem. Formal sensitivity analysis (\opref{algebraic-reduction}) would close this.
  \item \textbf{GPU-concurrent Merkle hashing} (0.5--1.5\% overhead, with 1--3\% transient memory-bandwidth contention) needs empirical validation on H100 with realistic training workloads.
  \item \textbf{Fusion assumptions} (Flash Attention truly fused, FFN GEMMs separable with lazy commitment) are kernel-dependent. A training stack that fuses FFN end-to-end (e.g.\ a memory-bound variant) materially increases per-sample FFN verification cost, and the sample count for fused blocks must be reduced correspondingly.
  \item \textbf{MoE cost numbers} depend on sampling-scheme design decisions open under \opref{moe}. The order-of-magnitude framing ($K \times$ per-expert-dense) holds across reasonable choices; concrete per-architecture figures require \opref{moe} resolution.
  \item \textbf{INT8 training with INT32 accumulation} (supported by H100/B200 Tensor Cores) would eliminate float non-associativity and make Freivalds directly applicable. No frontier model currently uses full INT8 pretraining, but the industry trend toward FP8 and FP4 is adjacent; verification cost at low-precision regimes is a worthwhile future study.
\end{enumerate}

\section{Determinism enforcement: detailed benchmarks and analysis}\label{app:determinism-details}

This appendix provides detailed measurements and analysis supporting the determinism
discussion in \Cref{enforce-determinism}.

\subsection{Kernel-level determinism configuration}

We enforce determinism through optimised kernel selection rather than PyTorch's blanket
\texttt{torch.use\_deterministic\_algorithms(True)} flag.
The configuration has three components. First, FlashAttention-2 runs in deterministic backward mode (\texttt{FLASH\_ATTENTION\_DETERMINISTIC=1}), which serializes cooperative-thread-array (CTA; GPU thread block) reductions at a measured cost. Second, cuBLAS uses pinned algorithm selection (\texttt{CUBLAS\_WORKSPACE\_CONFIG}). Third, NCCL uses NVLS-based reduction, which provides a fixed reduction order at near-zero overhead for FSDP allgather and reduce-scatter.

We measured the end-to-end overhead of this configuration on 8$\times$ H100 HBM3 with
NVSwitch, training Llama~7B with FSDP:

\begin{table}[H]
  \centering
  \small
  \begin{tabular}{llll}
    \toprule
    Seq.\ length & Non-det (ms/step) & Det (ms/step) & Overhead \\
    \midrule
    2048 & 531.8 & 540.3 & +1.6\% \\
    4096 & 999.1 & 1038.0 & +3.9\% \\
    8192 & 1129.9 & 1222.0 & +8.2\% \\
    \bottomrule
  \end{tabular}
  \caption{Measured full-determinism overhead (Llama~7B, FSDP, 8$\times$H100 HBM3). The
  overhead is attention-dominated: cuBLAS GEMMs contribute $\sim$0.8\% and NCCL
  communication (NVLS) contributes $\sim$0\%.}
  \label{tab:determinism-overhead}
\end{table}

\subsection{Parallelism dimensions beyond FSDP}

The measurements above apply to FSDP, which uses reduce-scatter and allgather (both
deterministic under NVLS at near-full bandwidth).
Frontier training runs typically combine multiple parallelism dimensions: tensor
parallelism (TP) within a node, pipeline parallelism (PP) and FSDP across nodes, and
expert parallelism (EP) for mixture-of-experts models.

\paragraph{Tensor parallelism.}
TP is the most challenging dimension.
It uses allreduce after each column-parallel and row-parallel GEMM.
Our NCCL benchmarks show that intra-node NVSwitch allreduce is \emph{not} deterministic
under NVLS above 128\,MB BF16.
The only deterministic configurations we found, sweeping algorithm, protocol and CTA
count at these message sizes, are Tree+Simple (64\% bandwidth loss) and Ring+1CTA
(89\% bandwidth loss, unusable).
This is a significant penalty: TP allreduce occurs at every layer, so even Tree+Simple
could add 6--10\% to total step time depending on the ratio of TP communication to
compute.

Interestingly, inter-node allreduce (over TCP/RoCE) is deterministic under default NCCL
configuration at all tested sizes (32--1024\,MB BF16, 2-node setup).
The determinism problem is specific to the intra-node NVSwitch code path.

\paragraph{Pipeline and expert parallelism.}
PP is straightforward: it sends activations between pipeline stages in a fixed schedule,
and determinism reduces to the same per-layer guarantees described above.
EP introduces all-to-all communication for expert routing. Our benchmarks show NVLS
all-to-all is deterministic at all tested sizes, and the routing decisions themselves are
deterministic given a fixed gating function.

\subsection{Towards a custom TP allreduce}

The overhead of Tree+Simple for TP can likely be eliminated by a custom deterministic
allreduce kernel optimised for the TP use case.
TP operates under narrow, exploitable constraints: fixed topology (8~GPUs on NVSwitch),
fixed message sizes (determined by model dimensions, constant across steps), and a fixed
reduction order is all that is needed for bit-exactness.
A custom kernel using NVLink direct writes with a hardcoded reduction tree for the
specific topology and message size could match NCCL's default bandwidth while guaranteeing
determinism by construction.
Projects such as MSCCL (Microsoft's custom collective compiler) demonstrate that
hand-written collectives can match or exceed NCCL for fixed topologies.

Importantly, PyTorch's process group architecture supports per-dimension configuration:
TP, FSDP, and PP each use separate communicators.
The custom kernel would replace NCCL only for TP allreduce calls, while FSDP and PP
continue to use NVLS under default configuration.
This avoids any performance regression on the already-solved FSDP and PP dimensions.

Our current measured overhead (1.6--8.2\%) applies to FSDP-only training.
For FSDP+TP setups, the E2E impact of the TP allreduce bandwidth loss is likely small in
practice: TP communication and layer computation overlap, and for typical TP=8 intra-node
configurations the compute time dominates.
For example, a 100\,MB allreduce at Tree+Simple bandwidth (169\,GB/s) takes $\sim$0.6\,ms,
which is well within the compute time of a single Transformer layer at frontier scale.
The overhead would become significant only if TP communication becomes the bottleneck, for
instance at very large TP degrees across nodes or with unusually small per-GPU compute.
This remains to be validated end-to-end.

\section{Network anchoring: hash choice, reconciliation, and open problems}\label{app:network-anchor}

This appendix supports \Cref{network-tap} by (i) justifying the dual-hash design, (ii)
describing the tree-structure and chunk-size tradeoffs, and (iii) discussing composite
precompile optimisations; the wire-to-tensor mapping problem and the remaining open
engineering problems are catalogued in \Cref{app:open-problems}
(\opref{wire-to-tensor}, \opref{tap-hardware} through \opref{intranode-witness}).

\subsection{Hash function choice: two regimes, two hashes}

The architecture uses two distinct hash functions for its Merkle commitments: a
zkVM-native hash (precompile~3, Poseidon today) for the trainer's on-the-fly tensor
commitments, and a silicon-native hash (precompile~8, SHA-256) for the network anchor.
This is not redundant: each hash is chosen for the bottleneck of the device that hashes at
line rate.

The two hash families have opposite cost profiles.
Poseidon operates in the trainer's native prime field: its round function is built from
$x^5 \bmod p$ S-boxes and MDS matrix multiplications over $\mathbb{F}_p$.
Poseidon is inexpensive inside the proof because of this arithmetic: it costs ${\sim}50$ constraints per compression and ${\sim}1{,}500$ constraints per path of depth 30. Outside the proof, however, the same arithmetic is expensive because prime-field multiplication is not a native silicon primitive. Commercial FPGAs and DPUs achieve at best sub-GB/s throughput, with significant area cost.
SHA-256 inverts this.
Its round function uses 32-bit bitwise operations (AND, XOR, rotation) and modular
addition, which map directly to Boolean gates and integer ALUs: Intel SHA-NI sustains
${\sim}1.9$ cycles/byte, and FPGA IP cores pipeline at 5--20\,Gbps per core and stack to
hundreds of GB/s per device.
The same simplicity is costly in-circuit, because bitwise operations over 32-bit words
must be encoded via lookup tables or bit decomposition (${\sim}7{,}500$ constraints per
compression, ${\sim}150{,}000$ constraints per path of depth 20).

The asymmetry in cluster-scale throughput is therefore two to three orders of magnitude.
Aggregate inter-node bandwidth on a modern training cluster reaches several hundred GB/s;
Poseidon-family hashes cannot be computed at those rates on commodity silicon.
The network anchor must use a hash with mature silicon support, which today means SHA-256.
Conversely, the trainer's Merkle paths are opened thousands of times per sampled GEMM
inside the zkVM; using SHA-256 for those paths would increase the proof budget for path
verification by roughly $150\times$, whereas the GPU overhead saved on the trainer side is
under 1\%.
Using one hash per regime is the only design that is efficient on both sides.

This split between an in-circuit-cheap hash and a silicon-cheap hash is a standard pattern
whenever a system bridges two cost models with incompatible primitives; Ethereum rollups,
for example, use Keccak at the L1 interface and zk-friendly hashes inside the rollup for
the same reason.

\subsection{Tree versus flat, and chunk size}

Committing to wire payloads as a Merkle tree rather than a single flat hash trades a small
amount of TAP-side work for a large reduction in challenge-time zkVM cost.
Flat SHA-256 commits a payload with one digest; verifying consistency at challenge time
then requires rehashing the entire payload inside the zkVM.
For a 1\,GB collective with the SHA-256 precompile this is roughly $16 \times 10^6$ blocks
at ${\sim}7{,}500$ constraints each, or ${\sim}10^{11}$ constraints per challenge.
Tree-structured commitments reduce this to $O(\log n)$ path verification plus a local
chunk rehash.

The tree structure costs almost nothing on the TAP side.
For $n$ leaves of size $\ell$, the total hashed volume is $n\ell + 64(n-1)$ bytes ---
each internal node hashes two 32-byte digests, not leaf-sized data --- so the overhead
over flat hashing is the factor $1 + 64/\ell$: with MB-sized leaves over a GB-sized
payload it is under $1.0001\times$.

The leaf size $\ell$ is a tuning parameter.
Smaller leaves shrink the chunk rehash but deepen the tree, inflating path verification.
Larger leaves invert the trade.

\begin{table}[H]
  \centering
  \small
  \begin{tabular}{lrrrr}
    \toprule
    Leaf size & Tree depth (1\,GB) & Path cost & Chunk rehash cost & Total per opening \\
    \midrule
    32\,B (one element) & 25 & ${\sim}190{,}000$ & ${\sim}7{,}500$ & ${\sim}200{,}000$ \\
    16\,KB & 16 & ${\sim}120{,}000$ & ${\sim}1{,}900{,}000$ & ${\sim}2{,}000{,}000$ \\
    64\,KB & 14 & ${\sim}105{,}000$ & ${\sim}7{,}500{,}000$ & ${\sim}7{,}600{,}000$ \\
    1\,MB & 10 & ${\sim}75{,}000$ & ${\sim}120{,}000{,}000$ & ${\sim}120{,}000{,}000$ \\
    \bottomrule
  \end{tabular}
  \caption{Constraint cost per wire-bound sample opening for a 1\,GB collective as a
  function of Merkle-leaf size. Path and rehash costs assume the SHA-256 precompile at
  ${\sim}7{,}500$ constraints per compression block.}
  \label{tab:leaf-size-tradeoff}
\end{table}

The zkVM cost is minimised at small leaves, which places the sweet spot at tens of
kilobytes when balanced against TAP-side chunk-boundary bookkeeping and packet-boundary
alignment (catalogued under \opref{tap-hardware} in \Cref{app:open-problems}).
Values in the 16--64\,KB range give opening costs in the 2--8 million constraint range per
sample, bounded and much smaller than the MAC-chain budget of a GEMM.

\subsection{Composite Merkle-path precompile}

The precompile catalogue in \Cref{tab:precompiles} reports costs assuming per-compression
precompile invocations composed in software (a loop over tree levels calling the SHA-256
or Poseidon primitive at each step).
A composite precompile that absorbs path-traversal logic (sibling loading, direction
bits, chained compressions) into a single dedicated circuit eliminates the per-level
RISC-V overhead and enables tighter constraint layouts across levels.
Cairo's Merkle builtins and ongoing work in SP1 and Jolt follow this pattern.
Estimated savings are 20--30\% for SHA-256 paths (where the compressor is heavy and glue
overhead relatively matters) and 15--25\% for Poseidon paths.
A further optimisation is a multi-path precompile that verifies several paths in one
invocation, amortising transition cost across openings; this gives another 10--20\% when
the same challenge opens multiple paths (as is typical for GEMM sampling, which opens 3--5
paths per sampled entry).

These are implementation-level optimisations; they refine the proof-cost figures
without changing the architectural claims.

Several open engineering problems attach to the network anchor: the hardware TAP itself,
wire-to-tensor mapping (with queue-pair auditability, Tier~2 attestation protocol,
and chunk-boundary alignment as subsidiary questions), and intra-node coverage. These
are catalogued in \Cref{app:open-problems} (\opref{tap-hardware} through \opref{intranode-witness}) rather than duplicated here.

\section{Toy example: end-to-end verification walkthrough}\label{appendix:toy-example}

This appendix walks through the complete verification protocol on a minimal example: a
two-layer MLP with ReLU activations, split across two nodes by pipeline parallelism. The
example exercises precompiles 1 (MAC), 2 (BF16 lookup), 3 (Merkle path with the
zkVM-native hash), and 8 (SHA-256 for network-anchor reconciliation). It does not exercise
precompiles 4--7 (FP32 nonlinear operations for softmax, normalisation, Adam, or loss);
those are triggered by additional operations in the full Transformer block and follow the
same pattern.

\subsection{Setup}

The model is a two-layer MLP,
$f(x) = \mathrm{ReLU}\big(\mathrm{ReLU}(x\,W_1)\,W_2\big)$. Inputs are rows of the batch
matrix, following the $(m \times d)\times(d \times n)$ GEMM convention used throughout
\Cref{appendix:proof-cost}. We train the model with pipeline parallelism over two nodes:
\begin{itemize}
  \item \textbf{Node 1} holds the first linear layer $W_1 \in \mathbb{R}^{d_{\text{in}}
  \times d_h}$.
  \item \textbf{Node 2} holds the second linear layer $W_2 \in \mathbb{R}^{d_h
  \times d_{\text{out}}}$.
  \item Loss is $L = \tfrac{1}{2} \|y - \mathrm{target}\|^2$ for simplicity of the
  backward pass.
  \item Optimiser is plain SGD ($W \leftarrow W - \eta \cdot \text{grad}$); no Adam state
  to track.
  \item Weights are BF16 with FP32 accumulators; rounding is RNE.
\end{itemize}
Each training step involves two inter-node messages: a forward activation passed from Node
1 to Node 2, and a backward gradient passed from Node 2 back to Node 1. The network anchor
observes both.

\paragraph{Notation.}
We write $R(\cdot)$ for a Merkle root over a tensor or byte sequence. Trainer-side Merkle
roots use a zkVM-native hash (Poseidon in the current instantiation, verified by
precompile~3). Network-anchor Merkle roots use a silicon-native hash (SHA-256, verified by
precompile~8). Both support $\log n$ membership proofs against the committed root; they
differ only in per-path cost inside the proof. We write $\mathrm{Hash}(\cdot)$ for the
outer compound commitment.

\subsection{Phase 1: pre-commit}

Before training begins, the trainer publishes the initial commitment:
\[
  h_{\text{commit}} = \mathrm{Hash}\!\big(\mathrm{Hash}(\mathtt{arch\_spec}) \;\|\;
  R(\mathtt{dataset}) \;\|\; R(W_0)\big)
\]
where $\mathtt{arch\_spec}$ records the layer dimensions, activation type (ReLU),
optimiser (SGD), precision (BF16/FP32, RNE), parallelism strategy (PP, world size 2), and
shuffling seed. $R(W_0)$ is the Merkle root over the initial weight shards across both
nodes; $R(\mathtt{dataset})$ is the Merkle root over the training data.

The auditor records $h_{\text{commit}}$ and installs a network anchor on the inter-node
link between Node 1 and Node 2.

\subsection{Phase 2: one training step}

We annotate a single step with every commitment published and every message observed. Node
1 processes batch shard $x$; arrows mark inter-node traffic.

The two nodes execute sequentially along the pipeline direction. Arrows mark inter-node messages observed by the network anchor.

\paragraph{Node 1 (holds $W_1$).}
\begin{verbatim}
mlp_1_in  <- x                                         # batch shard
mlp_1_out <- mlp_1_in @ W_1
act_1_out <- ReLU(mlp_1_out)
publish R(mlp_1_in), R(mlp_1_out), R(act_1_out)
send act_1_out  ---- forward msg_1 ---->  Node 2      (anchor: R(msg_1))
\end{verbatim}

\paragraph{Node 2 (holds $W_2$).}
\begin{verbatim}
recv act_1_out  from Node 1
mlp_2_in  <- act_1_out
mlp_2_out <- mlp_2_in @ W_2
act_2_out <- ReLU(mlp_2_out)
publish R(mlp_2_in), R(mlp_2_out), R(act_2_out)
L <- 0.5 * || act_2_out - target ||^2
grad_act_2 <- act_2_out - target
grad_mlp_2 <- grad_act_2 * (mlp_2_out > 0)
grad_W_2   <- mlp_2_in^T @ grad_mlp_2
grad_act_1 <- grad_mlp_2 @ W_2^T
publish R(grad_W_2)
send grad_act_1  ---- backward msg_2 ---->  Node 1    (anchor: R(msg_2))
\end{verbatim}

\paragraph{Node 1 (backward + optimiser).}
\begin{verbatim}
recv grad_act_1  from Node 2
grad_mlp_1 <- grad_act_1 * (mlp_1_out > 0)
grad_W_1   <- mlp_1_in^T @ grad_mlp_1
publish R(grad_W_1)
W_1 <- W_1 - eta * grad_W_1            # at Node 1
W_2 <- W_2 - eta * grad_W_2            # at Node 2
publish R(W_{t+1})                     # combined root
\end{verbatim}

After the step, the trainer's public hash chain contains $\mathrm{Hash}(\mathtt{arch\_spec})$, $R(W_t)$,
$R(\mathtt{dataset})$ (from pre-commit), plus per-step $R(\mathtt{mlp}_i\mathtt{\_in})$,
$R(\mathtt{mlp}_i\mathtt{\_out})$, $R(\mathtt{act}_i\mathtt{\_out})$, $R(\mathtt{grad\_W}_i)$
for $i \in \{1, 2\}$, and $R(W_{t+1})$. The network anchor's record contains
$R(\mathtt{msg}_1)$ (forward activation payload) and $R(\mathtt{msg}_2)$ (backward
gradient payload).

\subsection{Phase 3: challenge}

The auditor selects a step $t$, a node $i$, and $k$ random sample indices $(j_1, \ldots,
j_k)$ for the output entries to challenge. Because the indices are chosen \emph{after} the
Merkle roots have been published to the hash chain \emph{and} the chain has been frozen by
the auditor-signed terminal anchor --- the audit seed is opened only after freezing
(\Cref{constant-hashing}) --- the trainer cannot predict which entries will be sampled.

The challenge demands a proof of the following statement, phrased at Node~1 for
concreteness (\Cref{fig:proof-flow}):
\[
  \mathrm{Hash}(\mathtt{arch\_spec}),\; R(W_t),\; R(\mathtt{dataset}) \;\vdash\;
  R(\mathtt{mlp}_1\mathtt{\_in}) \;\rightarrow\; R(\mathtt{mlp}_1\mathtt{\_out})
  \;\rightarrow\; R(\mathtt{msg}_1)
\]
That is: executing the committed architecture with the committed weights on the committed
data produces intermediate tensors whose Merkle roots match the committed
$R(\mathtt{mlp}_1\mathtt{\_in})$ and $R(\mathtt{mlp}_1\mathtt{\_out})$, and the forward
message serialisation is consistent with the network anchor's observation
$R(\mathtt{msg}_1)$.

\subsection{Phase 4: verification inside the zkVM}

The proof checker is a public, auditable program. It reads the $\mathtt{arch\_spec}$ as a
private input, verifies it matches $h_{\text{commit}}$, then, for each sampled output
entry, runs the following checks. We illustrate one entry at Node~1: the output of the
first GEMM at row $r$, column $j$.

\begin{verbatim}
# ---- bind the private arch_spec to the public commitment ----
arch_spec = read_private_input()
assert Hash(arch_spec) == h_commit.arch_spec_hash

# ---- for each sampled (r, j) at Node 1 ----
for (r, j) in challenge_indices:

    # (a) input row: open against R(mlp_1_in)
    x_row    = read_hint()                 # private: mlp_1_in[r, :]
    path_in  = read_hint()
    MerklePath(x_row, path_in, R(mlp_1_in))          # precompile 3

    # (b) weight column: open against R(W_t)
    W1_col   = read_hint()                 # private: W_1[:, j]
    path_W   = read_hint()
    MerklePath(W1_col, path_W, R(W_t))               # precompile 3

    # (c) GEMM entry: recompute via MAC chain and bind to output root
    mlp_out_rj = read_hint()               # private: mlp_1_out[r, j]
    MAC_chain(x_row, W1_col, mlp_out_rj)             # precompile 1
    path_out = read_hint()
    MerklePath(mlp_out_rj, path_out, R(mlp_1_out))   # precompile 3

    # (d) ReLU output: verify via BF16 lookup, bind to activation root
    act_rj   = read_hint()                 # private: act_1_out[r, j]
    Lookup(mlp_out_rj, ReLU_TABLE, act_rj)           # precompile 2
    path_act = read_hint()
    MerklePath(act_rj, path_act, R(act_1_out))       # precompile 3

    # (e) anchor consistency: the same act_rj must appear at the
    #     computed byte offset in the forward message
    off_msg  = compute_msg_offset(arch_spec, r, j)   # offset within msg_1
    chunk    = read_hint()                 # chunk bytes containing off_msg
    path_msg = read_hint()
    MerklePath_SHA256(SHA256(chunk), path_msg, R(msg_1))  # precompile 8
    off_in_chunk = off_msg - chunk_start(off_msg)
    assert chunk[off_in_chunk : off_in_chunk + bf16_size] == act_rj

emit "PASS"
\end{verbatim}

Step (e) is the link between the trainer's tensor commitment and the network anchor's wire
commitment: the same BF16 value of $\mathtt{act}_1\mathtt{\_out}[r,j]$ must be present
both under the trainer-committed $R(\mathtt{act}_1\mathtt{\_out})$ (via Poseidon path) and
under the anchor-committed $R(\mathtt{msg}_1)$ (via SHA-256 path at the offset dictated by
the serialisation defined in $\mathtt{arch\_spec}$).

The guest emits as public outputs: $h_{\text{commit}}$, the committed roots
$R(\mathtt{mlp}_1\mathtt{\_in})$, $R(\mathtt{mlp}_1\mathtt{\_out})$,
$R(\mathtt{act}_1\mathtt{\_out})$, the anchor root $R(\mathtt{msg}_1)$, and a ``PASS''
bit. The auditor STARK-verifies the proof (milliseconds on a laptop), checks that the
public outputs match the published hash chain and the anchor record, and accepts.

\subsection{What each party sees}

\begin{table}[H]
  \centering
  \small
  \resizebox{\textwidth}{!}{%
  \begin{tabular}{llll}
    \toprule
                                 & Trainer (host) & Proof checker (guest) & Auditor \\
    \midrule
    $\mathtt{arch\_spec}$         & full          & private input, bound to $h_{\text{commit}}$ & only $\mathrm{Hash}(\mathtt{arch\_spec})$ \\
    Weights $W_1, W_2$            & full          & hints, bound to $R(W_t)$                    & never \\
    Activations, gradients        & full          & hints, bound via MAC + lookup               & never \\
    Merkle roots $R(\cdot)$       & computes      & public verification anchors                  & reads from hash chain \\
    Anchor root $R(\mathtt{msg}_i)$ & not involved  & public verification anchor                   & reads from anchor \\
    STARK proof                   & produces it (runs the zkVM prover) & its execution is what the proof attests & verifies it \\
    \bottomrule
  \end{tabular}%
  }
  \caption{Information visible to each role in the two-node toy protocol. Private
  tensors enter the guest only as hints bound to public commitments and are never
  revealed to the auditor.}
\end{table}

\subsection{Proof cost}

For illustration, take $d_{\text{in}} = d_h = d_{\text{out}} = 1024$ and $k = 100$ sampled
output entries per committed tensor at each of the two nodes (a small value chosen for
readability; the main-text analysis uses $k \approx 4{,}605$ for $10^{-20}$ soundness
error at $1\%$ deviation). The approximate constraint budget for one step:

\begin{table}[H]
  \centering
  \small
  \resizebox{\textwidth}{!}{%
  \begin{tabular}{lll}
    \toprule
    Verification work & Precompile & Constraints (order of magnitude) \\
    \midrule
    First GEMM ($d_{\text{in}} \cdot 90$ per sample $\times k$) & MAC (1) & ${\sim}9 \times 10^6$ \\
    ReLU lookups ($k$ per node, two nodes) & BF16 lookup (2) & ${\sim}3 \times 10^3$ \\
    Second GEMM (same magnitude) & MAC (1) & ${\sim}9 \times 10^6$ \\
    Merkle paths, trainer side ($4$ per sample $\times k$, two nodes) & Poseidon path (3) & ${\sim}1.2 \times 10^6$ \\
    Anchor consistency ($1$ path $+$ $1$ chunk rehash per sample) & SHA-256 (8) & ${\sim}6 \times 10^6$ \\
    \bottomrule
  \end{tabular}%
  }
  \caption{Illustrative one-step constraint budget for the two-layer MLP at
  $d_{\text{in}} = d_h = d_{\text{out}} = 1{,}024$ with $k=100$ samples per committed
  tensor; order-of-magnitude estimates.}
\end{table}

Total: roughly $2.6 \times 10^7$ constraints per step, which is well inside the capacity
of current zkVMs (STARK proof size ${\sim}200$\,KB after recursive composition). At
production scale the MAC chain dominates because the product $d \cdot k$ grows by roughly
three orders of magnitude; see \Cref{appendix:proof-cost} for the full breakdown.

\section{Training specification (\texttt{arch\_spec}) schema}\label{appendix:arch-spec}

This appendix gives the concrete structure of the \texttt{arch\_spec} committed during pre-training (\Cref{pre-commit}) and verified by the proof checker (\Cref{architecture}). The \texttt{arch\_spec} is a private input to the proof; the auditor sees only $\mathrm{Hash}(\mathtt{arch\_spec})$ as part of $h_{\mathrm{commit}}$. The schema below is the reference definition against which implementations are written. It is presented in Rust-style record syntax for concreteness; any structured serialisation with an equivalent set of fields is acceptable.

\subsection{Top-level \texttt{arch\_spec}}

{\footnotesize
\begin{verbatim}
struct ArchSpec {
    // Model architecture
    layers:       Vec<LayerSpec>,
    embedding:    EmbeddingSpec,
    output_head:  OutputHeadSpec,

    // Training configuration
    optimizer:    OptimizerSpec,         // Adam, AdamW, SGD, ...
    data_loading: DataLoadingSpec,       // batch derivation from committed dataset
    precision:    PrecisionSpec,         // BF16/FP32, rounding, hardware

    // Distributed training
    communication: CommSpec,             // AllReduce, AllGather, Send, Recv
    parallelism:   ParallelismSpec,      // TP, PP, DP, FSDP configuration

    // Verification
    merkle_points: Vec<MerklePoint>,     // tensors committed on-the-fly
    rng_spec:      RNGSpec,              // data-shuffling and dropout seeds
}
\end{verbatim}
}

Every operation that takes place during a training step must be expressible as a composition of the elements declared by \texttt{arch\_spec}. Any operation outside this vocabulary is rejected by the proof checker.

\subsection{Layer specification}

{\footnotesize
\begin{verbatim}
struct LayerSpec {
    forward_ops:   Vec<OpSpec>,
    backward_ops:  Vec<OpSpec>,          // gradient computations mirroring forward
    optimizer_ops: Vec<OpSpec>,          // parameter update for this layer
}

enum OpSpec {
    // Linear algebra
    GEMM { m: usize, k: usize, n: usize },

    // Attention (fused; internal score and softmax tensors never materialise;
    // verified end-to-end at Q/K/V/attention-output boundaries; see the
    // fused-attention sampling budget in the proof-cost appendix)
    FusedAttention {
        n_heads:     usize,
        n_kv_heads:  usize,              // GQA: < n_heads ; MHA: == n_heads
        d_head:      usize,
        seq_len:     usize,
        causal:      bool,
        positional:  PositionalEncoding, // RoPE, Learned, ALiBi, None
    },

    // Non-linear activations (BF16, verified via precompile 2)
    Activation { kind: ActivationType },  // GELU, SiLU, SwiGLU, ReLU

    // Normalisation (RMSNorm used by Llama family)
    LayerNorm { dim: usize },
    RMSNorm   { dim: usize },

    // Structural
    Residual,

    // Mixture-of-experts block (out of scope for base protocol;
    // see the mixture-of-experts open problem in the roadmap appendix)
    MoEBlock {
        router:   RouterSpec,
        experts:  Vec<Vec<OpSpec>>,
        combine:  CombineSpec,
    },
}
\end{verbatim}
}

A forward \texttt{GEMM} with inputs $A$ and $B$ produces two backward \texttt{GEMM}
operations. One computes the gradient with respect to the input
($\mathrm{grad}_A = \mathrm{grad}_C \cdot B^\top$); the other computes the gradient with
respect to the weight ($\mathrm{grad}_B = A^\top \cdot \mathrm{grad}_C$). The
\texttt{backward\_ops} list enumerates both operations explicitly.

Fused operations carry both a forward and a backward variant; both are sampled end-to-end at their input and output commitment boundaries (\Cref{appx:proof-cost:commitments}).

\subsection{Precision specification}

{\footnotesize
\begin{verbatim}
struct PrecisionSpec {
    compute:           Precision,        // BF16 for parameters/activations
    accumulate:        Precision,        // FP32 for GEMM accumulators
    rounding:          RoundingMode,     // RoundToNearestEven
    accum_order:       AccumOrder,       // LinearLeftToRight, ...
    hardware:          HardwareTarget,   // H100, MI300X, ...
    cublas_algo:       Option<String>,   // pinned cuBLAS algorithm ID
    flash_attn_version: String,          // e.g. "FlashAttention-3-deterministic"
}
\end{verbatim}
}

The \texttt{accum\_order} field is load-bearing for determinism: given non-associative floating-point addition, different orders can produce different bit-exact results. The committed order is what the proof checker verifies.

\subsection{Communication specification}

\texttt{CommSpec} governs operations that perform floating-point reductions
(\texttt{AllReduce}, \texttt{ReduceScatter}). The reduction order is explicit so the
proof can verify the arithmetic. Data-movement operations that do not perform reductions
(\texttt{Send} and \texttt{Recv} for pipeline parallelism, and \texttt{AllGather} for
FSDP) are verified through Merkle-root matching at endpoints plus network-anchor
consistency, without reduction-order specification.

{\footnotesize
\begin{verbatim}
struct CommSpec {
    // Global NCCL configuration pinned for determinism (see the
    // determinism-enforcement appendix)
    nccl_algo:     NCCLAlgo,             // Ring, Tree, NVLS
    nccl_proto:    NCCLProto,            // Simple
    nccl_channels: usize,

    ops: Vec<CommOp>,
}

enum CommOp {
    AllReduce {
        tensor_shape:    Vec<usize>,
        dtype:           Dtype,
        reduction_order: Vec<usize>,     // explicit FP addition order
        group:           ProcessGroup,
    },
    AllGather {
        shard_shape:  Vec<usize>,
        dtype:        Dtype,
        gather_order: Vec<usize>,        // rank-order (0, 1, 2, ...)
        group:        ProcessGroup,
    },
    ReduceScatter {
        tensor_shape:    Vec<usize>,
        dtype:           Dtype,
        reduction_order: Vec<usize>,
        group:           ProcessGroup,
    },
    Send { tensor_shape: Vec<usize>, src: usize, dst: usize },
    Recv { tensor_shape: Vec<usize>, src: usize, dst: usize },
}
\end{verbatim}
}

\subsection{Parallelism specification}

{\footnotesize
\begin{verbatim}
struct ParallelismSpec {
    tp_degree:        usize,             // tensor parallelism (typically 8)
    pp_stages:        usize,             // pipeline parallelism stages
    dp_degree:        usize,             // data parallelism replicas
    fsdp_enabled:     bool,
    fsdp_shard_degree: usize,            // ranks sharing each parameter shard

    // Expert parallelism (MoE models only; out of scope for base protocol)
    ep: Option<ExpertParallelismSpec>,

    // Process-group topology
    tp_groups: Vec<Vec<usize>>,
    pp_groups: Vec<Vec<usize>>,
    dp_groups: Vec<Vec<usize>>,
}
\end{verbatim}
}

\subsection{Commitment structure}

The outer public commitment, re-stated here for reference, combines the \texttt{arch\_spec} hash with the Merkle roots of initial weights and dataset:
\[
  h_{\mathrm{commit}} = \mathrm{Hash}\!\big(\mathrm{Hash}(\mathtt{arch\_spec}) \;\|\; R(\mathtt{dataset}) \;\|\; R(W_0)\big).
\]
Any ex-ante attestation claims (\Cref{ex-ante-attestation}) extend this structure by appending $\mathrm{Hash}(\mathtt{ex\_ante\_claims})$.

\subsection{Converter pipeline}

Generating the \texttt{arch\_spec} from the trainer's actual PyTorch (or JAX) training code requires a converter pipeline; this is treated as its own open problem (see \opref{converter} in \Cref{app:open-problems}).

\section{Protocol-aligned formalisation of execution verification}\label{appendix:formalization}

\paragraph{Convention.} In this appendix and in \Cref{sec:formalisation-defs}, we revert to the cryptographic convention: the \emph{prover} is the trainer of \Cref{sec:proposed-solution}, and the \emph{verifier} is the auditor.

In the scope of this paper, our protocol in~\Cref{sec:verifying} proves a narrower object: \emph{relation-qualified execution-conformance} to a declared training-execution relation.
The verifier never evaluates a held-out loss \(L_{\cT}(M)\).
What is checked is that the public commitments, TAP tags, and zkVM proofs are jointly consistent with a witness for the committed run.
The PAC-style definitions in \Cref{sec:formalisation-defs} motivate a complexity-theoretic
viewpoint for a more general notion of \emph{AI model} verification that encompasses
learning capacity. This PAC formalisation adapts existing approaches~\citep{goldwasser_et_al:PACverif,amit2024:sefl-verif-models}, which focus mainly on ML models, to a circuit-valued domain, with soundness against computationally bounded provers.

\subsection{Background notions}\label{sec:formal-background}

\paragraph{NP relations and arguments.}
An NP relation is a polynomial-time decidable predicate \(\cR \subseteq \cX \times \cW\) on a public instance \(x \in \cX\) and a private witness \(w \in \cW\).
Here \(x\) is the public training transcript: commitments, tags, public hyperparameters, the genesis challenge, verifier signatures, and the opened sampling seed after streaming has been frozen.
The witness \(w\) is the private execution trace: architecture specification, compiled program, data, weights, batches, randomness, optimizer state, intermediate tensors, and traffic.
Because the protocol relies on computational assumptions, the proof object is formally an \emph{argument} rather than an information-theoretic proof: a polynomial-time cheating prover should not make the verifier accept a false instance except with the stated soundness error.

\paragraph{Commitments and Merkle binding.}
We use two kinds of commitments.
The tensor and weight commitments are Merkle-style commitments to values that the prover later opens inside audited proofs.
Their relevant property for soundness is binding: after a root has been published, the prover cannot open the same root to two different tensor values except with probability \(\varepsilon_{\mathrm{bind}}\).
The verifier's seed commitment \(\mathrm{Com}(\sigma;\rho)\) has two distinct roles.
Binding prevents the verifier from changing \(\sigma\) after the transcript is frozen (except with probability \(\varepsilon_{\mathrm{com,bind}}\)), while hiding prevents the prover from learning \(\sigma\) during training.
For soundness we use the prover-side prediction advantage \(\varepsilon_{\mathrm{com,pred}}\), not the verifier-side CZK hiding term \(\varepsilon_{\mathrm{com,hide}}\): for standard hiding commitments \(\varepsilon_{\mathrm{com,pred}}\) is bounded by the usual hiding advantage, but the games are conceptually different.

\paragraph{Commit-then-reveal sampling.}
Before training, the verifier samples a seed \(\sigma\) and publishes \(\mathrm{Com}(\sigma;\rho)\).
During training, the prover streams \((\mathrm{Com}(w_t),h_t)\) and the public anchor chain binds these values in order.
Only after the terminal anchor is signed by the verifier is \((\sigma,\rho)\) opened.
The step set \(S\), audited layers, and audited entries are then derived from \(F_\sigma\) using fixed-length encodings and separate tags
\[
\texttt{SAMP/STEP},\qquad \texttt{SAMP/LAYER},\qquad \texttt{SAMP/ENTRY}.
\]
This ordering is load-bearing: if \(\sigma\) is known before the stream is frozen, an adaptive prover can place deviations entirely outside the audited positions.
The terminal anchor freezes only the ordered public commitment stream.
It does not certify that the prover still stores the private opening material for every old commitment, so availability of sampled openings is a separate operational precondition of the spot-check protocol.

\paragraph{Knowledge soundness.}
Knowledge soundness strengthens plain soundness by requiring that an accepting prover can be converted into an extractor that outputs a witness for the accepted statement.
In this appendix the statement is deliberately qualitative.
For the BCS-/FRI-style compiled STARK layer, the theorem-level accounting keeps the quadratic random-oracle contribution; outer commitment-chain extraction and wrapper extraction are separate losses~\citep{bensasson2016iop,bensasson2018ethstark}.
We do not package these terms into a clean deployed-stack \(2^{-128}\) theorem.

\paragraph{Auxiliary-input computational zero knowledge.}
Auxiliary-input CZK means that for every polynomial-time verifier with auxiliary input \(z\), there is a simulator whose output is computationally indistinguishable from the real interaction, where the simulated object is the public transcript together with the accept/reject bit~\citep{goldreich2001foundations,lindell2015efficient}.
The claim made here is sequential and per-proof: it covers the proof objects in one audit session under the usual programmable-random-oracle modeling of Fiat--Shamir.
It is not a generic concurrent-composition, GUC, or EUC claim.

\subsection{Protocol-aligned formalisation}\label{sec:formalization}
We formalise and analyse security of the protocol of~\Cref{sec:verifying} 
in the same manner as for proof systems for NP relations.\footnote{PAC-style
generalisation is given in \Cref{sec:pac-formalisation}, which has the potential to go beyond NP relations and capture learnability, as previously done for ML models~\citep{amit2024:sefl-verif-models,goldwasser_et_al:PACverif}.} An \textbf{NP relation}
$\cR\subseteq\cX\times\cW$ pairs a public instance $x$ (commitments, tags,
hyperparameters, genesis challenge, signatures, opened seed) with a private witness
$w$ (\texttt{arch\_spec}, $P_{\mathrm{prog}}$, data, weights, batches, randomness,
optimizer state, intermediate tensors, traffic). \textbf{Merkle commitments} bind
tensor and weight values with binding error $\varepsilon_{\mathrm{bind}}$. The
verifier's \textbf{seed commitment} $\mathrm{Com}(\sigma;\rho)$ is additionally
hiding; for soundness we use the prover-side prediction advantage
$\varepsilon_{\mathrm{com,pred}}$ (bounded by hiding) rather than the CZK term
$\varepsilon_{\mathrm{com,hide}}$. In \textbf{commit-then-reveal}, the verifier
publishes $\mathrm{Com}(\sigma;\rho)$ before training; the prover streams
$(\mathrm{Com}(w_t),h_t)$ into the anchor chain; $(\sigma,\rho)$ opens only after
the verifier signs the terminal anchor; $S$ and audited positions are then derived
from $F_\sigma$ under tags $\texttt{SAMP/STEP}$, $\texttt{SAMP/LAYER}$,
$\texttt{SAMP/ENTRY}$. Required properties\footnote{Sequential, per-proof, under
programmable-RO modelling of Fiat--Shamir; no concurrent-composition, GUC, or EUC
claim.} include \textbf{(i) knowledge soundness} that allows extracting the witness from convincing provers (BCS/FRI accounting retains the quadratic
random-oracle term~\citep{bensasson2016iop,bensasson2018ethstark});
\textbf{(ii) auxiliary-input computational
ZK}~\citep{goldreich2001foundations,lindell2015efficient} — for every polynomial-time verifier with auxiliary input \(z\), there is a simulator whose output is computationally indistinguishable from the real interaction, where the simulated object is the public transcript together with the accept/reject bit;
\textbf{(iii) completeness} ensures that a prover knowing a valid witness can convince the verifier.
The relation against which {\bf (i)--(iii)} are stated is defined in~\Cref{sec:defs-proto-formal}.
Security analysis in view of the above errors $\varepsilon_{\mathrm{bind}}$, $\varepsilon_{\mathrm{com,pred}}$, and $\varepsilon_{\mathrm{com,bind}}$ is given in~\Cref{sec:protocol-sec-stmts}.

\subsection{Universal target relation and sampled verifier predicate}\label{sec:defs-proto-formal}

We keep separate the relation the protocol aims to certify from the predicate actually opened by the current sparse-auditing protocol.
The target relation \(\cR_{\mathrm{train}}\) is the universal execution relation over the full training run.
The current spot-check verifier checks a sampled restriction, denoted \(\cR_{\mathrm{spot}}^{S,Q}\), and soundness for \(\cR_{\mathrm{train}}\) pays the M2/M3 sampling terms below.

\begin{definition}[Universal training-execution relation]
Let \(x_{\mathrm{run}}\) be the public run instance containing:
(i) the pre-commitment
\[
R = H\!\left(H(\texttt{arch\_spec}) \concat \mathrm{MerkleRoot}(\mathrm{dataset}) \concat \mathrm{MerkleRoot}(w_0)\right),
\]
(ii) public hyperparameters \(\theta\), including \(T\) and the training schedule,
(iii) the verifier public key \(\mathsf{vk}_V\),
(iv) the genesis challenge batch and TAP tag,
(v) streamed commitments \(\{ \mathrm{Com}(w_t),h_t\}_{t=1}^T\), and
(vi) the terminal anchor \(\mathrm{anchor}_T\) and verifier signature \(\sigma_{\mathrm{chain}}\).

The witness \(w\) contains the private execution trace:
\(\texttt{arch\_spec}\), the compiled zkVM program \(P_{\mathrm{prog}}\), the dataset, the weight trajectory \(\{w_t\}_{t=0}^T\), the batches \(\{B_t\}_{t=1}^T\), prover hints, intermediate tensors, optimizer state, registered stochastic-component states, and the raw network traffic \(\{\mathrm{traffic}_t\}_{t=1}^T\).
In the base TAP model, the keyed-hash key \(K\) is a fixed run secret in the prover/TAP trust base and hidden from the verifier; the proof either treats \(K\) as a private witness value shared with the TAP or treats TAP-tag correctness as an external functionality assumption.

We write \((x_{\mathrm{run}},w) \in \cR_{\mathrm{train}}\) if all of the following hold:
\begin{enumerate}[label=(U\arabic*), leftmargin=2.4em]
\item \textbf{Commitment and anchor consistency.}
The pre-commitment equation defining \(R\) holds; each \(\mathrm{Com}(w_t)\) is the Merkle apex of the declared committed tensors and optimizer state for step \(t\); and the streamed values form the anchor chain
\[
\begin{aligned}
\mathrm{anchor}_0 &= H(\text{``ANCHOR/INIT''}\concat R),\\
\mathrm{anchor}_t &= H\bigl(\text{``ANCHOR/LINK''}\concat \mathrm{anchor}_{t-1}\\
&\qquad\concat \mathrm{Com}(w_t)\concat h_t\concat t\bigr).
\end{aligned}
\]
The terminal signature verifies:
\[
\begin{aligned}
\mathsf{Verify}(&\mathsf{vk}_V,
\text{``ANCHOR/LINK''}\concat R\concat T\concat \mathrm{anchor}_T,
\sigma_{\mathrm{chain}})=1.
\end{aligned}
\]

\item \textbf{Program/spec binding.}
The zkVM program is the publicly specified deterministic compilation of the committed architecture:
\[
P_{\mathrm{prog}} = f_{\mathrm{compile}}(\texttt{arch\_spec}).
\]

\item \textbf{Batch and RNG derivation.}
The batch schedule and every stochastic component output are derived from the committed dataset root, the master RNG seed, and the registered stochastic-component list in \(\texttt{arch\_spec}\).
Stateless randomness uses domain-separated keyed derivations such as \(\texttt{RNG/STEP}\) and \(\texttt{RNG/<COMP>}\) with fixed-length encodings.
Stateful stochastic rules, such as dynamic loss scaling, are updated by their declared deterministic transition functions.

\item \textbf{Genesis.}
The genesis execution on the verifier's challenge batch is consistent with \(P_{\mathrm{prog}}\), \(w_0\), and the TAP tag
\[
h_{\mathrm{gen}} = F_K(\mathrm{traffic}_{\mathrm{gen}}).
\]

\item \textbf{Universal step consistency.}
For every \(t \in [T]\), the committed transition from \((B_t,w_{t-1})\) to \(w_t\) is the transition prescribed by \(P_{\mathrm{prog}}\), and the produced traffic matches the TAP tag:
\[
\mathsf{Exec}(P_{\mathrm{prog}},B_t,w_{t-1};\mathbf{h}_t)=(w_t,\mathrm{traffic}_t),
\qquad
h_t=F_K(\mathrm{traffic}_t).
\]

\item \textbf{Operation-level consistency.}
Every committed operation value used in the transition is consistent with the relevant operation semantics.
For GEMMs this means bit-exact MAC-chain consistency against the committed operands; for nonlinear and lookup-heavy operations this means the corresponding precompile semantics for exp, sqrt/rsqrt, division, log, and BF16 lookup, composed along the attention and normalization paths.

\item \textbf{Layer-boundary consistency.}
Layer boundaries chain correctly: a layer's committed output root is the next layer's committed input root wherever the architecture declares such a boundary.
\end{enumerate}
\end{definition}

\begin{definition}[Sampled spot-check predicate]
Let the spot-check transcript be
\[
x_{\mathrm{spot}}=(x_{\mathrm{run}},\mathrm{Com}(\sigma),\sigma,\rho,S,Q).
\]
Here \(Q\) denotes all sampled layer and entry positions.
The predicate \(\cR_{\mathrm{spot}}^{S,Q}\) first checks
\[
\mathsf{Open}(\mathrm{Com}(\sigma),\sigma,\rho)=1,
\]
and then checks that \(S\), the audited layers, and the audited entries are exactly the outputs of \(F_\sigma\) under the fixed tags \(\texttt{SAMP/STEP}\), \(\texttt{SAMP/LAYER}\), and \(\texttt{SAMP/ENTRY}\), using the protocol's fixed-length encoding convention.
It checks (U1)--(U4) as public consistency conditions, and it checks (U5)--(U7) only at the sampled steps, layers, and entries specified by \((S,Q)\).
\end{definition}

Thus the current proof objects are arguments for \(\cR_{\mathrm{spot}}^{S,Q}\).
They are arguments for the universal target relation \(\cR_{\mathrm{train}}\) only up to the sampling miss terms in the soundness bound.
In a V5-IVC architecture, the step restriction \(t \in S\) is replaced by \(t \in [T]\); this removes the M3 branch but replaces it with recursive-proof, commitment, and extractor costs.
It does not by itself remove all operation-level sampling costs unless the per-step verifier is also strengthened.

\subsection{Security analysis for \Cref{sec:verifying}} \label{sec:protocol-sec-stmts}

\paragraph{For ex-ante attestations (\ref{ex-ante-attestation} of~\Cref{sec:verifying}) - Attack surface for compute under-declaration.}
Compute-threshold enforcement is our first target, and the adversary of interest
under-declares FLOPs to stay below a regulatory tier. Three cheating modes alter the
executed computation relative to the declared one: a larger model than declared, more
training data than declared, or more training steps than declared. Each has a cheap
detection path beyond the generic sampling argument. A
wider or deeper model fails the first weight Merkle-path opening because commitment
shapes do not match \texttt{arch\_spec}. Extra training data fails input-Merkle openings
at steps that use uncommitted samples: hiding a factor $\alpha_{\mathrm{data}}$ of extra data forces
$1 - 1/\alpha_{\mathrm{data}}$ of openings to fail, which two challenges catch at $99\%$ confidence for
$\alpha_{\mathrm{data}} = 10$. Extra training steps are visible to the network anchor, whose observed
collective count exceeds the committed chain length by a single integer comparison.
A compute-threshold attestation under these three detection mechanisms is cheaper to
enforce than generic $\texttt{arch\_spec}$ verification against an adaptive adversary.
These three do not exhaust the adversary's options: a trainer that executes the declared
computation faithfully while \emph{preparing its committed inputs} with undeclared
compute --- distillation from an undeclared model being the standard instance --- trips
none of the three detections, yet under-declares cumulative training compute. That
mode is analysed in \Cref{appx:proof-cost:defense-depth}; no cheap structural detection
for it is known, and it is the subject of open problems in
\Cref{app:open-problems} (in particular \opref{reachability} and \opref{compute-accounting}).
Other attestation classes (for instance data-content filters targeting semantic smuggling,
or procedural attestations on fine-grained regime compliance) lack the structural
detection surface of the three modes above and require denser sampling. A case-by-case analysis is already done in
\Cref{appx:proof-cost:defense-depth}.

\paragraph{For ex-ante attestations (\ref{ex-ante-attestation} of~\Cref{sec:verifying}) - Ex-post variant.}
Emergent properties of the completed run that are not knowable at initialisation (final-weight norms, benchmark performance under a public evaluator) admit a natural ex-post variant of the same $g(X) = y$ structure, executed once against the completed hash chain and final weights rather than maintained as a running invariant. Ex-post attestations do not modify the online-phase protocol; they reuse the precompile toolkit on final commitments.

\paragraph{Completeness.}
Suppose \((x_{\mathrm{run}},w)\in\cR_{\mathrm{train}}\), the prover follows the committed procedure honestly, the implementation satisfies the bit-exact determinism requirements of \Cref{enforce-determinism}, and the verifier derives \((S,Q)\) from a correctly opened \(\sigma\).
Then the verifier accepts the spot-check protocol with probability \(1-\negl(\lambda)\).

\emph{Proof sketch.}
All public commitments and anchor links verify by (U1).
The sampled predicate only restricts universal conditions (U5)--(U7), so every sampled transition, layer boundary, GEMM entry, and nonlinear/precompile check is correct.
The TAP tags match by (U4)--(U5), and zkVM completeness gives accepting genesis and per-step proofs except with negligible proof-system completeness error.

\paragraph{Five-branch soundness.}
Let \(\alpha_{\mathrm{step}}\) be the fraction of corrupted steps in a transcript that is
not in \(\cR_{\mathrm{train}}\). Over corrupted sampled steps, let
\(\alpha_{\mathrm{layer}}\) be a uniform lower bound on the fraction of corrupted audited
layers. Over corrupted audited operations, let \(\alpha_{\mathrm{entry}}\) be a uniform
lower bound on the fraction of corrupted output entries.
Let \(k_{\mathrm{step}},k_{\mathrm{layer}},k_{\mathrm{entry}}\) be the corresponding query budgets, and let \(N_{\mathrm{gemm,layer}}\) denote the number of audited GEMM-like operations per layer.
(The entry budget is per-operation-class in the deployed cost model --- e.g.\ the reduced attention budget of \Cref{appx:proof-cost:commitments} --- and the displayed bound should be read with the minimum applicable \(k_{\mathrm{entry}}\).)
For the current spot-check protocol,
\[
\begin{aligned}
\varepsilon_{\mathrm{total}}^{\mathrm{spot}} \le\;&
\underbrace{T \cdot \varepsilon_{\mathrm{bind}}}_{\mathbf{M1:}\ \mathrm{commitment\ binding}}
+ \underbrace{k_{\mathrm{step}}\!\left((1-\alpha_{\mathrm{layer}})^{k_{\mathrm{layer}}}
+ k_{\mathrm{layer}}N_{\mathrm{gemm,layer}}(1-\alpha_{\mathrm{entry}})^{k_{\mathrm{entry}}}\right)}_{\mathbf{M2:}\ \mathrm{intra\text{-}step\ auditing}} \\
&+ \underbrace{(1-\alpha_{\mathrm{step}})^{k_{\mathrm{step}}}}_{\mathbf{M3:}\ \mathrm{step\ sampling}}
+ \underbrace{(1+k_{\mathrm{step}})\varepsilon_{\mathrm{zkVM}}}_{\mathbf{M4:}\ \mathrm{proof\ soundness}}
+ \underbrace{\varepsilon_{\mathrm{com,pred}}+\varepsilon_{\mathrm{com,bind}}}_{\mathbf{M5:}\ \mathrm{commit\text{-}then\text{-}reveal}} \\
&+ \varepsilon_{\mathrm{PRF}}+\varepsilon_{\mathrm{sig}}.
\end{aligned}
\]
The five protocol mechanisms are M1--M5.
The PRF and signature terms are auxiliary cryptographic correction terms: \(\varepsilon_{\mathrm{PRF}}\) accounts for replacing the domain-separated sampler outputs by independent uniform samples, and \(\varepsilon_{\mathrm{sig}}\) accounts for forging the verifier's terminal-anchor signature.

\emph{Proof sketch.}
First condition on the commit-then-reveal event: the seed commitment hides \(\sigma\) from the prover until the stream is frozen and binds the verifier to the opened seed.
The complement of this event contributes M5,
\(\varepsilon_{\mathrm{com,pred}}+\varepsilon_{\mathrm{com,bind}}\).
Under this event and the PRF hybrid, \(S\), layers, and entries are uniform samples from their domains.
If the transcript is not in \(\cR_{\mathrm{train}}\), then one of the following must occur.
The prover equivocates on a committed tensor or anchor value, giving M1.
The sampled steps miss all corrupted steps, giving M3; the displayed binomial term is a convenient upper bound on the exact hypergeometric miss probability.
If a corrupted step is sampled, the sampled layers or entries miss the corrupted local operation values, giving the two M2 branches.
If the verifier accepts despite a locally false zkVM statement, the proof-system soundness event M4 occurs.
Finally, if the terminal chain can be changed after \(\sigma\) is known, the verifier signature has been forged or the protocol flow has been violated; the cryptographic part is \(\varepsilon_{\mathrm{sig}}\).
A union bound over these events gives the display.

At the paper's working point \(k_{\mathrm{step}}=10^3\) and \(\alpha_{\mathrm{step}}=10^{-2}\), and assuming the intra-step audit parameters are set so that the M2 branch is subdominant (corruption touching most audited layers and entries at the default budgets), the dominant term is still
\[
(1-\alpha_{\mathrm{step}})^{k_{\mathrm{step}}}\approx 4.3\times 10^{-5}.
\]
The current mainline is therefore sparse-auditing, detection-grade execution verification rather than a clean full-stack \(2^{-128}\) claim.

\paragraph{Knowledge soundness.}
Assume the inner zkVM proof system is a proof of knowledge in the programmable random-oracle model, with the BCS-/FRI-style extractor accounting just noted, and assume the commitment openings recorded in the public transcript are extractable except with their binding losses.
Then an accepting prover for the spot-check protocol can be converted into an extractor that outputs a witness \(w_{\mathrm{spot}}\) for \(\cR_{\mathrm{spot}}^{S,Q}\), except with the spot-check soundness error above and the inner-proof and outer-commitment extraction losses.
When the sampling events do not miss the actual deviations, this extracted sampled witness is consistent with the universal witness required for \(\cR_{\mathrm{train}}\).

\emph{Proof sketch.}
Run the proof-of-knowledge extractor on the accepted genesis proof and each accepted audited-step proof.
Oracle recording fixes the Fiat--Shamir challenges used by the proofs, and commitment binding fixes the Merkle openings and anchor-chain values to unique tensors except with the stated binding probabilities.
The extracted local witnesses agree on adjacent committed roots by the boundary checks and agree with the frozen transcript by the terminal signature.
This stitches the extracted local executions into a sampled trace witness.
The step from sampled trace to universal trace is exactly where M2 and M3 enter; without IVC or a stronger per-step verifier, knowledge soundness remains a qualified sampled-argument statement.
This appendix does not claim a referee-grade deployed-stack extractor theorem with a concrete \(2^{-128}\) end-to-end loss.

\paragraph{Sequential auxiliary-input CZK.}
For any non-uniform polynomial-time verifier with auxiliary input \(z\), the intended privacy target is computational zero knowledge for the proof objects viewed sequentially within one audit session.
The public leakage is the public transcript itself: \(R\), \(\theta\), training length, commitment roots, TAP tags, proof sizes, sampled positions after reveal, and the accept/reject outcome.
Under the usual honest-verifier-ZK predecessor plus Fiat--Shamir-preserves-ZK package, together with the keyed TAP channel and a CZK outer wrapper when used~\citep{groth2016size}, the verifier learns no additional witness information from the proof transcripts in this sequential/per-proof model.

\emph{Proof sketch.}
The simulator samples the public coins and programmed random-oracle answers needed by the Fiat--Shamir proofs, simulates each accepted proof object in sequence, and uses the hiding of the private commitments and the secrecy of the TAP key \(K\) to avoid exposing tensors, weights, data, optimizer state, or raw traffic.
Sequential composition allows the simulator to carry forward the public transcript state proof by proof.
The exact STARK/FRI setting still contains folklore edges in the bounded-query lift and the transplant of auxiliary-input random-oracle tables into the BCS simulator, and the illustrative concrete inner statistical-ZK residue is not a clean \(2^{-128}\) number.
Accordingly, this appendix uses only the qualitative sequential/per-proof consequence.
It makes no generic concurrent-composition claim and no shared-setup GUC/EUC closure claim.

\subsection{Operational caveats and upgrade boundary}

\paragraph{Sparse auditing is detection-grade, not universal.}
Bulk cheating is detected according to the M2/M3 branches above, and the M3 term is structural for a protocol that opens only \(k_{\mathrm{step}}\) sampled steps from a length-\(T\) stream.
A single carefully targeted bad step is caught with probability roughly \(k_{\mathrm{step}}/T\), not with cryptographic certainty.
Adaptive weighting or PAC-style interpretations, when added, remain overlays with extra assumptions rather than replacements for \(\cR_{\mathrm{train}}\)-conformance.

\paragraph{Lazy challenge-time commitments require a nested seed.}
The five-branch statement models a single commit-then-reveal audit seed. Tensors
committed only at challenge time (the lazy FFN intermediates of
\Cref{appx:proof-cost:commitments}) require a second, domain-separated entry seed
opened only after the lazy roots have been published; the M2/M5 accounting for
this nested round --- its own prediction and PRF losses, and the inclusion of the
lazy roots in the audited instance --- is not yet part of the displayed bound.
Absent the nested round, the $(1-\alpha_{\mathrm{entry}})^{k_{\mathrm{entry}}}$
factor does not apply to those tensors.

\paragraph{Domain-separated sampling is part of the statement.}
The step, layer, and entry samplers must be derived from the opened \(\sigma\) by a domain-separated PRF with fixed-length encodings.
The tags \(\texttt{SAMP/STEP}\), \(\texttt{SAMP/LAYER}\), and \(\texttt{SAMP/ENTRY}\) must not be reused for training RNG, anchor hashing, TAP MACs, or any other protocol role.
Without this discipline the M2/M3 probabilities in the theorem are not the probabilities of the implemented verifier.

\paragraph{Batch and RNG derivation are load-bearing.}
The relation requires the batch \(B_t\) and every registered stochastic component to be derived from the committed dataset root, the master RNG seed, and the deterministic transition rules in \(\texttt{arch\_spec}\).
This is what rules out a wrong-batch or RNG-divergence witness.
The verifier audit seed \(\sigma\) is separate from the master RNG seed in \(\texttt{arch\_spec}\); it only drives the sampled set \((S,Q)\) and is opened after the relevant commitment stream has been frozen.
If a component's randomness, dynamic state update, dtype conversion, or data-ordering convention is left outside the registered schedule, the appendix does not certify it.

\paragraph{Terminal anchors and rolling storage.}
Post-training spot checks remain valid only when the prover can open every sampled commitment.
Arbitrary whole-run post-training sampling therefore needs full retention, or deterministic replay from durable checkpoints together with committed replay metadata: optimizer state, data-loader state, stochastic-component state, kernel/reduction-tree and precision metadata, checkpoint interval commitments, and the committed seed schedule declared in \(\texttt{arch\_spec}\).
A strictly rolling deployment is therefore a separate windowed variant: each window needs its own challenge schedule, with the audit seed for that window committed before the window is frozen and opened before the required opening material is evicted.
If a sampled point cannot be opened, the verifier rejects rather than resampling.

\paragraph{Windowed audit variant.}
For proof-before-eviction storage, fix public windows \(I_j=\{a_j,\ldots,b_j\}\) and require either a per-window audit commitment \(c_j=\mathrm{Com}(\sigma_j;\rho_j)\) posted publicly before step \(a_j\), or a pre-registered public-beacon rule fixed before the window starts.
At window close, a signed or timestamped freeze certificate binds the interval, previous anchor, frozen window endpoint \(A_j\) (for example \(A_j=\mathrm{anchor}_{b_j}\)), the seed commitment or beacon rule, sampler parameters, and opening deadlines; this freeze must be public before \((\sigma_j,\rho_j)\) is opened or the beacon output is used.
The opened seed or beacon output derives \(S_j\subseteq I_j\), and every sampled proof/opening, or exact committed replay producing that opening, must arrive before the window proof deadline and before private material for \(I_j\) is evicted.
Missing seed openings, missing sampled openings, failed replay, late proofs, or abort after challenge reveal are rejection/non-compliance events, never occasions to resample or shrink \(S_j\).
In such a theorem, M3 and M5 are window-indexed; a full statement needs per-window soundness accounting together with an explicit timing/bulletin-board model, and does not inherit the theorem statement or numerical bound of the single terminal-anchor protocol.

\paragraph{TAP key custody is an assumption in the base model.}
The keyed tag \(h_t=F_K(\mathrm{traffic}_t)\) protects against bounded-entropy traffic-guessing by a verifier who knows plausible traffic patterns but not \(K\).
In the base model, \(K\) lives in the prover/TAP trust base for the run and is hidden from the verifier.
This appendix does not close TAP compromise, prover--TAP collusion, key-rotation ambiguity, side channels, or vendor-root questions.
TAP-TEE changes the custody and attestation story for \(K\), but it is an upgrade path with its own attestation, side-channel, and manufacturer-trust assumptions, not a mainline theorem here.

\paragraph{Concrete cryptographic residues remain.}
The knowledge-soundness and CZK paragraphs are intentionally scoped.
The inner BCS/FRI knowledge extractor keeps the quadratic random-oracle term; the outer commitment-layer extraction story is separate.
The sequential CZK story is theorem-composed only after keeping the bounded-query and auxiliary-input simulator-transplant edges explicit.
The concrete ZK residue at illustrative parameters blocks any clean deployed-stack \(2^{-128}\) slogan until the parameter story is tightened.

\paragraph{V8, TAP-TEE, and V5-IVC target different gaps.}
Exact V8 targets genesis-weight provenance, but a paper-ready version still needs public initialization metadata, registry-grade initializer semantics, canonical serialization, dtype/cast rules, and normal-family sampler specifications.
TAP-TEE targets telemetry custody and provenance, not the dominant statistical M3 branch.
V5-IVC targets step coverage and can remove M3 only by replacing sparse spot-checking with universal per-step proof coverage; it then inherits recursive-verifier, outer-PCS, binding, and extraction costs.
None of these upgrades is treated here as deployed-stack ready, and a Groth16 outer wrapper does not retroactively tighten the inner IVC or sampled-verifier soundness terms.

\paragraph{Composition boundary.}
This appendix makes no generic concurrent-composition, shared-random-oracle, shared-CRS, GUC, or EUC closure claim.
The safe paper posture is the sequential/per-proof one stated above, with stronger shared-setup composability left as future work.

\subsection{PAC-style threat model and formalisation}\label{sec:pac-formalisation}
Generalising the formalisation in~\Cref{sec:formalization}, we extend AI verification
beyond execution relations to encompass PAC-style learnability. We then state precisely
which relations we seek to verify under this formalisation. Our modelling and formalisation
use the framework of \emph{computational complexity}. The architecture in
\Cref{pre-commit,constant-hashing,network-tap} ultimately verifies faithful execution of a
committed training procedure rather than low loss on an external distribution. Accordingly,
\Cref{sec:formalization} states the protocol-aligned execution-verification relation and
its soundness, knowledge-soundness, and zero-knowledge guarantees.

\subsubsection{Definitions}\label{sec:formalisation-defs}
In this part, we take a \emph{computational-complexity-theoretic} approach to AI training
verification. This approach is inspired by recent work on verification in learning theory.
The work in~\citep{goldwasser_et_al:PACverif} provides a foundation for \emph{PAC
verification}, in which a \emph{verifier} checks whether a learning algorithm, acting as a
\emph{prover}, has produced a near-optimal hypothesis.\footnote{Machine-learning algorithms
provide examples of systems for these PAC-verification applications.} The work
in~\citep{amit2024:sefl-verif-models} provides a foundation for \emph{self-proving models},
in which learning algorithms acting as provers successfully convince a verifier that their
answers are correct.\footnote{Large language models provide examples of systems for these
self-proving applications.} Both works draw on a computational-complexity perspective in
their theoretical modelling.
In the seminal work of~\citep{goldwasser_et_al:PACverif} it is stated that \emph{``The P vs. NP problem asks whether finding a solution ourselves is harder than verifying a solution supplied by someone else. It is natural to ask a similar question in learning theory: Are there machine learning problems for which learning a good hypothesis is harder than
verifying one proposed by someone else? We find this theoretical motivation compelling in
and of itself.''}

We propose a similar conceptual approach in our formalisation, i.e., based on computational complexity, for a similar perspective:
because it is widely believed to be \emph{hard} to train an AI model for some specific task so that it can be \emph{safely} deployed at a national/international level, 
is there a difficulty gap between such a training process and verifying one proposed by some third party?
And it is computational complexity that provides us with the better lens to look into these questions of hardness and verification.
The PAC-style definitions below are a learning-theoretic motivation and comparison point.
They are not the cryptographic theorem proved by the protocol in this paper: \Cref{appendix:formalization} states the narrower execution-conformance claim, namely consistency with a declared training-execution relation.

\paragraph{Our setting of training verification.} Let \(\cM \subseteq \zo^\cC\) be a set of models, as a subset of some family of Boolean circuits \(\cC\defeq \{C_{n,d}\}_{n,d \in \N}\) indexed by size \(n\) and depth \(d\). 
Let \(\cT\) be a distribution over \(\cC \times \zo\). 
This represents the desirable results of the training process. The circuits \(C_{n,d}\) such that \(\cT(C_{n,d}) = 1\) are obtained after training.
The prover \(P\) and verifier \(V\) have access to their own \emph{oracles} \(\cO_P\) and \(\cO_V\) respectively. 
In our prototype later, \(\cO_P\) represents a zkVM, for instance.
The prover \(P\) and verifier \(V\) have access to a private source of random coins as well, denoted \(r_P \in \zo^*\) and \(r_V\in \zo^*\) in that order.

A \emph{training verification protocol} can consist of an \emph{offline phase} and an
\emph{online phase}. During the offline phase, \(P\) publishes a \emph{common reference
string} \(\crs\). During the online phase, \(P\) and \(V[\crs]\) take turns sending each
other messages \(w_1,w_2,\dots\); here \(V[\crs]\) indicates that \(V\) has access to the
published \(\crs\). At the end of the online phase, \(V[\crs]\) outputs either a model
\(\ctildeM \in \cM\) or ``reject''. For example, \(P\) can send \(V[\crs]\) a trained
model \(\ctildeM\) during the online phase and try to convince \(V[\crs]\) of its properties
with respect to \(\cT\). If \(\ctildeM\) is ``good'', meaning
\(\cT(\ctildeM)=1\), then \(V[\crs]\) outputs \(\ctildeM\); otherwise, it rejects.


\begin{myremark}
  First, we represent the models to be verified as Boolean circuits. This representation is
  nonuniform because we later allow precompiled algorithms with hints to be parts of these
  models: different hints at different input lengths produce different behaviours. The
  distribution \(\cT\) encompasses the properties/relations that we would like to check
  on a given model \(C_{n,d}\). This generalises the notion of relations usually seen in the
  context of NP problems. For example, one can parameterise \(\cT\) with threshold
  constraints and consider a ``good'' model \(C_{n,d}\) that satisfies
  \(\cT(C_{n,d})=1\).
\end{myremark}

\paragraph{Our definitions.} We start the formal definitions.
\begin{definition}[Agnostic trainability]\label{def:agnos-trainable}
  A class of models \(\cM\) is \emph{\(\alpha\)-agnostically trainable} if there exists a circuit \(T\)
  such that for every distribution \(\cT\) over \(\cC\times \zo\) and every \(\epsilon,\delta > 0\),
  \(T\) outputs, with probability \(1-\delta\) over its random coins, a model \(M\) that satisfies
  \[
    L_\cT(M) \le \alpha\cdot L_\cT(\cM) +\epsilon
  \]
  where \(L_\cT: \zo^\cC \to [0,1]\) is a \emph{loss function} that is defined as \(L_\cT(M) \defeq \Pr_{(C,y) \gets \cT}[M(C) \neq y]\)
  and \(L_\cT(\cM) \defeq \inf_{M\in \cM}(L_{\cT}(M))\).
\end{definition}
\Cref{def:agnos-trainable} is inspired by PAC learnability. With high probability, the
circuit \(T\) outputs a trained model that is \(\epsilon\)-close to the best trained model
with respect to \(\cT\); the qualifier ``agnostic'' allows a multiplicative factor
\(\alpha\) on the best loss with respect to \(\cT\).
The agnostic aspect follows Definition 4 of~\citep{goldwasser_et_al:PACverif}, adapting it to a circuit-valued domain and stating soundness against computationally bounded provers (a weaker guarantee than the statistical soundness there).

We now proceed to define what is \emph{training verifiability}.
\begin{definition}[Training verifiability]\label{def:training-verifblt}
  We say that \(\cM\) is \emph{\(\alpha\)-verifiable} with respect to a family of distributions \(\mathfrak{T}\)
  containing distributions \(\cT\) over \(\cC\times \zo\) using oracles \(\cO_P, \cO_V\) if there exist probabilistic polynomial-time (PPT) Turing machines \((P,V)\)
  such that for all \(\delta, \epsilon > 0\)
  \begin{description}
    \item[Completeness] For any \(\cT \in \mathfrak{T}\), the random variable \(m \defeq [P^{\cO_P}(\delta,\epsilon),V^{\cO_V}(\delta,\epsilon)]\) satisfies
    \[
      \Pr[m \neq \text{ ``reject''} \wedge L_\cT(m) \le \alpha\cdot L_\cT(\cM) +\epsilon] \ge 1 - \delta\enspace.
    \]
    \item[Soundness] For any \(\cT \in \mathfrak{T}\), for any PPT \(P'\), the random variable \(m \defeq [{P'}^{\cO_P}(\delta,\epsilon),V^{\cO_V}(\delta,\epsilon)]\) satisfies
    \[
      \Pr[m \neq \text{ ``reject''} \wedge  L_\cT(m) > \alpha\cdot L_\cT(\cM) +\epsilon] \le \delta\enspace.
    \]
    
  \end{description}
  In the above the random variable \(m \defeq [P^{\cO_P}(\delta,\epsilon),V^{\cO_V}(\delta,\epsilon)]\) denotes the output of the verifier after the completion of the protocol.
\end{definition}

\paragraph{Relation to prior models.}
The setting of \Cref{def:agnos-trainable,def:training-verifblt} is \emph{designed to subsume}
the PAC-verification~\citep{goldwasser_et_al:PACverif} and
self-proving~\citep{amit2024:sefl-verif-models} settings. If the two reductions hold, the
setting would provide a useful common frame rather than a third incomparable one. Formally
establishing those reductions is left to future work.

\section{Glossary}\label{app:glossary}

Entries in the proof-system and protocol subsections follow the cryptographic
convention of \Cref{appendix:formalization}: the \emph{prover} is the trainer of
\Cref{sec:proposed-solution}, and the \emph{verifier} is the auditor.

\subsection{Hardware and distributed training}

\begin{description}[style=nextline, leftmargin=3.7cm, labelwidth=3.5cm]
  \item[\glstarget{cublas}{cuBLAS}] NVIDIA's GPU-accelerated linear algebra library, used here for GEMM-heavy training kernels.
  \item[\glstarget{cudnn}{cuDNN}] NVIDIA's deep learning primitives library, providing optimised GPU implementations of neural-network operations such as convolutions, normalization, and pooling.
  \item[\glstarget{nccl}{NCCL}] NVIDIA Collective Communication Library, the standard library used for multi-GPU and multi-node collectives such as allreduce, allgather, and reduce-scatter.
  \item[\glstarget{nvlink}{NVLink}] High-bandwidth GPU-to-GPU interconnect used for in-node transfers.
  \item[\glstarget{nvswitch}{NVSwitch}] In-node switching fabric that connects GPUs over NVLink and supports high-bandwidth any-to-any communication.
  \item[\glstarget{nvls}{NVLS}] NVLink SHARP, an NCCL path that can offload reductions to NVSwitch hardware and impose a fixed reduction order for supported collectives.
  \item[\glstarget{allreduce}{Allreduce}] Collective operation that combines tensors across participants and returns the combined tensor to each participant. Floating-point allreduce can be order-sensitive.
  \item[\glstarget{allgather}{Allgather}] Collective operation that gathers tensor shards from all participants so that each participant receives the concatenated result.
  \item[\glstarget{reduce-scatter}{Reduce-scatter}] Collective operation that reduces tensors across participants and scatters disjoint output shards back to them.
  \item[\glstarget{all-to-all}{All-to-all}] Collective communication pattern in which each participant sends a distinct shard to every other participant.
  \item[\glstarget{fsdp}{FSDP}] Fully Sharded Data Parallelism: a data-parallel training scheme that shards model state across devices and uses collectives to materialize and reduce shards.
  \item[\glstarget{tensor-parallelism}{Tensor parallelism}] Parallelism that splits individual tensor operations, typically matrix multiplications, across devices and often requires allreduce within a layer.
  \item[\glstarget{pipeline-parallelism}{Pipeline parallelism}] Parallelism that assigns different layers or layer blocks to different devices and streams microbatches through the resulting pipeline.
  \item[\glstarget{expert-parallelism}{Expert parallelism}] Parallelism for mixture-of-experts models in which experts are distributed across devices and token routing induces all-to-all traffic.
  \item[\glstarget{hbm}{HBM}] High-Bandwidth Memory: stacked DRAM packaged on the GPU module, providing the bulk of training-tensor storage and the highest-bandwidth path between compute and memory.
  \item[\glstarget{smartnic}{SmartNIC}] Programmable network interface card with on-card compute that can perform packet inspection, telemetry, hashing, or attestation in line with traffic, used here as a tier of network anchoring.
  \item[\glstarget{dpu}{DPU}] Data Processing Unit: a programmable accelerator (often combined with a SmartNIC) that offloads networking, storage, and security functions from the host CPU.
  \item[\glstarget{tee}{TEE}] Trusted Execution Environment: an isolated processor mode (e.g.\ Intel TDX, NVIDIA Confidential Computing) providing remote attestation and integrity guarantees for code and data executed inside it.
\end{description}

\subsection{ZK and proof systems}

\begin{description}[style=nextline, leftmargin=3.7cm, labelwidth=3.5cm]
  \item[\glstarget{pac}{PAC}] Probably approximately correct: a learning-theoretic guarantee over a distribution. In this paper PAC-style claims are separate from the cryptographic execution-conformance statement.
  \item[\glstarget{np-relation}{NP relation}] Polynomial-time decidable relation between a public instance and a private witness; the proof system argues that such a witness exists.
  \item[\glstarget{zk}{ZK}] Zero knowledge: a privacy property saying that the verifier's view of the proof can be efficiently simulated from the public statement and the accept/reject outcome alone, so the proof conveys no witness information.
  \item[\glstarget{zkvm}{zkVM}] Zero-knowledge virtual machine: a proof environment that proves correct execution of a guest program. In this paper the guest is the proof checker, not the full training run.
  \item[\glstarget{snark}{SNARK}] Succinct non-interactive argument of knowledge: a proof system with short proofs and efficient verification for NP statements, under system-specific assumptions and setup.
  \item[\glstarget{stark}{STARK}] Scalable transparent argument of knowledge, typically a hash-based proof system using polynomial commitment and low-degree testing machinery such as FRI.
  \item[\glstarget{fri}{FRI}] Fast Reed--Solomon interactive oracle proof of proximity, the low-degree testing component used in many STARK constructions.
  \item[\glstarget{bcs}{BCS transform}] Ben-Sasson--Chiesa--Spooner compilation pattern that turns an interactive oracle proof into a non-interactive argument using commitments and Fiat--Shamir challenges.
  \item[\glstarget{pcs}{PCS}] Polynomial commitment scheme: a commitment primitive for polynomials with succinct opening proofs at selected points.
  \item[\glstarget{groth16}{Groth16}] Pairing-based SNARK with very small proofs and fast verification, requiring a structured common reference string for the proved circuit.
  \item[\glstarget{fiat-shamir}{Fiat--Shamir}] Transformation that makes a public-coin interactive proof non-interactive by deriving verifier challenges from a hash of the transcript.
  \item[\glstarget{grinding}{Grinding}] Attack on hash-derived (Fiat--Shamir) challenges in which the prover repeatedly re-randomises its committed transcript and recomputes the challenge derivation offline until the derived challenges avoid its deviation; the security loss is bounded by the prover's hash-query budget.
  \item[\glstarget{transferability}{Transferability}] Property of a verification transcript that it convinces third parties, not only the original verifier. Third parties need authenticated evidence and publicly checkable claims. For sampled audits, they also need assurance about challenge selection (\Cref{sec:proof-options}). Publishing an auditor's seed does not establish non-collusion with the trainer.
  \item[\glstarget{designated-verifier}{Designated-verifier}] Proof or audit transcript that convinces only a specific verifier (for instance because the challenges came from that verifier's private coins); third parties must additionally trust the designated verifier. Orthogonal to zero knowledge, which concerns privacy.
  \item[\glstarget{detection-grade}{Detection-grade}] Assurance level of a sampled protocol: deviations in the covered class are caught with high probability under the stated sampling distribution, rather than being infeasible except with cryptographically negligible probability. Contrasts with cryptographic-grade soundness; the floors of \Cref{sec:sampling-guarantees} quantify the gap.
  \item[\glstarget{vdf}{VDF}] Verifiable delay function: a function whose evaluation requires a prescribed amount of inherently sequential work but whose output is efficiently verifiable; a building block for public randomness beacons resistant to last-moment manipulation.
  \item[\glstarget{hybrid-protocol}{Hybrid verification protocol}] Proposed two-tier
  protocol family that remains under investigation (\Cref{sec:hybrid}). It is intended to
  bind every coefficient of every step at a declared tolerance width (full coverage,
  relaxed precision) and to recover exactness through a sparse, unpredictable schedule of
  bit-exact spot-checks and randomly scheduled exact anchors.
  \item[\glstarget{random-oracle}{Random oracle}] Idealized hash-function model in which each new query receives an independent random response, with repeated queries answered consistently.
  \item[\glstarget{knowledge-soundness}{Knowledge soundness}] Property that an accepting prover can be converted, except with stated losses, into an extractor that outputs a witness for the accepted statement.
  \item[\glstarget{computational-zk}{Computational zero knowledge (CZK)}] Privacy property requiring that proof transcripts can be efficiently simulated up to computational indistinguishability, revealing no witness information beyond the public transcript.
  \item[\glstarget{auxiliary-input}{Auxiliary input}] Extra side information supplied to an adversarial verifier or distinguisher; auxiliary-input security asks that the guarantee survive such side information.
  \item[\glstarget{ivc}{IVC}] Incrementally verifiable computation: recursive proof composition for a long computation, maintaining a compact proof as steps are appended.
  \item[\glstarget{crs}{CRS}] Common reference string: public proof-system parameters, sometimes generated by a trusted or structured setup depending on the proof system.
  \item[\glstarget{public-instance}{Public instance}] Public statement and inputs checked by the verifier, such as commitment roots, tags, parameters, and proof-system public inputs.
  \item[\glstarget{witness}{Witness}] Private data that makes the public instance true; here it may include tensors, weights, batches, randomness, optimizer state, and traffic.
  \item[\glstarget{precompile}{Precompile}] Specialized proof-system circuit or primitive for a fixed operation, such as a MAC chain or Merkle path check, called by the proof checker instead of emulating the operation generically.
  \item[\glstarget{extractor}{Extractor}] Algorithm used in a knowledge-soundness argument to recover a witness from a prover that convinces the verifier.
  \item[\glstarget{guc-euc}{GUC/EUC}] Generalized or externalized universal-composability frameworks for protocols with shared setup or global state. The appendix explicitly does not claim these composition guarantees.
\end{description}

\subsection{Protocol and commitments}

\begin{description}[style=nextline, leftmargin=3.7cm, labelwidth=3.5cm]
  \item[\glstarget{tap}{TAP}] Traffic Access Point: a passive observation point on training-cluster links. In the base protocol it records or tags traffic rather than deciding whether the run is valid.
  \item[\glstarget{tap-tag}{TAP tag}] Keyed digest of observed traffic used as a public binding value for later proof checks.
  \item[\glstarget{tap-tee}{TAP-TEE}] Upgrade path in which TAP key custody or telemetry attestation is moved into a trusted execution environment (TEE) or similar hardware root. It changes trust assumptions rather than the sampled-auditing theorem.
  \item[\glstarget{prf}{PRF}] Pseudorandom function: a deterministic keyed function whose outputs are computationally indistinguishable from random to parties without the key.
  \item[\glstarget{keyed-prf}{Keyed PRF}] A PRF evaluated under a secret key \(K\), used here for domain-separated tags or sampler outputs depending on the protocol role.
  \item[\glstarget{commit-then-reveal}{Commit-then-reveal}] Pattern in which a value, such as a sampling seed, is first committed and only later opened after the relevant transcript has been fixed.
  \item[\glstarget{audit-seed}{Audit seed}] Verifier or public randomness used to derive sampled steps, layers, or entries after the relevant commitment stream has been frozen. It is distinct from the training RNG seed.
  \item[\glstarget{merkle-root}{Merkle root}] Root hash of a Merkle tree, serving as a compact public digest of many leaves.
  \item[\glstarget{merkle-commitment}{Merkle commitment}] Commitment to a vector or tensor represented by a Merkle tree; individual entries can be opened with authentication paths to the public root.
  \item[\glstarget{anchor-chain}{Anchor chain}] Hash chain linking per-step commitments, TAP tags, and run metadata in order, so later entries bind to the earlier transcript.
  \item[\glstarget{terminal-anchor}{Terminal anchor}] Final value of the anchor chain, used as the point at which the streamed transcript is frozen before challenge material is opened.
  \item[\glstarget{windowed-audit}{Windowed audit}] Rolling-storage variant that partitions the run into public windows, freezes each window before revealing its audit seed, and requires sampled openings before private material is evicted.
  \item[\glstarget{window-root}{Window root}] Commitment root, digest, or anchor for the public step records in one window. It freezes the public record stream, not the private witness material.
  \item[\glstarget{freeze-certificate}{Freeze certificate}] Signed or timestamped public message binding a window interval, previous anchor, current window root or anchor, seed commitment, sampler parameters, deadlines, and protocol context before seed reveal.
  \item[\glstarget{bulletin-board}{Bulletin board}] Append-only public log or ordering service for protocol events such as seed commitments, freeze certificates, seed openings, proof submissions, aborts, and timeouts.
  \item[\glstarget{logical-clock}{Logical clock}] Monotone counter or timestamp used by a bulletin board to compare event order. It is a public ordering abstraction, not a claim about synchronized hardware clocks.
  \item[\glstarget{public-beacon}{Public beacon}] External randomness source whose output is unpredictable before a specified time and publicly verifiable after release; it can replace a hidden verifier seed in some windowed variants.
  \item[\glstarget{quorum}{Quorum}] Threshold set of verifiers or authorities whose joint action authorizes a freeze, randomness contribution, or audit decision under an explicitly stated threshold policy.
  \item[\glstarget{domain-separation}{Domain separation}] Use of distinct labels or encodings so that hashes or PRF outputs for one protocol role cannot be reused as another role.
  \item[\glstarget{poseidon}{Poseidon}] Hash function designed to be efficient inside algebraic proof systems; used here for Merkle-path checks in the proof circuit.
  \item[\glstarget{spot-checking}{Spot-checking}] Sampled verification of selected steps, layers, or entries rather than exhaustive checking; soundness then includes the probability that sampling misses a deviation.
  \item[\glstarget{m1-m5}{M1--M5 branches}] Shorthand for the five soundness mechanisms tracked in the appendix: commitment binding, intra-step auditing, step sampling, proof soundness, and commit-then-reveal.
  \item[\glstarget{opening-material}{Opening material}] Private data needed to open and prove a sampled statement, including committed values, Merkle paths, commitment randomness when relevant, hints, stochastic state, and proof witnesses.
  \item[\glstarget{full-retention}{Full retention}] Storage mode in which private opening material for every challengeable step is kept until the audit seed has been opened and all sampled proofs have been accepted or rejected.
  \item[\glstarget{deterministic-replay}{Deterministic replay}] Reconstruction of sampled opening material from committed checkpoints and metadata as a single-valued, bit-exact function of the frozen transcript. Failed replay is a rejection event, not a reason to resample.
  \item[\glstarget{checkpoint}{Checkpoint}] Durable committed snapshot used for recovery or replay; for training verification it must include enough model, optimizer, RNG, data-loader, kernel, and metadata state to reproduce relevant commitments.
  \item[\glstarget{proof-before-eviction}{Proof-before-eviction}~] Rolling-window discipline in which a window is frozen, sampled, and proved before the prover deletes the private material needed to answer challenges about that window.
  \item[\glstarget{acceptance-watermark}{Acceptance watermark}] Operational status saying that a window's required sampled proofs have been accepted, or the window has publicly failed; eviction policies should be keyed to this status rather than to age alone.
  \item[\glstarget{v5-ivc}{V5-IVC}] Roadmap shorthand for an IVC-based upgrade path that aims at universal per-step coverage rather than sampled step coverage. It is referenced as a target, not as a deployed theorem here.
  \item[\glstarget{v8}{V8}] Roadmap shorthand for an upgrade path targeting genesis-weight provenance and initialization semantics. It is outside the appendix's main theorem.
\end{description}

\subsection{ML training and numerical terms}

\begin{description}[style=nextline, leftmargin=3.7cm, labelwidth=3.5cm]
  \item[\glstarget{arch-spec}{\texttt{arch\_spec}}] Architecture specification consumed by the proof checker, declaring model operations, precision, communication pattern, data-loading parameters, and commitment boundaries.
  \item[\glstarget{bf16}{BF16}] Bfloat16 floating-point format with a compact mantissa and an FP32-sized exponent range; training commonly uses BF16 values with FP32 accumulation.
  \item[\glstarget{fp32}{FP32}] IEEE 754 single-precision floating point, used here for accumulators and selected nonlinear operations.
  \item[\glstarget{rne}{RNE}] Round-to-nearest-even, the IEEE 754 rounding mode assumed by the numerical checks unless otherwise stated.
  \item[\glstarget{gemm}{GEMM}] General matrix-matrix multiplication, the dense linear-algebra primitive underlying most Transformer linear layers and their gradients.
  \item[\glstarget{mac-chain}{MAC chain}] Ordered multiply-accumulate computation for a dot product, including the rounding behavior of each multiplication and accumulation step.
  \item[\glstarget{training-rng-seed}{Training RNG seed}] Random-number-generator seed, usually committed through \texttt{arch\_spec}, that determines batches and stochastic training components. It is separate from the verifier's audit seed.
  \item[\glstarget{stochastic-component}{Stochastic component}] Declared source of training randomness or stateful stochastic behaviour, such as dropout, data augmentation, stochastic rounding, dynamic loss scaling, or router noise.
  \item[\glstarget{hint}{Hint}] Private value supplied by the host to the zkVM guest; it is not trusted until checked against commitments and operation-specific constraints.
  \item[\glstarget{host}{Host}] The trainer-side process or GPU environment that performs the native computation and supplies private hints to the proof checker.
  \item[\glstarget{guest}{Guest}] The program executed inside the zkVM, here the proof checker that validates committed hints and sampled operations.
  \item[\glstarget{fused-operation}{Fused operation}] GPU kernel or operation block whose internal intermediates are not separately materialized; the proof treats it through declared inputs, outputs, and operation semantics.
\end{description}


\newpage
\end{document}